%% file: main.tex
\documentclass{article}
\UseRawInputEncoding

\usepackage{PRIMEarxiv}
\usepackage{url}
\usepackage{multirow}
\usepackage{multicol}
\usepackage{makecell}
\usepackage{csvsimple}
\usepackage{siunitx}
\usepackage{bbm}
\usepackage{bbold}
\usepackage{amsmath}
\usepackage{amssymb}

\usepackage{graphicx}
\graphicspath{ {./images/} }

\usepackage{subcaption}
\usepackage[font=small]{caption}

\usepackage{algorithm}
\usepackage{algpseudocode}

\usepackage{rotating}
\usepackage[T1]{fontenc}

\usepackage{caption} 
\usepackage[stable]{footmisc}

\author{%
\begin{tabular}{c} Chenhao Tong \\ The University of Melbourne \\ Melbourne, VIC \\ chetong@unimelb.edu.au \end{tabular} \and
\begin{tabular}{c} Aaron Harwood \textsuperscript{\textsection} \\ The University of Melbourne \\ Melbourne, VIC \\ arnhrwd@gmail.com \end{tabular}
\and
\begin{tabular}{c} Maria A. Rodriguez \\ The University of Melbourne \\ Melbourne, VIC \\ marodriguez@unimelb.edu.au \end{tabular} 
\and
\\
\begin{tabular}{c} Richard O. Sinnott \\ The University of Melbourne \\ Melbourne, VIC  \\ rsinnott@unimelb.edu.au\\
\end{tabular}}

\begin{document}

\title{An Energy-aware and Fault-tolerant Deep Reinforcement Learning-based Approach for Multi-agent Patrolling Problems}

\maketitle

\begingroup\renewcommand\thefootnote{\textsection}
\footnotetext{Work done while the author was at the University of Melbourne.}
\endgroup

\begin{abstract}
    
Autonomous vehicles are suited for continuous area patrolling problems. However, finding an optimal patrolling strategy can be challenging for many reasons. Firstly, patrolling environments are often complex and can include unknown environmental factors, such as wind or landscape. Secondly, autonomous vehicles can have failures or hardware constraints, such as limited battery life. Importantly, patrolling large areas often requires multiple agents that need to collectively coordinate their actions. In this work, we consider these limitations and propose an approach based on model-free, deep multi-agent reinforcement learning. In this approach, the agents are trained to patrol an environment with various unknown dynamics and factors. They can automatically recharge themselves to support continuous collective patrolling. A distributed homogeneous multi-agent architecture is proposed, where all patrolling agents execute identical policies locally based on their local observations and shared location information. This architecture provides a patrolling system that can tolerate agent failures and allow supplementary agents to be added to replace failed agents or to increase the overall patrol performance. The solution is validated through simulation experiments from multiple perspectives, including the overall patrol performance, the efficiency of battery recharging strategies, the overall fault tolerance, and the ability to cooperate with supplementary agents.
\end{abstract}


\input{01_introduction.tex}

\input{02_related_work.tex}

\input{03_preliminary.tex}

\input{04_modelling.tex}

\input{05_methods.tex}

\input{06_evaluation.tex}

\clearpage
\section{Conclusion}
This work has proposed a deep multi-agent reinforcement learning-based approach for multi-agent patrolling problems that encompasses battery and failure constraints of patrolling agents, which are essential for providing continuous patrolling performance. A distributed homogeneous multi-agent architecture is proposed where agents execute identical policies. This supports a patrolling system that can tolerate agent failures and allow supplementary agents to be added to either replace failed agents or increase the patrolling performance. We use a modified state-of-the-art reinforcement learning algorithm (Proximal Policy Optimization) to train the patrolling agents, and the simulation experiments show that our approach can successfully reduce both the average and maximum idleness incurred during patrolling. The performance surpasses the Conscientious Reactive (CR) strategy and the individual learner approach. In addition, the system's fault tolerance and the ability of existing agents to cooperate with supplementary agents have been demonstrated.

There are several areas that could be explored in future extensions to the work, including:
\begin{itemize}
    \item taking into account the uncertainties in agents' communication; 
    \item dealing with non-fully-cooperative scenarios, e.g. agents may collide.
    \item deploying the solution to real-world patrolling vehicles.
\end{itemize}

\clearpage
\section{Acknowledgement}
This research was supported by the use of the Nectar Research Cloud, a collaborative Australian research platform supported by the NCRIS-funded Australian Research Data Commons (ARDC) \\

This study was supported by the \emph{Melbourne Research Scholarship} from the Univerisity of Melbourne, VIC, Australia.\\

This research was undertaken using the LIEF HPC-GPGPU Facility hosted at the University of Melbourne. This Facility was established with the assistance of LIEF Grant LE17a0100200. \\

\clearpage
\input{07_appendix}

\clearpage
\bibliographystyle{plain} 
\bibliography{main}

\end{document}

%% file: 01_introduction.tex
 \section{Introduction} \label{intro}
\emph{Patrolling} can be defined as travelling regularly through a given area so that emerging events of interest, e.g., traces of intrusion, can be identified as early as possible. Autonomous vehicles are highly suited to carry out patrolling tasks. They are designed for continuous and repetitive work, and their usage in hazardous areas can lead to reduced safety risks (for humans). The benefits of using autonomous vehicles have been demonstrated in different patrolling use cases such as environment monitoring \cite{env1,env2} disaster management \cite{disaster1,disaster2}, and security management \cite{security1, security2}. 

Solutions to the patrolling problem aim to minimize the time between visits to any location in a given area \cite{p1-first}. However, finding a solution is non-trivial. Chevaleyre~\cite{p2-theoretical} demonstrates that the patrolling problem is highly related to the well-known Travelling Salesman Problem (TSP) and thus is NP-hard. In addition, it is often required to use multiple patrolling agents to patrol large areas, which can give rise to multi-agent coordination problems. Existing literature mainly focuses on finding approximate solutions for environments comprising agents that can be accurately modelled. Examples include solutions based on finding short cycles that cover a given area/map~\cite{cyclic1, p2-theoretical, freq-Area}, graph partitioning~\cite{p2-theoretical, partition1, partition2, partition3}, heuristic searches \cite{p3-hs}, auction-based algorithms \cite{p4-auction-review2004} and Bayesian inference \cite{dsf1, dsf2, dsf3}. However, it is typically infeasible to model real environments accurately due to their complexity and uncertainty. In addition, the use of autonomous vehicles as patrolling agents has associated constraints and uncertainties,  e.g., limited battery capacity, potential failure problems, and uncertainties in movements. This further increases the complexity of the problem. Model-free solutions to multi-agent patrolling (MAP) problems that consider unknown environmental dynamics and constraints of the patrolling agents are thus highly desirable.

Multi-Agent Reinforcement Learning (MARL) is a suitable technique for solving model-free MAP problems. It enables agents to learn how to make sequential decisions to achieve specific goals in model-free environments. Deep MARL is based on a combination of deep learning and MARL. It has demonstrated its suitability to solve complex multi-agent systems problems, such as outperforming professional players in multi-player real-time strategy games \cite{starcraft, dota}. The success of deep MARL for the automation of multi-agent systems has led to recent interest in applying deep MARL techniques to solving broader patrolling problems \cite{rl_recent_1, rl_recent_2}. Despite the recent interest, the majority of existing MAP solutions based on reinforcement learning~\cite{p6-RL, rl_recent_1, rl_recent_2, rl_recent_3} are naively based on ideal patrolling agents, and hence are not suited to autonomous vehicle patrolling scenarios where agents typically have constraints.

This work proposes a deep MARL-based solution for MAP problems that incorporates various environmental dynamics and associated constraints of patrolling agents. Specifically, the agents patrol in a model-free and unknown dynamic environment, and automatically return to battery charging stations to recharge themselves as/when required. In addition, the patrolling system can tolerate agent failures and allow the introduction of supplementary agents. We consider a typical MAP problem setting \cite{p4-auction-review2004, dsf1, rl_recent_1}, where agents are fully cooperative with each other when patrolling a given area. The detailed contributions of this paper are as follows:

\begin{itemize}

    \item we take into account two typical real-world limitations of autonomous patrolling agents: battery limitations and failure problems. In addition, we consider environmental dynamics and uncertainties affecting the agents' initial deployment, movement, battery charging time, and battery discharging rate;

    \item we introduce a multi-objective reward function that evaluates the performance of the agents' patrolling and battery recharging strategies. A modified multi-agent Proximal Policy Optimisation algorithm \cite{ppo} is proposed to train the agents;
    
    \item we present a distributed homogeneous multi-agent architecture, where agents execute an identical policy locally based on their own environmental observations and information shared by other agents. This is used to develop a fault-tolerant patrolling system.
        
\end{itemize}
The performance of the proposed solution is explored through simulation experiments from three perspectives: patrolling performance, the efficiency of battery recharging strategies, the agents' adaptability to failure, and the agents' adaptability to environment dynamics and uncertainties.

The rest of this paper is organised as follows. Section 2 reviews work on existing MAP solutions. Section 3 illustrates the modelling of the MAP problem as a MARL problem. Section 4 presents the learning algorithm. Section 5 introduces the experiments and discusses the results. Finally, Section 6 concludes the paper and discusses potential areas of future work.

%% file: 02_related_work.tex
\section{Related Work} \label{related_work}

MAP problems can be largely divided into three categories \cite{recent_trend, dsf2}: \emph{adversarial patrolling}, where agents attempt to detect intruders in an area; \emph{area patrolling}, where agents monitor an area for diverse purposes, e.g., information collection; and \emph{perimeter patrolling}, which is a special type of area patrolling where agents monitor the edges of a patrol map. This work focuses on area patrolling problems, where agents fully cooperate with each other.

Machado et al.~\cite{p1-first} formalise the definition of the MAP problem as multiple agents continuously traversing a graph $G(V,E)$ with the aim of minimising the idleness of every vertex. Here, the term \emph{idleness} refers to the time between two visits to the same vertex. The authors compare the performance of several approaches categorised by the agents' basic type (cognitive or reactive), the agents' observability, the communication method, and the coordination strategy. The authors propose two coordination strategies: i) a Conscientious Reactive (CR) strategy, where agents patrol neighbour vertices with the highest degree of idleness, and ii) a Conscientious Cognitive (CC) strategy, where agents follow the shortest path to patrol to the vertex with the highest degree of idleness. A centralised coordinator is used to ensure no agents target the same vertex. Both methods have been used as baseline strategies for performance comparison in related literature \cite{p2-theoretical, p3-hs, p4-auction-review2004, dsf1, rl_recent_1}. Chevaleyre~\cite{p2-theoretical} demonstrates that the MAP problem is closely related to the \textit{Travelling Salesman Problem (TSP)} and thus is NP-hard. The author proposes two graph theory-based strategies: i) a cycle-based strategy where agents follow a cyclic path that connects all vertices, and ii) a graph partition-based strategy where each agent is responsible for patrolling a section of the graph. These two approaches have been further explored by many subsequent works \cite{cyclic1, freq-Area, ACO1, ACO2, ACO3}. The aforementioned works are largely theoretical and assume ideal agents and environments with no constraints, uncertainties, or dynamics. As a result, their applicability to patrolling in real-world environments with autonomous vehicles is limited.

Several approaches have been proposed to address the limitations of patrolling agents. For example, Portugal et al.~\cite{dsf1, dsf2, dsf3} proposed a fault-tolerant, distributed, and scalable solution based on Bayesian decision rules. Poulet et al.~\cite{auction2, auction3} introduce a fault-tolerant auction-based patrolling strategy that introduces entry/exit mechanisms for agents that are entering/leaving the patrolling system. The energy constraints of patrolling agents, although important, have only been studied by a limited number of researchers \cite{recent_trend}. Sipahioglu et al.~\cite{b_path_planning} proposed an approach based on the Ulusoy partitioning algorithm that modelled the MAP problem with battery constraints based on a capacitated arc routing problem. Jensen et al.~\cite{b-hot-swap} proposed a hot-swap recharging strategy that swapped low-battery agents with fully charged ones. Sugiyama et al.~\cite{b-HRL} used hierarchical reinforcement learning to train a coordinator that decided how agents should trade off patrolling time and battery recharging time, and a heuristic path-finding algorithm is used to compute routes for agents back to the charging station. Basilico and Nicola~\cite{recent_trend} provided a comprehensive survey of MAP-related literature.

However, the aforementioned works typically rely on the construction of models to accurately represent the characteristics of agents and/or the environment. Portugal et al.~\cite{dsf1, dsf2, dsf3}'s Bayesian inference solution relies on the knowledge of the consequence of agents' actions, e.g. the probability of agents reaching a vertex and the gain on performance when an action is taken. However, in a dynamic environment, the consequence of agents' actions is typically unpredictable. Poulet et al.~\cite{auction2, auction3}, Sugiyama et al.~\cite{b-HRL} and many other strategies \cite{p1-first, p3-hs} required prior knowledge of the shortest path from agent locations. However, the length of the shortest path between two locations is typically non-deterministic in real-world scenarios, as the landscape or wind speed and direction can affect the travel time and distance. Even if it is feasible to acquire knowledge of the environment a priori, there is no guarantee that the environmental conditions will remain the same at the time of the actual patrolling. 

On the other hand, multi-agent reinforcement learning (MARL) can solve multi-agent optimization problems without a detailed and accurate model of the environment. This approach allows agents to learn the dynamics of the environment, coordination strategies, and other sub-tasks without prior knowledge of the system. Santana et al.~\cite{p6-RL} proposed an RL-based patrolling strategy based on tabular-based Q-Learning to train agents. This was shown to outperform CC and CR strategies \cite{p4-auction-review2004}. However, tabular-based RL algorithms are predominantly suited to small state space problems. For large or infinite state spaces, one alternative is deep MARL, which uses deep neural networks to abstract information related to the agent states. Jana et al.~\cite{rl_recent_1} trained agents using the Individual Deep Q-Learning algorithm \cite{IQL}, whilst Luis et al.~\cite{rl_recent_2} did so using a centralised learning approach and carried out real-world  patrolling experiments to show how deep MARL could be applied to real-world MAP problems. 

Although existing RL-based MAP approaches have achieved good performance, their assumption of ideal agents limits their applicability to real patrolling problems. For example, the centralised patrolling system proposed by Luis et al. ~\cite{rl_recent_2}, where a central coordinator assigns actions to each agent during patrolling, cannot tolerate the failure of the central coordinator or network partitions between agents and the coordinator. Regarding the individual learner approach~\cite{IQL}, the number of agent failure scenarios grows exponentially with respect to the number of patrolling agents, consequently requiring a vast number of training samples. Most importantly, battery constraints and automatic recharging requirements of the agents have not been studied in the existing RL-based approaches.

In this work, we address this gap by taking into account two typical patrolling agents' limitations: energy constraints and failures, alongside the assumption of a dynamic patrolling environment.

%-- with unlimited energy and no failure issues -- limits their application to real-world patrolling problems. In this work, we extend the real-world MAP problem settings from existing RL-based works, where the environmental dynamics are unknown and cannot be modelled, while we take into account two typical patrolling agents' limitations: energy constraints and failure constraints.

%% file: 03_preliminary.tex
\section{Preliminaries} \label{preliminary}

\subsection{Single-Agent Reinforcement Learning \cite{RL}}
A single-agent RL problem is typically modelled as a Markov Decision Process (MDP) given as a set $\langle S, A, P, R, \gamma \rangle$ where:
\begin{itemize}
    \item $S$ is the set of all possible states of the environment;
    \item $A$ is the agent's action space which defines the agents' available actions;
    \item $P$ is the state transition function of the environment, which represents the probability of the environment transitioning from one state to another given the agent's action;
    \item $R$ is the reward function, and
    \item $\gamma$ is a discount factor, where $\gamma \in [0, 1) $.
\end{itemize}

The interaction between an agent and the environment in an RL problem is modelled as follows. In each time step, the agent chooses an action $a$ from its action space $A$ according to its policy $\pi(a \mid s)$ and based on the current environment state $s$. The environment will then transition to another state $s'$ according to the state transition function $P(s' \mid s,a)$, and the agent will receive a reward $r$ according to the reward function $R(s,a, s')$. A sequence of interactions ($\langle s,a,r \rangle$) between the agent and the environment, beginning at the initial state of the environment and ending at the terminal state of the environment, is called an episode. The goal of an agent is to find the optimal policy $\pi^*$ that maximises the expected cumulative reward given as $\mathcal{R}$ (Eq.~\ref{reward})
\begin{equation} \label{reward}
    \mathcal{R}_t=\sum_{i=0}^{\infty}\gamma^ir_{t+i+1},\gamma \in [0,1)
\end{equation}
where $\gamma$ is the discount factor used to control the influence of future rewards. The larger the value of $\gamma$, the more the agent pursues future rewards. The discount factor is also used to converge the cumulative reward to a finite value when dealing with episodes of potentially infinite length.

There are two main types of RL: model-based RL and model-free RL. Model-based RL assumes a known environment state transition function so that the agent can predict the consequence of a given action. Model-free RL does not know the environment model, and the agent learns the consequences of actions only from its past experience when interacting with the environment. In this work, we focus on model-free RL, as environment models are generally inaccessible in many real-world problems.

Model-free RL algorithms can be classified into three groups: value-based, policy-based, and actor-critic-based. Value-based learning algorithms, such as Q-Learning, use the value iteration method to approach the optimal action-value function $Q^*$, and hence obtain the maximum possible expected future return of each available action when the agent is in a given state. An optimal action $a$ in a given state $s$ is given as $\underset{a}{argmax}(Q^*(s,a))$. Policy-based algorithms, such as the REINFORCE \cite{reinforce} algorithm, use a policy gradient method to directly update the policy in the direction that results in higher expected returns. Actor-critic algorithms, such as Proximal Policy Optimisation (PPO), combine the previous two types of algorithms to learn a value function (the \textit{critic}), which evaluates the performance of the policy (the \textit{actor}). 

\subsection{Multiagent Reinforcement Learning}
Multi-agent Reinforcement Learning (MARL) focuses on decision-making problems where multiple agents are involved. In this work, we are mainly focused on multi-agent patrolling problems with fully-cooperative and fully-observable settings, where agents have the same goal -- minimise the idleness of every vertex in the graph, and can fully observe the states of the environment and know other agents' information through communication. Fully-cooperative and fully-observable MARL problems can be modelled by a  Multi-Agent MDP (MMDP) given as a set $\langle \mathbb{D}, \mathbf{S}, \mathbb{A}, \mathbf{T}, \mathbf{R}, \gamma\rangle$ where

\begin{itemize}
    \item $\mathbb{D}$ is the set of agents in the environment;
    \item $\mathbf{S}$ is the set of states of the environment;
    \item $\mathbb{A}$ is the joint actions of all agents, given as $\mathbb{A} = \underset{i}{\times} \mathbf{A}_i$ where $\mathbf{A}_i$ is the actions of agent $i$;
    \item $\mathbf{T}$ is an environment state transition function;
    \item $\mathbf{R}$ is a reward function;
    \item $\gamma$ is the discount factor and $\gamma \in [0, 1]$;
\end{itemize}

% A special subset of the Dec-POMDP, called Dec-MDP, models situations where the combination of all agent observations is the true state of the environment, that is, $\mathbb{\Omega} = \mathbf{S}$ and $\mathbf{O}$ becomes deterministic. In addition, if the agents can communicate and form a global observation, Dec-MDP becomes the Multi-Agent MDP (MMDP, also called Markov Game) -- a set $\langle \mathbb{D}, \mathbf{S}, \mathbb{A}, \mathbf{T}, \mathbf{R}, \gamma\rangle$ where the components are identical to those in the Dec-POMDP set. Generally speaking, the MAP problem can be simulated by Dec-MDP or MMDP, since it is normally assumed that the agents can communicate. However, due to message loss or delay, the agents may not be able to form a global observation of the environment.

 There are three commonly used MARL training frameworks: centralised training centralised execution (CTCE), centralised training decentralised execution (CTDE), and decentralised training decentralised execution (DTDE). In CTCE, a meta-agent, trained by a single-agent reinforcement learning algorithm, learns a joint policy directly and assigns actions to agents. In CTDE frameworks, such as the multi-agent PPO algorithm \cite{multi-agent_ppo}, agents are trained based on global observations, while during the run time (execution), they act based on local observations and information shared between them. Finally, DTDE algorithms, such as the Individual Q Learning algorithm \cite{IQL}, each agent learns its own policy and executes it based on local observations or information shared by other agents.

% In traditional CTDE and DTDE frameworks, each agent will learn their own policies. However, when failure is introduced, a system with $n$ number of agents has $2^n$ possible failure scenarios, and when scale-up is required, it is unclear what policy additional agents should execute.
% However, as we previously discussed, a centralised coordinator may not always be available during the runtime. In addition, since the output size of many neural network architectures (CNN, DNN) is normally fixed, the meta-agent cannot control extra agents introduced to the system. 

\subsection{Multi-agent Proximal Policy Optimisation}

The Multi-agent Proximal Policy Optimisation (MAPPO) algorithm \cite{multi-agent_ppo, ppo} follows the CTDE framework, where it uses a critic network $V_{\theta_1}(s)$ with parameter set $\theta_1$ to evaluate the value of a given state, and uses actor networks $\pi_{\theta_{2,i}}(s)$ with parameter set $\theta_{2,i}$ to approximate agent $i$'s policy. The critic network is used only during training.

Eq.~\ref{ppo_a} and Eq.~\ref{ppo_c} are the loss functions of the actor and critic networks for the MAPPO algorithm \cite{ppo, multi-agent_ppo}. Here, $T$ is the maximum horizon; $\theta$ is the neural network parameters; $r_{t,i}$ is agent $i$'s reward at time $t$; $a_{t,i}$ is agent $i$'s action at time $t$; $s_t$ is the observed environment state at time $t$; $\epsilon$ is a clipping constant which limits the value of $ratio_t(\theta)$ between $1-\epsilon$ and $1+\epsilon$; $\gamma$ is the discount factor;and $A_{t,i}$ is the \emph{Generalized Advantage Estimation} of agent $i$, which estimates whether an action leads to a better or worse long-term reward. $\lambda$ is the discount factor used when calculating $A_{t,i}$; $V^{targ}_t(s_t)$ is the actual value of the state, which is approximated based on the cumulative reward at time $t$: $\sum_{t'=t}^T\gamma^{t'-t} r_{t'}$; and $MSE$ is the \emph{mean square error}.

\begin{equation} \label{ppo_a}
    \begin{split}
        &L^{CLIP}_{t,i}(\theta) = \mathbb{E}_{t,i}[min(ratio_{t,i}(\theta)A_{t,i}, clip(ratio_{t,i}(\theta), 1-\epsilon, 1+\epsilon)A_{t,i})] \\
        &\text{where  } ratio_{t,i}(\theta) = \frac{\pi_{\theta_{new}}(a_{t,i} \mid s_{t})}{\pi_{\theta_{old}}(a_{t,i} \mid s_{t})} \\
        &\text{and  } A_{t,i} = \delta_{t,i}+(\gamma\lambda)\delta_{t+1}+...+(\gamma\lambda)^{T-t+1}\delta_{T-1} \\
        &\text{where  } \delta_{t,i} = r_{t,i} + \gamma V_\theta(s_{t+1}) - V_\theta(s_{t}) \\
    \end{split}
\end{equation}

\begin{equation}\label{ppo_c}
L^{VF}_t(\theta) = MSE(V_\theta(s_t)-V^{targ}_t(s_t))  
\end{equation}

The critic network is updated towards the direction where the estimate of the value of the state $s_t$ $V_\theta(s_t)$ is closer to the actual value of the state $V^{targ}_t(s_t)$, which is approximated based on the cumulative reward at time $t$: $\sum_{t'=t}^T\gamma^{t'-t} r_{t'}$. The actor network is updated towards the direction that increases the probability of agents choosing the action that can lead to a better long-term reward. To train the agent with the MAPPO algorithm, a set $\langle s_t, p(a_t \mid \theta), r_t, V_\theta(s_t), a_t \rangle$ needs to be collected at each training step to form a trajectory, which will be used to compute the value of loss functions.

% \begin{equation}\label{ppo_ae}
% \begin{split}
% &\overline{L^{CLIP}_t(\theta)} =\\
% &\frac{\sum_{t \in K} min(ratio_t(\theta)A_t, clip(ratio_t(\theta), 1-\epsilon, 1+\epsilon)A_t)}{ \mid K \mid }
% \end{split}
% \end{equation}

%% file: 04_modelling.tex
\section{Problem Modelling} \label{sec:modelling}
Following the work of Machado et al.~\cite{p1-first}, we consider the MAP problem as a set of agents continuously traversing a graph $G(V,E)$ with the goal of minimising the idleness of every vertex. For modelling purposes, we define the following:
\begin{itemize}
    \item $Idle(v_t)$ -- the idleness of vertex $v$ at time $t$;
    \item $Idle(G_t)$ -- the idleness of graph $G(V,E)$ at time $t$, which is the average of the idleness of all vertices in the graph $G(V,E)$;
    \item $AVG^h(G)$ -- the average of $Idle(G_t)$ over $h$ steps of a patrolling scenario, and
    \item $MAX^h(G)$ -- the maximum $Idle(v_t)$ that occurs during $h$ steps of a given patrolling scenario.
\end{itemize}

The patrolling problem can be modelled as an optimisation problem. $AVG^h(G)$ and $MAX^h(G)$ are two commonly used optimisation criteria \cite{p1-first, p2-theoretical, p3-hs, rl_recent_1, dsf1}, where $AVG^h(G)$ measures the agents' average patrolling performance, and $MAX^h(G)$ measures the agents' worst patrolling performance. Many existing works only consider one of the two optimisation criteria \cite{p2-theoretical, p3-hs, dsf1, rl_recent_1, rl_recent_2}. However, Santana et al.~\cite{p6-RL} have shown that only optimising the average idleness during patrolling may leave a vertex unvisited for a long time, and only minimising the maximum idleness during patrolling may result in high average idleness. Therefore, both optimisation criteria are taken into account in this work. However, the environmental dynamics may have an unpredictable effect on agents' worst patrolling performance, e.g. pushing agents away from their targets and resulting in a sudden degradation in agents' patrolling performance. 

Specifically, this work aims to minimise $AVG^h(G)$ and $\overline{MAX^h(G)}$, where $\overline{MAX^h(G)}$ is the average of the largest $Idle(v_t)$ measured at each step over $h$ steps of a given patrolling scenario. $\overline{MAX^h(G)}$  is used instead of $MAX^h(G)$ as it provides a more accurate measure of the worst patrolling performance in dynamic environments where agents are subject to unpredictable behaviour such as being pushed away from target locations.

\subsection{Environment and Agent Modelling}

\begin{figure}[!hbt]

    \begin{subfigure}[t]{.27\linewidth}
        \centering
        \includegraphics[width=\linewidth]{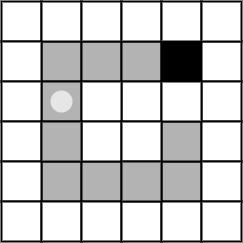}
        \caption{An example grid map}
        \label{fig:grid_map}
    \end{subfigure}\hfill
    %%%%%
    \begin{subfigure}[t]{.27\linewidth}
        \centering
        \includegraphics[width=\linewidth]{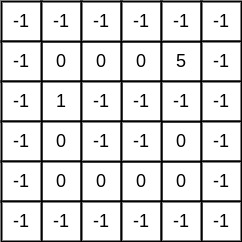}
        \caption{Matrix expression of (a)}
        \label{fig:grid_map_matrix}
    \end{subfigure}\hfill
    %%%%%
    \begin{subfigure}[t]{.27\linewidth}
        \centering
        \includegraphics[width=\linewidth]{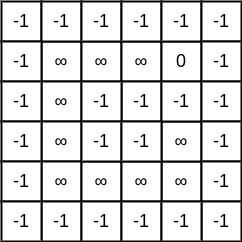}
        \caption{Idleness matrix of (a) at timestep 0}
        \label{fig:idleness_matrix}
    \end{subfigure}
    
    \caption{An example grid map (a), its corresponding matrix expression (b), and its idleness matrix at timestep 0 (c).}
\end{figure}

In this work, we consider modern autonomous vehicles, such as drones, with the ability to communicate and access their location information (GPS), to be patrolling agents. Hence, we consider the graph that is patrolled to be based on a geometric map, which can be discretised into a grid and represented as a matrix. Fig.~\ref{fig:grid_map} shows an example of a grid map, where grey blocks are vertices (i.e., locations that agents can visit), white blocks are obstacles (i.e., locations that agents cannot occupy), the black block is the battery charging station, and the circle represents a given agent. Fig.~\ref{fig:grid_map_matrix} is the matrix representation of the grid map in Fig.~\ref{fig:grid_map}, where $0$ represents the vertices, $-1$ represents obstacles, and $5$ represents battery charging stations. In addition, We consider the patrolling graph on a large scale so that each vertex is large enough for multiple agents to visit at the same time.

The location of the agent is represented by the matrix index of the vertex occupied by the agent, which simulates the GPS coordinates. For example, in Fig.~\ref{fig:grid_map}, if the top-left block has index $(0,0)$, then the bottom-right block has index $(5,5)$, and the location of the agent is $(1,2)$. The agent can also mark its location on the matrix based on its coordinate location information. In Fig.~\ref{fig:grid_map_matrix}, $1$ represents the agent's location. In addition, as agents are patrolling grid maps, it is assumed that the action space of the agent is one of $\langle \allowbreak Up,\allowbreak Down, \allowbreak Left, \allowbreak Right\rangle$. 

Similarly, the idleness of the grid map can also be represented by a matrix. It is assumed that the idleness of vertices is $+\infty$ at the beginning of a patrolling scenario since no vertices have been visited yet. If an agent visits a vertex, the idleness of that vertex will be set to $0$. The idleness of obstacles remains constant at $-1$, and the idleness of the battery charging station remains constant at $0$. Fig.~\ref{fig:idleness_matrix} shows the idleness matrix for the grid map in Fig.~\ref{fig:grid_map} at time $0$. As we assume that agents can communicate, they can share their current location with each other, and mark the idleness of the vertices occupied by other agents to $0$ to form a global observation of the idleness matrix of the graph.

In this work, we take into account battery-related constraints that affect the ability of agents to continuously patrol a given map. Specifically, we assume that agents are battery-operated and that the battery has limited charge and drains at an unknown rate. As a result, agents must recharge their batteries before they run out to enable continuous patrolling. The time required to recharge a modern autonomous vehicle, e.g., a UAV, can be four times longer than the actual flight time \cite{UAV}. Therefore, waiting until a vehicle is fully charged before deploying it to continue a given patrol is impractical. Instead, in this work, we assume a hot-swap battery recharging scheme \cite{b-hot-swap}, i.e., when a vehicle goes to the battery charging station to recharge, a charged vehicle will be deployed to replace it. We assume the battery hot-swap process takes a non-negligible amount of time since pre-flight preparation needs to be carried out before an agent can be deployed. In addition, for safety purposes and to account for features present in modern autonomous vehicles (e.g., a drone may force-land when its battery is lower than a preset threshold \cite{dji}), we define $b_{l}$, a tunable parameter defining the amount of battery that should remain when an agent visits the battery charging station.

We also take into account agent failures. In particular, we consider catastrophic failures that cause failed agents to become non-operational -- they can no longer patrol (move), observe the environment, or communicate. If an agent's shared information is not received by the rest of the active agents after one step of patrolling, the agent is considered failed. The remaining patrolling agents are expected to keep patrolling when one or multiple agents are failed. In addition, supplementary agents may be introduced to either replace the failed agents or to increase the patrolling performance. We assume that the supplementary agents will be deployed from the battery charging stations.

%agents cease to function --> no longer patrol, observer, communicate 
%remaining agents should be able to continue to patrol

Alongside, we consider a dynamic patrolling environment. We assume that the environmental dynamics affect an agent's movements and battery draining rate with an unknown probability $p^i_{dyn}$. Specifically, an agent's moving direction and speed may be changed by the dynamics, e.g., wind, resulting in a change to their moving direction, and their traveling time between vertices. As a result, agents may consume more/less battery, if the dynamics push them away/towards their targets.

% Overall, the observable information for an agent at time $t$ is given as a set $\langle \allowbreak G(V, E), \allowbreak Idle(G_t), \allowbreak B_t, \allowbreak Loc_t \rangle$, where $Idle(G_t)$ is the idleness matrix of the graph $G(V, E)$ at time $t$, $B_t$ is all agents' remaining battery information at time $t$, and $Loc_t$ is all agents location information at time $t$. 

\subsection{Environment State Transitions}

In a given patrolling scenario, agents interact with the environment in the following way:

\begin{enumerate}
  \item At the beginning of a patrolling scenario, all agents are randomly placed on vertices with a random battery remaining in the graph since we do not assume any initial deployment strategy.
  
  \item At the beginning of a step, each agent will first make observations, then communicate and synchronise their global observations, and finally choose an action and move to the corresponding vertex contemporaneously. However, the environmental dynamics may affect agents' movement. After every agent arrives at the vertices, the current step is considered terminated. An agent's battery will drop for a non-deterministic amount due to environmental dynamics.
  
  \item At the end of each step, the idleness of all vertices will increase by the time that agents spend completing the step. And at the end of each step, the idleness of the vertices occupied by agents will be set to $0$.
  
  \item If the agent lands on the battery charging station intentionally, i.e., not caused by environmental dynamics, the hot-swap procedure will start immediately. Otherwise, the agent will not be charged or replaced. After the hot-swap procedure is completed, a charged agent will be deployed. 

\end{enumerate}

We summarise the parameters of the environment and agents and their notations in Table~\ref{tab:parameters}.

\renewcommand{\arraystretch}{1.5}
\begin{table}[htb]
    \centering
    \begin{tabular}{ll} 
    \hline
    \textbf{Parameter}  & \textbf{Description} \\
    \hline
    $b_{max}$           & The maximum battery capacity of the agent. \\
    \hline
    $b^i_{init}$        & Agent $i$'s battery remaining at deployment. \\
    \hline
    $b^i_{swap}$        & \makecell[l]{The time to replace agent $i$ with a charged agent when agent $i$ wants to recharge.} \\
    \hline
    $p^i_{dyn}$         & \makecell[l]{The non-deterministic probability that agent $i$'s movement will be affected due \\ to environmental dynamics and uncertainties.} \\
    % $\Delta^k_{idle}$   & \makecell[l]{The time the agents spend completing a step \\ and the idleness increase in a step $k$.} \\
    % \hline
    % $Idle_T$            & The predefined idleness threshold.\\
    \hline
    \end{tabular}
    \caption{The parameters of the environment and agents.}
    \label{tab:parameters}
\end{table}

%% file: 05_methods.tex
 \section{Methods} \label{sec:methods}

% In this work, agents can communicate and share their local observations, and the number of agents will vary due to failed agents or the introduction of supplementary ones. Therefore, a distribution of Multi-agent Markov Decision Processes (MMDP) $\tau_t \sim p(\tau)$ is proposed to model the MAP problem, where $\tau_t = \langle \mathbb{D}_t, \mathbf{S}, \mathbb{A}_t, \mathbf{T}, \mathbf{R}, \gamma\rangle$ is an MMDP where: $\mathbb{D}_t$ is the set of agents at time step $t$; and $\mathbb{A}_t$ is the agents' corresponding joint action space at time step $t$. The other components are identical to those in Dec-POMDP. 

In this section, we detail the architecture of the patrolling system, the reward function, and the deep MARL learning algorithm proposed to train the patrolling system.

%%%%%%%%%%%% System Architecture %%%%%%%%%%%%

\subsection{System Architecture}

This works aims to design a patrolling system that tolerates agent failures. This makes a MARL approach based on the centralised training centralised execution architecture (CTCE) unsuitable, as the failure of the central coordinator would result in a catastrophic failure of the system. With respect to the decentralised training and decentralised (DTDE) architecture, as each agent is distinct, in a MAS with $n$ agents, the number of possible combinations of agents that are online (i.e., patrolling) at any given point in time is $2^n-1$, without considering the case in which all agents are offline. In order to train a strategy that tolerates changes to the number of patrolling agents, the agents would have to learn to cooperate with each other in $2^n-1$ distinct cases, requiring a vast amount of training samples to cover all patrolling situations. In addition, scaling up becomes challenging under such an architecture, as determining the policy that newly added agents will follow is not straightforward. One approach is to let additional agents execute random policies from existing agents, but this may lead to inconsistent performance. 

Therefore, in this work, we propose a distributed homogeneous MAS architecture that follows the centralised training decentralised execution (CTDE) architecture, where agents have an \emph{identical} policy based on local and shared information. In this architecture, with $n$ agents, the number of possible combinations of online agents at any time is $n-1$, requiring fewer samples to train a robust (patrolling) policy. In addition, since agents are identical, the experience (learning) can be shared between them to improve training efficiency. Moreover, agents are potentially trained to have a strategy that can cooperate with a varying number of agents. Therefore, when scaling up the system, it is expected that the additional agents can cooperate well with the existing agents in the system.

%%%%%%%%%%%% The reward function %%%%%%%%%%%%

\subsection{The Reward Function}

The reward function reflects the performance of the agent's policy on solving a given task. In this work, the reward function evaluates agents from two perspectives: i) the performance of the agents' patrolling strategy, and ii) the performance of the agents' battery charging strategy. Therefore, the MAP problem we consider can be treated as a multi-objective reinforcement learning problem, with two reward functions, $R_p$ and $R_b$, evaluating the agents from each of the two perspectives. The linear scalarization method is used to combine the two reward functions into a single function with a summation function, as shown in Eq.~\ref{fn_highlevelreward}.

\begin{equation}
\label{fn_highlevelreward}
r = R_p + R_b
\end{equation}

\paragraph{Patrolling performance ($R_p$)} 
With respect to the agents' patrolling goal, two optimisation criteria are considered: $AVG^h(G)$ and $\overline{MAX^h(G)}$. Similarly, the linear scalarization method is used; we combine the two criteria with the summation function since we consider both criteria to be equally important. The optimisation goal is then to minimise $AVG^h(G) + \overline{MAX^h(G)}$, which is equivalent to maximising $-AVG^h(G)-\overline{MAX^h(G)}$. Therefore, the $R_p$ can be defined as Eq.~\ref{rp1}, given a time step $t$.

\begin{equation} \label{rp1}
R'_p(t) = -Idle(G_t)-max(Idle(v_t))
\end{equation}

where $max(Idle(v_t))$ is the maximum idleness among the vertices in graph G at time step $t$. The corresponding cumulative reward function with finite horizon $h$ is shown in Eq.~\ref{c_rp1}. The reinforcement learning algorithm will find a strategy that maximises Eq.~\ref{c_rp1}, hence, solving the defined problem.

\begin{equation} \label{c_rp1}
\begin{split}
\mathcal{R}_p &= \sum_{t=0}^{h}\gamma^t \cdot (-Idle(G_t)-max(Idle(v_t))) \\
                &= -\sum_{t=0}^{h}\gamma^t \cdot Idle(G_t) -\sum_{t=0}^{h}\gamma^t \cdot max(Idle(v_t)) \approx h \cdot (-AVG^h(G)-\overline{MAX^h(G)})
\end{split}
\end{equation}

\begin{figure}[!hbt]
    \centering
    \includegraphics[width=0.44\textwidth]{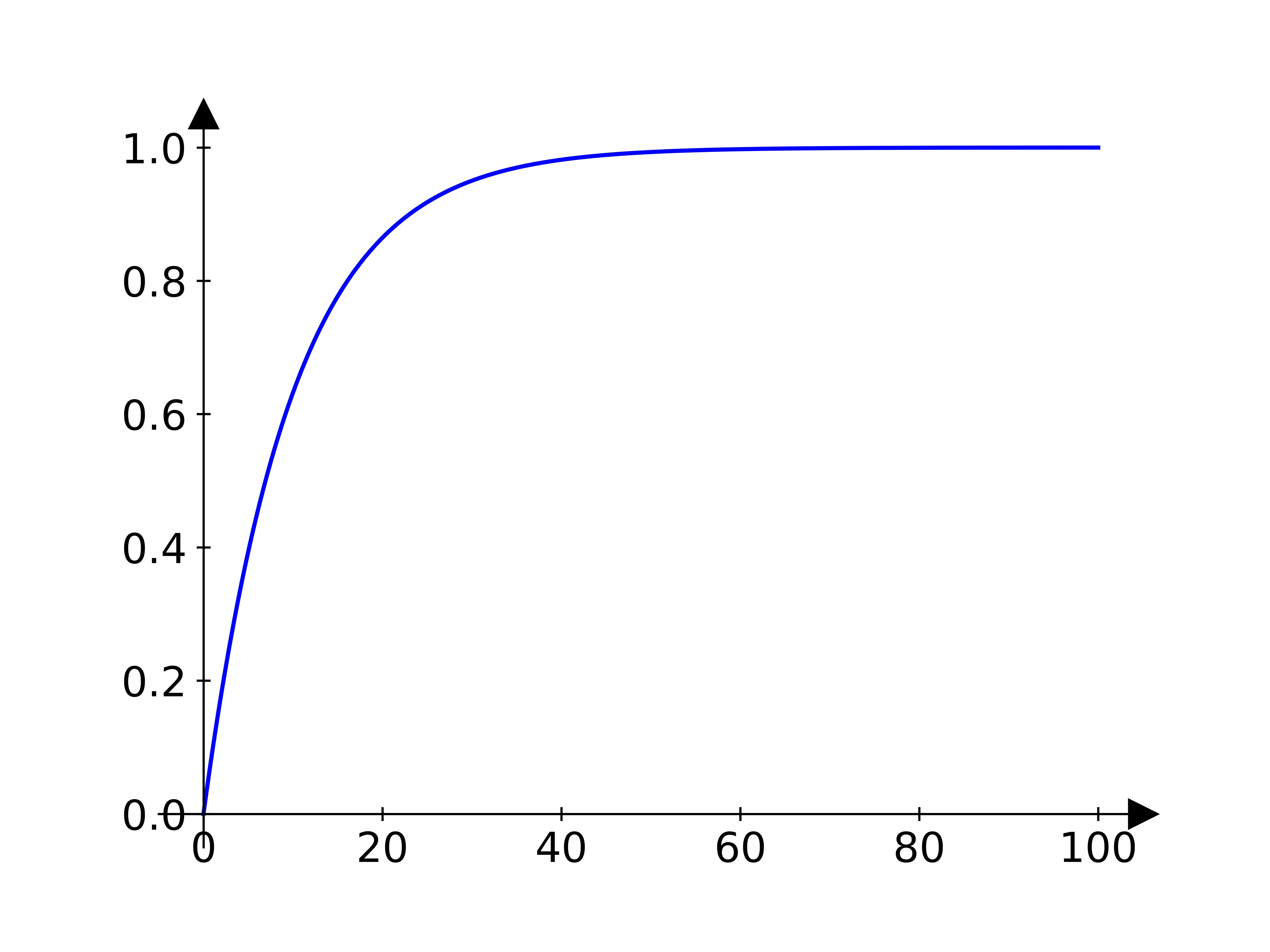}
    \caption{Function plot of $f(i) = -e^{-\frac{i}{c_{norm}}} + 1$, $i \in [0,100]$}
    \label{fig:function_plot_norm}
\end{figure}

The idleness of a vertex ranges from $0$ to $+\infty$. We normalise the idleness of vertices between $0$ and $1$ with the function $f(i) = -e^{-\frac{i}{c_{norm}}} + 1$, where $i$ is the idleness and $c_{norm}$ is a constant. The function is depicted in Fig.~\ref{fig:function_plot_norm} when $c_{norm} = 10$.

In addition, $R_p$ is scaled to a positive reward with value between $0$ and $1$, as shown in Eq.~\ref{rp2}
\begin{equation} \label{rp2}
    R''_p(t) = \frac{2-Idle(G_t)-max(Idle(v_t))}{2}
\end{equation}

Although agents run an identical policy, the agents may have differences in their patrolling performance. For example, the stochasticity in an agent's policy, if using RL algorithms such as Q-Learning or PPO, allows the agent to explore different patrolling behaviours. Therefore, some agents may contribute more to the patrolling task than other agents. However, the reward function in Eq.~\ref{rp2} assigns all agents the same reward, where agents with poor performance will receive the same reward as good-performing agents, thus, introducing instability to training and degrading the training efficiency. To address this issue, we introduce the \emph{difference reward} \cite{difference_reward} to $R''_p$. The difference reward indicates the \emph{contribution} of each agent, which is shown in Eq.~\ref{r_diff}, where $D_{k}$ is agent $k$'s difference reward, $G$ is the reward that takes into account all agents' actions, and $G_{-k}$ is the reward that takes into account all agents' actions except agent $k$'s action. $G$, in our cases, is $R''_p$, which is shown in Eq.~\ref{rp2}. And $G_{-k}$ is the $R''_p$ calculated by assuming agent $k$ is not moving. Each agent wants to maximise the cumulative sum of $D_{k}$ so that they contribute as much as possible to the patrolling in a scenario.

\begin{equation} \label{r_diff}
    D_{k}(t) = G(t) - G_{-k}(t)
\end{equation}

Since the agent's goal is to maximise both $R''_p$ and $D_{k}$, we use the linear scalarization method, specifically the summation, to combine the two reward functions. The reward function of $R_p$ for an agent $k$ is shown in Eq.~\ref{rp3}, where $c_{R_p}$ and $c_{R_d}$ are the corresponding scaling factors for $R''_p(t)$ and the difference reward respectively.

\begin{equation} \label{rp3}
    R_p(k,t) = R''_p(t) \cdot c_{R_p} + D_{k}(t) \cdot c_{R_d}
\end{equation}

% In general, the higher the $\frac{w_{R_d}}{w_{R_p}}$ is, the more the agents will focus on increasing their own contribution. However, competition may occur between the agents, e.g., all agents are planning to visit a vertex with the highest idleness, leaving other vertices not being planned to visit. Therefore, $w_{p}$ and $w_{d}$ need to be fine-tuned during training.

\paragraph{Battery usage ($R_b$)} 

The recharging strategy of an agent is evaluated based on two aspects: i) the ability of an agent to recharge its battery before it runs out as measured by $R_{b1}$, and ii) the ability of an agent to recharge with a battery remaining of $b_l$ (the predefined battery level that the agents should remain when recharging) as measured by $R_{b2}$. Again, the linear scalarization method is used, and $R_b$ is defined as:
\begin{equation} \label{rwb}
R_b = R_{b1} + R_{b2}
\end{equation}

An agent $k$ failing to recharge its battery will be penalised by a constant value, as shown in Eq.~\ref{rwb1}.
\begin{equation} \label{rwb1}
R_{b1}(k) = 
    \begin{cases}
      -c_b & \text{if agent $k$ runs out of battery} \\
      0 & \text{Otherwise}
    \end{cases}       
\end{equation}

% As the patrolling environment has dynamics and the battery drainage of agents has uncertainties, the more battery the agent uses before it decides to recharge, the higher the chance that it will run out of battery on the way to the battery charging station. On the other hand, agents remaining more when recharging reduces the chance of running out of battery on the way to recharge, but the agents need to recharge more often, leading to a worst patrolling performance. In addition, 

In $ R_{b2}$, if the agent recharges with a battery remaining higher/lower than $b_{l}$, it will receive a penalty ($R_{b2-recharge}$) with respect to its remaining battery level when it recharges. In addition, if the agent patrols with a battery lower than $b_{l}$, it will receive a penalty ($R_{b2-patrol}$) in each step with respect to its remaining battery level. The linear scalarization method, specifically the summation, is used to combine $R_{b2-recharge}$ and $R_{b2-patrol}$. Eq.~\ref{rwb2} shows $R_{b2}$ for agent $k$, where $b_k$ is agent $k$'s remaining battery level, and $c_{recharge}$ and $c_{patrol}$ are the corresponding scaling factors for $R_{b2-recharge}$ and $R_{b2-patrol}$.

\begin{equation} \label{rwb2}
\begin{split}
    &R_{b2}(k) = c_{recharge} \cdot R_{b2-recharge}(k) + c_{patrol} \cdot R_{b2-patrol}(k) \\
    &R_{b2-recharge}(k) = 
    \begin{cases}
      -\frac{1}{b_l}b_k + 1 & 0 \leq b_k \leq b_l \> \textbf{and} \> *\\
      \frac{1}{1-b_l}(b_k-b_l) & b_l < b_k \leq 1 \> \textbf{and} \> *\\
      0 & \text{Otherwise}
    \end{cases} \\
    &R_{b2-patrol}(k) = b_l - b_k,\> \textbf{when } b_k\in [0,b_l] \\
    &* - \textit{if agent $k$ lands on the battery charging station}
\end{split}
\end{equation}

\begin{figure}[!hbt]
    \centering
    \includegraphics[width=0.44\textwidth]{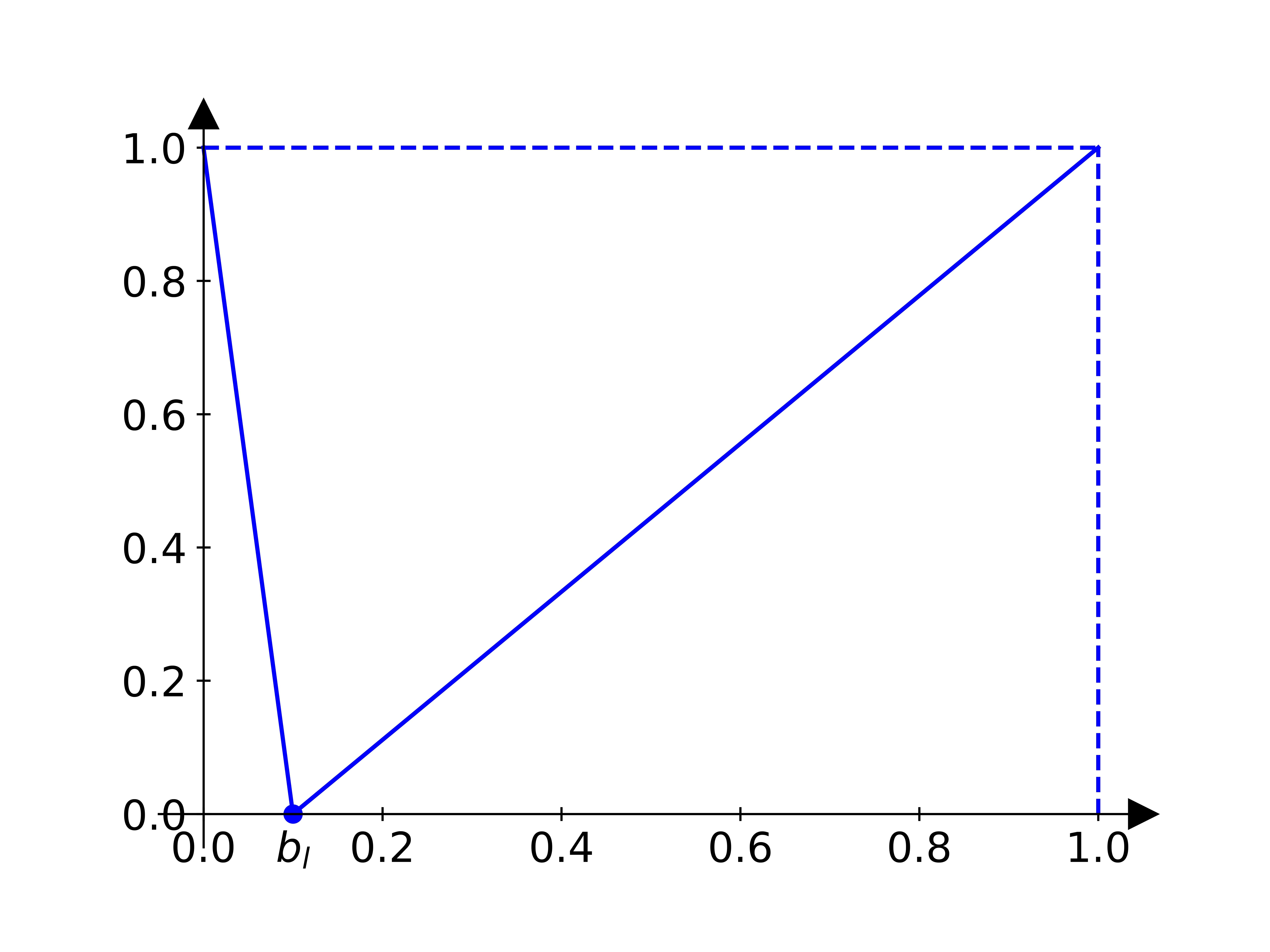}
    \caption{Function plot of $R_{b2-recharge}$, $x \in [0,1]$}
    \label{fig:function_plot_rb2}
\end{figure}

Fig.~\ref{fig:function_plot_rb2} plots $R_{b2-recharge}$, $x \in [0,1]$. From the graph, we can see that when an agent recharges with $b_l$ of remaining battery, the agent's recharging strategy is considered optimal and, therefore, will not receive a penalty. Otherwise, a penalty will be assigned to the agent.

% Regarding the value selection for $w_{recharge}$, $c_b$, and $w_{patrol}$, the larger the $w_{recharge} \cdot R_{b2-recharge}$ is, the more the agents are encouraged to learn to recharge efficiently. However, it may discourage the agents from recharging at the early stage of learning, as the efficient recharging strategy is not discovered, and penalties are received for charging. On the other hand, the larger the $c_b$ and $w_{patrol} \cdot R_{b2-patrol}(k)$ are, the agents are encouraged more to recharge, as large penalties are received for running out of battery. Therefore, these parameters need to be fine-tuned to achieve optimal training performance for each patrolling scenario.  

%%%%%%%%%%%% The Learning Algorithm %%%%%%%%%%%%

\subsection{The Learning Algorithm}

To train our strategy, we use the Multi-agent Proximal Policy Optimization (MAPPO) algorithm \cite{ppo, multi-agent_ppo} which aligns with the CTDE architecture of the proposed homogeneous MAS architecture.

As we consider agents' failure, recharging needs, and the introduction of supplementary agents, the number of patrolling agents varies over time, resulting in an observation space with inconsistent size, which cannot be easily handled by many deep neural network architectures, such as Convolutional Neural Networks (CNNs) and Artificial Neural Network (ANNs). 

An agent's observable information consists of a set $\langle \allowbreak G(V, E), \allowbreak Idle(G_t), \allowbreak B_t, \allowbreak Loc_t\rangle$, where $G(V, E)$ is the patrolling graph, $Idle(G_t)$ is the idleness matrix, $B_t$ are agents' battery information, and $Loc_t$ are agents' location information. $G$ and $Idle(G_t)$ have fixed sizes, while $Loc_t$ and $B_t$ varies with respect to the number of active agents. 

The agents' critic network evaluates the value of the current state of patrolling based on global information. During training, we assume a maximum number of patrolling agents so the size of all agents' battery information and location information is bounded. Therefore, the observation of an agent $i$'s critic network is the set $\langle \allowbreak G(V, E), \allowbreak Idle(G_t), \allowbreak B_t, \allowbreak Loc_t\rangle$. If fewer agents than the assumed maximum are patrolling in a scenario, the vacant space in $B_t$ will be filled with $1$, and the vacant space in $Loc_t$ will be filled with the location of battery charging stations.

The agent's actor network needs adaption to the introduction of supplementary agents during the run time, which will increase the observation size. Hence, $Loc_t$ and $B_t$ need to be transformed into fixed-size observations. 

Regarding agents' location $Loc_t$, an agent will set the idleness of the vertices occupied by other agents to $0$ in the idleness matrix after establishing the agents' locations. Since only the vertices, except the battery charging station, occupied by agents will have an idleness value of $0$, all other agents' location information can be approximately derived from the idleness matrix. However, agents cannot differentiate whether a vertex is occupied by one or multiple agents. But the deep neural network can generalise the patrolling situations, so we assume the agents can still find the near-optimal patrolling strategy. 

Regarding the battery information, we designed that the agent's actor network can only observe its own battery information. As the reward function $R_b$ only depends on the agents' own battery information, agents can still learn to recharge near optimally without global battery information.

In addition, to efficiently train the agents to avoid obstacles, the invalid action masking method is used \cite{starcraft}. The method inputs the validity of actions as an observation to the agent's actor network, and renormalises the agents' probability distribution of the action space so that only valid actions can be chosen. In this paper, a set with four elements is used to represent the validity of each of the four actions, with entries in the set being $0$ if the action is invalid or $1$ otherwise. For example, in the scenario depicted in Figure \ref{fig:grid_map}, the agent can only move \textit{Up} or \textit{Down}, resulting in the following action masking set: $\langle \allowbreak 1,\allowbreak 1, \allowbreak 0, \allowbreak 0\rangle$. If assume the agent's probability distribution over the action space is $\langle 0.4, 0.1, 0.3, 0.2 \rangle$, since actions "Left" and "Right" are invalid, the probability distribution will be renormalised to $\langle 0.8, 0.2, 0, 0 \rangle$. In real applications, the invalid action masking set can be acquired by obstacle detection hardware, e.g. radar.

Thus, an agent $i$'s actor network's observation is a set $\langle \allowbreak G(V, E), \allowbreak Idle(G_t), \allowbreak b_{it}, \allowbreak Loc_t, \allowbreak ACT_{it}\rangle$, where $b_{it}$ is agent $i$'s battery information at time $t$, and $ACT_{it}$ is agent $i$'s invalid action masking set at time $t$. 

As a result, agents need to share their location information at each step of patrolling, and their location and battery information during training. If using a $32bit$ floating point number to represent an agent's latitude or longitude, $8$ bytes are used to represent an agent's location information. In addition, a $32bit$ floating point number can be used to represent an agent's battery remaining information. Therefore, $12$ bytes of information need to be exchanged between agents in each step during patrolling. With modern communication technology, such as WiFi, 4G, and 5G cellular networks, with $100Mbps$ transmission bandwidth, sending $12$ bytes would take approximately $9.6\cdot10^{-7}$ seconds in ideal conditions. Compared with the time of autonomous vehicles traveling between vertices, e.g. seconds in duration, the communication time between agents has a minor impact on the idleness increase, hence, is negligible. 

\begin{figure}[!hbt]
    \begin{subfigure}[b]{.48\linewidth}
        \centering
        \includegraphics[width=.8\linewidth]{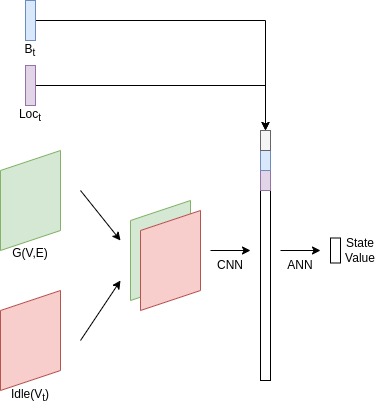}
        \caption{Actor Network}
        \label{fig:actor}
    \end{subfigure}\hfill
    %%%%%
    \begin{subfigure}[b]{.48\linewidth}
        \centering
        \includegraphics[width=.8\linewidth]{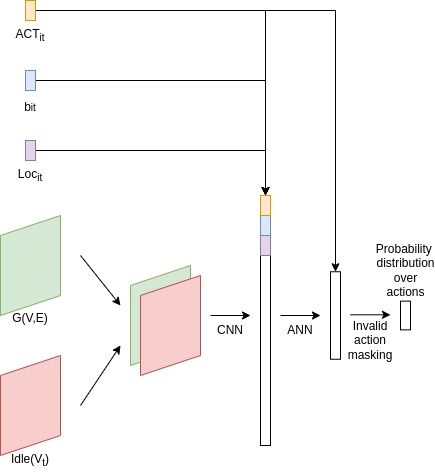}
        \caption{Critic Network}
        \label{fig:critic}
    \end{subfigure}%
    %%%%%
    \caption{Architecture of the Actor Network \ref{fig:actor} and the Critic Network \ref{fig:critic} in the MAPPO. The arrow represents the direction of the data flow.}
\end{figure}

We extend the neural architectures of the actor and critic networks from the previous deep MARL-based solution \cite{rl_recent_2} with some modifications. The networks' architectures are illustrated in Figure \ref{fig:actor} and \ref{fig:critic}. 

First, the grid map and the idleness matrix are concatenated and processed by a CNN. The output of the CNN is then flattened as a one-dimensional array. For clarity, the flattened CNN output of the critic network is named $O^c_{CNN}$, and the flattened CNN output of the actor network is named $O^a_{CNN}$. For the critic network, observations $Loc_t$, $B_t$ are included as part of $O^c_{CNN}$, which is the input to an ANN, producing as output an estimate of the value of the state at time step $t$ ($V_{\theta_1}(s_t)$). For the actor network, $Loc_{it}$, $b_{it}$ and $ACT_{it}$ are included as part of $O^a_{CNN}$, which is the input to an ANN, producing as out the probability distribution over the action space ($p_{\theta_2}(a \mid s_t)$); then, renormalised according to the invalid masking set to avoid agents choosing invalid actions. 

In our multi-agent patrolling scenario, agents receive different rewards in each step. This results in different agents calculating different values of $V^{targ}_t(s_t)$, which introduces instabilities for the critic network when learning the value of a state. Therefore, a scalarization function is used to map all agents' $V^{targ}_t(s_t)$ to a single value $V^{targ'}_t(s_t)$. Since all agents are using the same policy $\pi$, the performance of the policy $\pi$ can be approximated by the average performance of all agents, represented by the average of the cumulative rewards of all agents. Therefore, the average function is used as the scalarization function when calculating $V^{targ'}_t(s_t)$. The expression of $V^{targ'}_t(s_t)$ is shown in Eq.~\ref{new_vtar}:
\begin{equation} \label{new_vtar}
    \begin{split}
        & V^{targ'}_t(s_t) =  \frac{\sum_{d \in D}\sum_{t'=t}^T\gamma^{t'-t} r_{dt'}}{ \mid D \mid } \\
    \end{split}
\end{equation}
where $D$ is the set of agents, $r_{dt'}$ is agent $d$'s reward at time $t'$. The actor loss function remains the same.\\

% \subsubsection{Trajectory Collection}

In Section.~\ref{preliminary}, we discussed that training agents via the MAPPO algorithm require a trajectory containing $\langle s_t, p(a_t \mid  \theta), r_t, V_\theta(s_t), a_t \rangle$ to be collected in each time step of a training episode. As the homogeneous MAS architecture is used, agents will collect their trajectories locally during the training episode, and the trajectories will be combined together after the training episode is terminated. In addition, MAPPO allows collecting trajectories from $n$ parallelly running episodes to improve the sampling efficiency. This also supports training the fault tolerance of the agents by running the $i$th episode with $i$ patrolling agents. Consequently, the agents' policy is trained to patrol with $1$ to $n$ number of agents, thus, having fault tolerance.

When an agent is recharging, we consider it offline and will be replaced by a supplementary agent through battery hot-swapping. The deployed supplementary agent will continue the trajectory collection. However, the battery hot-swapping procedure takes time, which means that the trajectory is not collected when preparing the supplementary agent for deployment. This will lead to agents calculating different values of $V_t^{targ}(s_t)$ for the same state $s_t$, even when identical rewards are received.

For example, considering two agents, $a$ and $b$, run an episode for $10$ steps and are receiving identical rewards with the discount factor $\gamma = 1$. Agent $a$ recharges itself at timestamp $1$, and agent $b$ recharges itself at timestamp $3$, and battery hot-swapping takes $6$ timesteps. The $V_t^{targ}(s_0)$ calculated by agent $a$ is $\mathcal{R}_a(s_0) = r_0 + r_7 + r_8 + r_9$, and the $V_t^{targ}(s_0)$ calculated by agent $b$ is $\mathcal{R}_b(s_0) = r_0 + r_1 + r_2 + r_9$. And $\mathcal{R}_a(s_0) \neq \mathcal{R}_b(s_0)$. This introduces instability when training the critic network, which learns from $\mathcal{R}_a(s_0)$ and $\mathcal{R}_b(s_0)$ to approximate $V_t^{targ}(s_0)$.

However, the proposed homogeneous MAS architecture allows trajectory data to be shared between agents. Therefore, the missing data not collected during the deployment of a supplementary agent can be reconstructed based on the trajectories of other running agents. In the example above, the cumulative reward of $s_0$ calculated by both agents $a$ and $b$ now becomes $r_0 + r_1 + r_2 + r_7 + r_8 + r_9$. During time $3$ to time $6$, no agents are patrolling, so no data can be shared between the agents. By using this method, all agents will calculate the same $V_t^{targ}(s_t)$, which reduces the instability. 

The pseudocode of the learning algorithm is shown in Algorithm \ref{alg:mappo}.

\begin{algorithm}[htb]
\caption{Homogeneous Multi-agent Proximal Policy Optimisation Algorithm}\label{alg:mappo}
\begin{algorithmic}[1]

\State $i \Leftarrow 0$
\State $\text{Create an agent $(agent_0)$}$
\State $\text{Initialise Critic Network with random parameters}$
\State $\text{Initialise $agent_0$'s Actor Network with random parameters}$
\State $\text{Clone the $agent_0$ into $N$ agents}$

\While{$i \leq episodes$}
    \State $\text{Reset the environment}$
    \State $s \Leftarrow \text{starting state $s_0$}$
    \State $step \Leftarrow 0$
    
    \While{$step \leq horizon$ \textbf{or} $\text{episode is not terminated}$}
        
        \State $joint\_action \Leftarrow [\:]$\;
        \State \begin{@empty}Each agent $i$ makes an observation and shares its location with other agents to form $O^c_t$ and $O^a_{it}$\end{@empty}
        \State $V_{\theta_1}(s_t) \Leftarrow critic\_network(O^c_t)$

        \For{$\text{each \textbf{online} agent $a$}$}
            \If{$\text{a is not charging}$}
                \State $joint\_action.append(a.actor(O^a_{it}))$
            \EndIf
        \EndFor
        \State $\text{update environment}$
        \State $s \Leftarrow new\_state$
        \State \begin{@empty}each agent store $\langle O^c_t, O^a_{it}, p(a_{it} \mid \theta), r_{t}(i), V_{\theta_1}(s_t), a_{it} \rangle$\end{@empty}
        \State \begin{@empty}replace hot-swapping agents' trajectory with an active agent's trajectory \end{@empty}

        \If{$\text{any agent's battery expires}$ \textbf{or} $\text{idleness expires}$}
            \State $\text{terminate episode}$
        \Else
            \State $\text{$step \Leftarrow step + 1$}$
        \EndIf

        \State $\text{all agents except $agent_0$ copy trajectory to $agent_0$}$
        \State $\text{update $agent_0$'s actor network}$
        \State $\text{update critic network based on $agent_0$'s trajectory}$
        
        \For{$\textbf{each agent $a$ except $agent_0$}$}
            \State $a.actor \Leftarrow agent_0.actor$
        \EndFor
        
    \EndWhile
\EndWhile

\end{algorithmic}
\end{algorithm}

\clearpage

%% file: 06_evaluation.tex
\section{Performance evaluation} \label{sec:exp}

\begin{figure}[!hbt]
    \begin{subfigure}[b]{.2\linewidth}
        \centering
        \includegraphics[width=\linewidth]{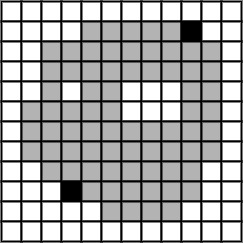}
        \caption{Map A}
        \label{fig:street}
    \end{subfigure}\hfill
    %%%%%
    \begin{subfigure}[b]{.2\linewidth}
        \centering
        \includegraphics[width=\linewidth]{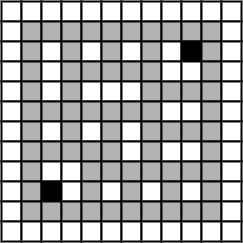}
        \caption{Map B}
        \label{fig:field}
        \end{subfigure}\hfill
    %%%%%
    \begin{subfigure}[b]{.2\linewidth}
        \centering
        \includegraphics[width=\linewidth]{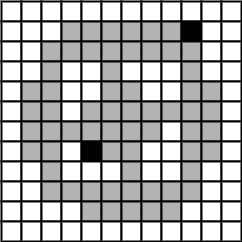}
        \caption{Map C}
        \label{fig:island}
        \end{subfigure}\hfill
    %%%%%
    \begin{subfigure}[b]{.2\linewidth}
        \centering
        \includegraphics[width=\linewidth]{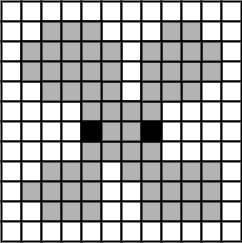}
        \caption{Map D}
        \label{fig:room}
        \end{subfigure}\hfill
    \caption{Four patrolling maps.}
    \label{fig:maps}
\end{figure}

To provide a general performance analysis of the proposed deep MARL-based approach, we train models on $4$ patrolling maps (Fig.~\ref{fig:maps}) with $3$ different $b_l$ settings -- $0.1$, $0.15$ and $0.2$. We use the following convention to name different models being trained: \emph{A/B/C/D-$b_l$}, For example, \emph{A-$0.1$} refers to the model that is trained on map A with $b_l$ set to $0.1$. The maps will be loaded onto agents prior to patrolling. During training, a maximum of $5$ patrolling agents are assumed.

The performance of the proposed deep MARL-based strategy is evaluated from the following perspectives: 

\begin{itemize}

    \item \textbf{Battery recharging performance} -- evaluated based on the battery failure rate (when an agent runs out of battery) and on the remaining battery level when agents recharge. The lower the battery failure rate and the closer the agent's remaining battery is to $b_l$ when recharging, the better the strategy's battery recharging performance.

    \item \textbf{Patrolling performance} -- evaluated based on the performance criteria $AVG^h(G)$ and $\overline{MAX^h(G)}$ (as defined in Section~\ref{sec:modelling}). The lower the values of the performance criteria, the better the strategy's patrolling performance;

    \item \textbf{Fault tolerance and the ability to cooperate with supplementary agents} -- evaluated based on the agents' patrolling and battery recharging performance in scenarios in which agent failures occur, and when supplementary agents are introduced. 

    % \item \textbf{The ability to cooperate with supplementary agents} -- evaluated based on the agents' \emph{Patrolling performance} and \emph{Battery recharging performance} when supplementary agents are introduced to the system. The existing agents are expected to automatically cooperate with supplementary agents, which should result in an increase in the system's patrolling performance without a decrease in agents' battery charging performance. A model trained with a maximum $5$ number of agents will be applied to $6$ to $8$ number of agents for performance analysis.

\end{itemize}

As baseline strategies for performance comparison, we consider the individual learner approach and the Conscientious Reactive (CR) strategy. 

The individual learner approach is commonly used in previous works to train patrolling agents \cite{p6-RL, rl_recent_3, rl_recent_1} and outperforms the CR strategy and heuristic search-based strategy in a \emph{deterministic} patrolling environment with \emph{ideal} patrolling agents \cite{rl_recent_3}. For fairness, the PPO algorithm is used as the reinforcement learning algorithm in the individual learner approach, and the same neural network architectures and the same training hyperparameters will be used when training the proposed MARL-based model.

The CR strategy is a reactive strategy that does not rely on a precise model of the environment, such as the shortest path or distance between vertices. This makes the CR strategy less susceptible to environmental dynamics compared with other strategies, such as the Conscientious Cognitive, heuristic search-based, or auction algorithm-based strategies. For fairness, the agents with CR strategies are allowed to communicate each other's location and form a global observation of the idleness matrix. The original CR algorithm does not include a battery charging strategy. We extend the CR strategy to enable agents to follow the shortest path to the nearest battery charging station when their battery level reaches a critical point. If the agent's moving direction is affected by the environment dynamics, the shortest path to the nearest battery charging station will be recomputed based on the agent's current location.

In this work, we configure the parameters of the environment and agents based on the settings of real-world experiments from previous works \cite{rl_recent_2} and the specifications of real-world UAVs, specifically, the DJI Matrice 300 RTK \cite{dji}, DJI's latest commercial drone at the time of writing. The Matrice 300 RTK has a maximum flight time of $55$ minutes under ideal conditions. In regards to the battery hot-swapping time, according to its manual, the UAV's batteries must be warmed up before installation. The battery heating process, using a DJI heater \cite{dji-bh}, takes $8$ to $13$ minutes and consumes $3\%$ to $7\%$ of the battery. 

We assume that the size of a grid on the graph is $50m$ by $50m$, and the time interval between two discrete timesteps is assumed to be $0.1$ minutes. The suggested horizontal flying speed of DJI Matrice 300 RTK should be no more than $10m/s$ with the radar obstacle avoidance feature, so we assume that in ideal conditions, it takes the UAV 1 discrete timestep to patrol to a neighbour vertex (with an $\approx8.33m/s$ flying speed).

As for the probability $p_{dyn}$ of an agent being affected by environmental dynamics, Luis et al. ~\cite{rl_recent_2} assume a constant $5\%$ for $p^i_{dyn}$d, which results in a successful application in a lake patrolling scenario. However, the effect of real-world environmental dynamics on agents is non-deterministic. Hence, in this work, we assume that in each step, $p^i_{dyn}$ has a random value between $0\%$ to $5\%$ for each agent. In addition, we assume the agent's battery can be drained up to $5\%$ faster than flying in ideal conditions, and the idleness increase in each step can be $-5\%$ to $5\%$ less/more than patrolling in ideal conditions.

The selected values of the parameters of agents and environment are shown in Table \ref{tab:parameters-1}.

\renewcommand{\arraystretch}{1.5}
\begin{table}[htb]
    \centering
    \begin{tabular}{ll} 
    \hline
    \textbf{Parameter} & \textbf{Value} \\
    \hline
    $b_{max}$        & $550$ time steps \\
    \hline
    $b^i_{swap}$     & $b^i_{swap} \sim [80, 150]$ time steps \\
    \hline
    $p^i_{dyn}$      & $p^i_{dyn} \sim [0, 0.05]$ per step and per agent\\
    \hline
    \end{tabular}
    \caption{The values of parameters of the environment and agents.}
    \label{tab:parameters-1}
\end{table}

\subsection{Model Training}

When training the proposed MARL approach, trajectories from $8$ episodes are collected to train the agents, where $4$ episodes are run with $2$, $3$, $4$ and $5$ agents and $4$ episodes are run with $1$ patrolling agent. More than $1$ episode of single-agent patrolling scenarios are included, aiming at improving the training efficiency, since the experience-sharing feature of the proposed homogeneous MAS architecture is inapplicable to a single agent. For a fair comparison, a similar amount of trajectories are used to train individual learner agents. Trajectories collected from $9$ episodes are used, where $5$ episodes are run with each individual patrolling agent, $3$ episodes are run with randomly selected $2$, $3$ and $4$ agents, and $1$ episode is run with $5$ patrolling agents.

The selected values of reward function parameters are shown in Table~\ref{tab:rw-parameters}. The details of the architecture of the agent's actor and critic networks are shown in Table~\ref{tab:Actor} and ~\ref{tab:Critic}, and the training hyperparameters of the MAPPO algorithm are shown in Table~\ref{tab:mappo}.

Each model was trained for $3,000$ steps, with each training episode having $5,000$ steps. If any agent failed to recharge, the training episode was terminated. 

\begin{table}[htb]
    \centering
    \begin{tabular}{ll} 
    \hline
    \textbf{Parameter} & \textbf{Value} \\
    \hline
    $c_{norm}$       & $150$\\
    \hline
    $c_b$            & $50$\\
    \hline
    $c_{R_p}$        & $0.5$\\
    \hline
    $c_{R_d}$        & $50$\\
    \hline
    $c_{recharge}$   & $2$ (Map A), $2$ (Map B), $1$ (Map C), $1$ (Map D)\\
    \hline
    $c_{patrol}$     & $10$ ($b_l$ = 0.2), $15$ ($b_l$ = 0.15), $25$ ($b_l$ = 0.1)\\
    \hline
    
    \end{tabular}
    \caption{The parameters of the reward functions.}
    \label{tab:rw-parameters}
\end{table}

%actor
\begin{table}[htb]
    \centering
    \begin{tabular}{ccc} 
    \hline
    \textbf{Layer} & \textbf{Parameter} & \textbf{Activation Function} \\
    \hline
    Conv &  \makecell{ic=2, oc=4, ks=(3,3), \\ s=1, p=0} & Tanh \\
    \hline
    Conv &  \makecell{ic=4, oc=8, ks=(3,3), \\ s=1, p=0} & Tanh \\
    \hline
    Dense & 519 $\times$ 512 & Tanh \\
    \hline
    Dense & 512 $\times$ 341 & Tanh \\
    \hline
    Dense & 341 $\times$ 227 & Tanh \\
    \hline
    Dense & 227 $\times$ 1 & None \\
    \hline
    \end{tabular}
    \caption{Neural network architecture of the agent's actor network. "Conv" -- convolutional layer, "Dense" -- dense layer, "ic" -- "input channel". "oc" -- "output channel". "ks" -- "kernel size". "s" -- "stride". "p" -- "padding". }
    \label{tab:Actor}
\end{table}

%critic
\begin{table}[htb]
    \centering
    \begin{tabular}{ccc}     
    \hline
    \textbf{Layer} & \textbf{Parameter} & \textbf{Activation Function} \\
    \hline
    Conv &  \makecell{ic=2, oc=4, ks=(3,3), \\ s=1, p=0} & Tanh \\
    \hline
    Conv &  \makecell{ic=4, oc=8, ks=(3,3), \\ s=1, p=0} & Tanh \\
    \hline
    Dense & 527 $\times$ 512 & Tanh \\
    \hline
    Dense & 512 $\times$ 341 & Tanh \\
    \hline
    Dense & 341 $\times$ 227 & Tanh \\
    \hline
    Dense & 227 $\times$ 1 & None \\
    \hline
    \end{tabular}
    \caption{Neural network architecture of the agent's critic network.}
    \label{tab:Critic}
\end{table}

% mappo
\begin{table}[htb]
    \centering
    \begin{tabular}{cc}
        \hline
        \textbf{Parameter} & \textbf{Parameter Value} \\
        \hline
        $\gamma$           & 0.95 \\
        \hline
        GAE $\lambda$      & 0.95 \\
        \hline
        Policy clip        & 0.15 \\
        \hline
        Number of Batches  & 50 \\
        \hline
        Epoch              & 3 \\
        \hline
        Entropy coefficient & \makecell[l]{Start from 0.04 and decrease 0.01 every 500 episodes, \\ with a minimum value of 0.005}  \\ 
        \hline
        Learning rate & \makecell[l]{Start from 2e-4 and decrease 5e-5 every 1000 episodes}  \\
        \hline
        \end{tabular}
    \caption{Values of MAPPO hyperparameters.}
    \label{tab:mappo}
\end{table}

\begin{figure}[htb]

    \begin{subfigure}{0.23\textwidth}
            \includegraphics[width=\linewidth]{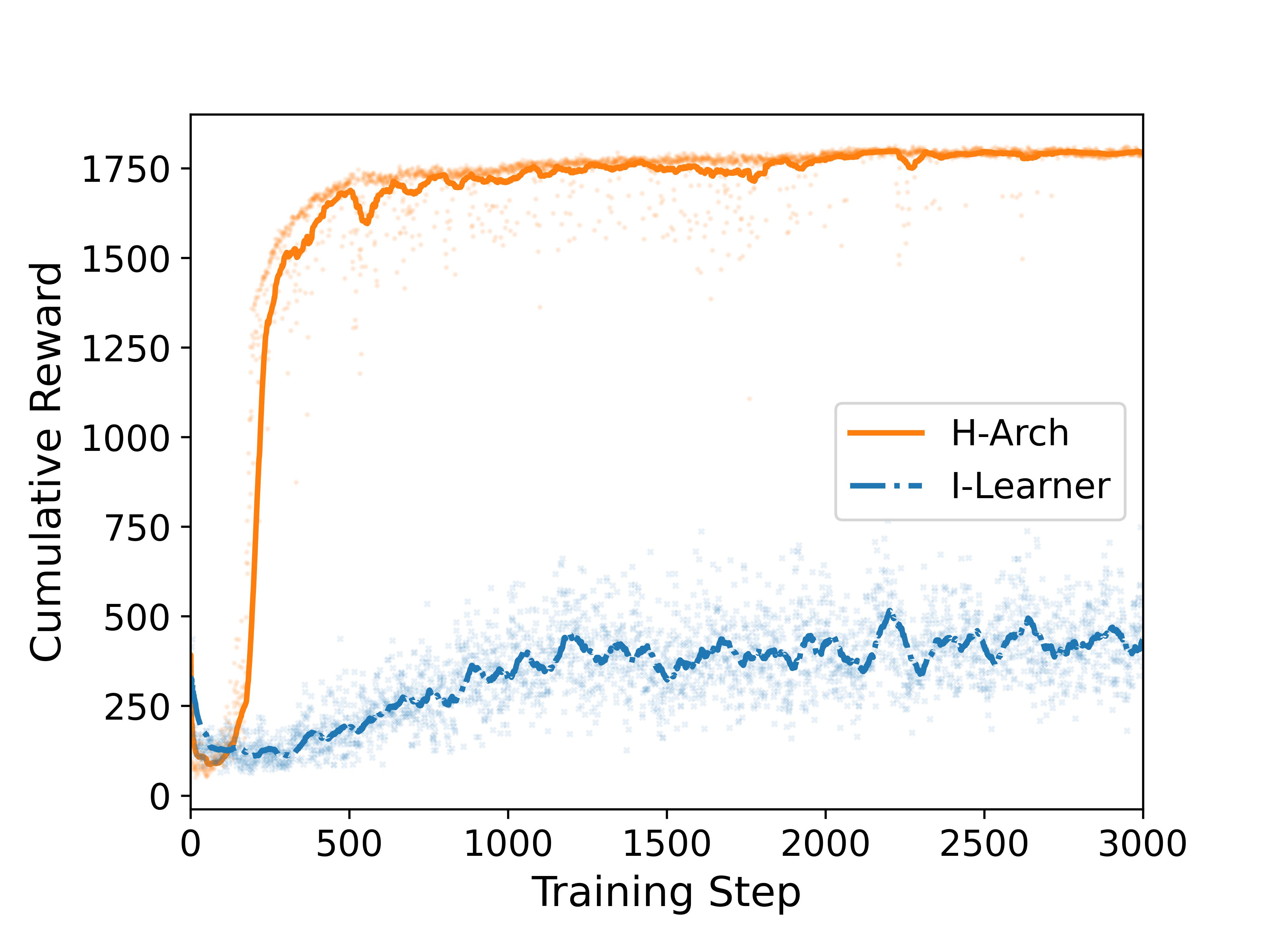}
            \caption{A-0.1 $\mathcal{R}$}
            \label{fig:6a}
        \end{subfigure} 
        \begin{subfigure}{0.23\textwidth}
            \includegraphics[width=\linewidth]{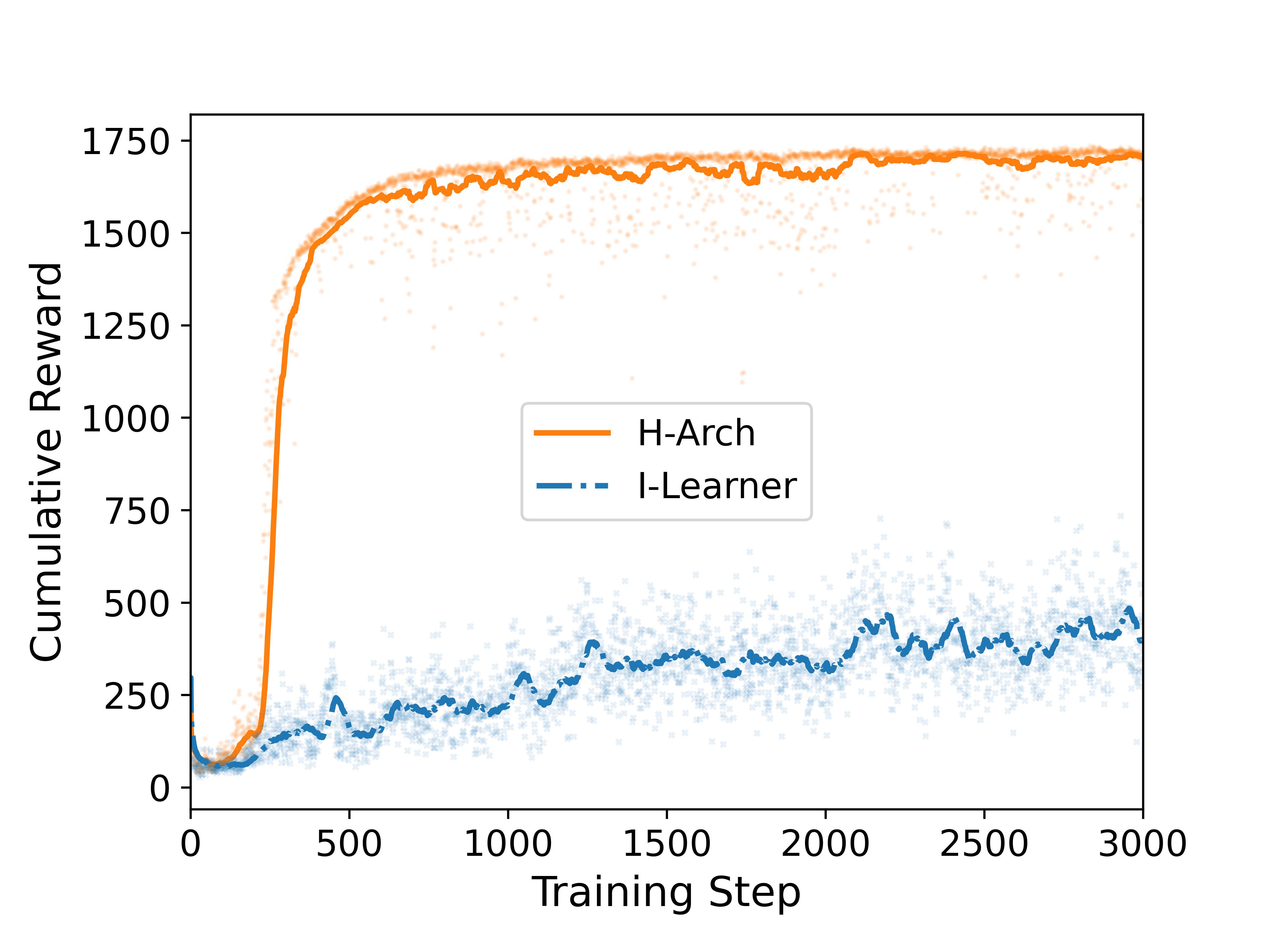}
            \caption{B-0.1 $\mathcal{R}$}
            \label{fig:6b}
        \end{subfigure} 
        \begin{subfigure}{0.23\textwidth}
            \includegraphics[width=\linewidth]{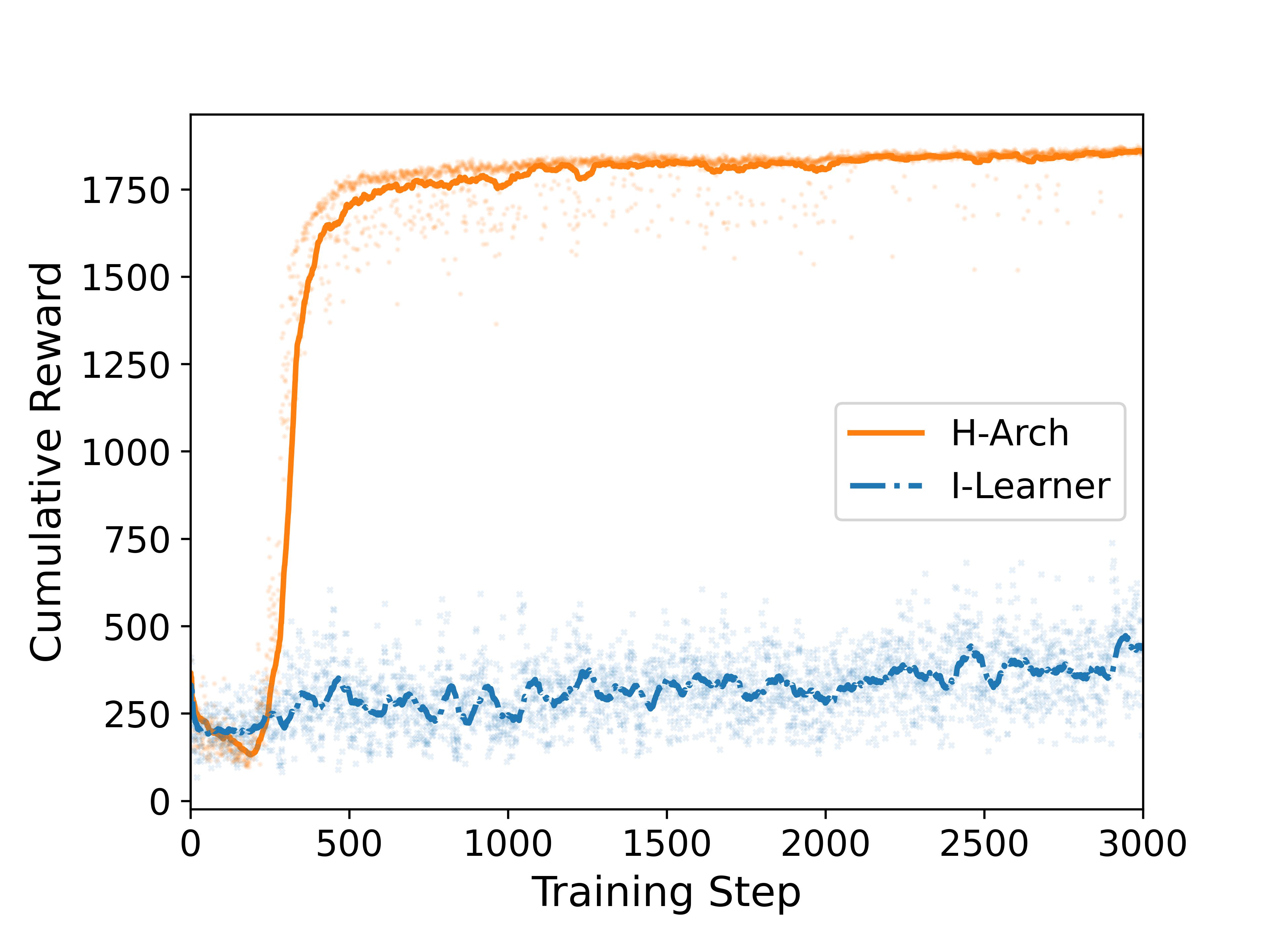}
            \caption{C-0.1 $\mathcal{R}$}
            \label{fig:6c}
        \end{subfigure} 
        \begin{subfigure}{0.23\textwidth}
            \includegraphics[width=\linewidth]{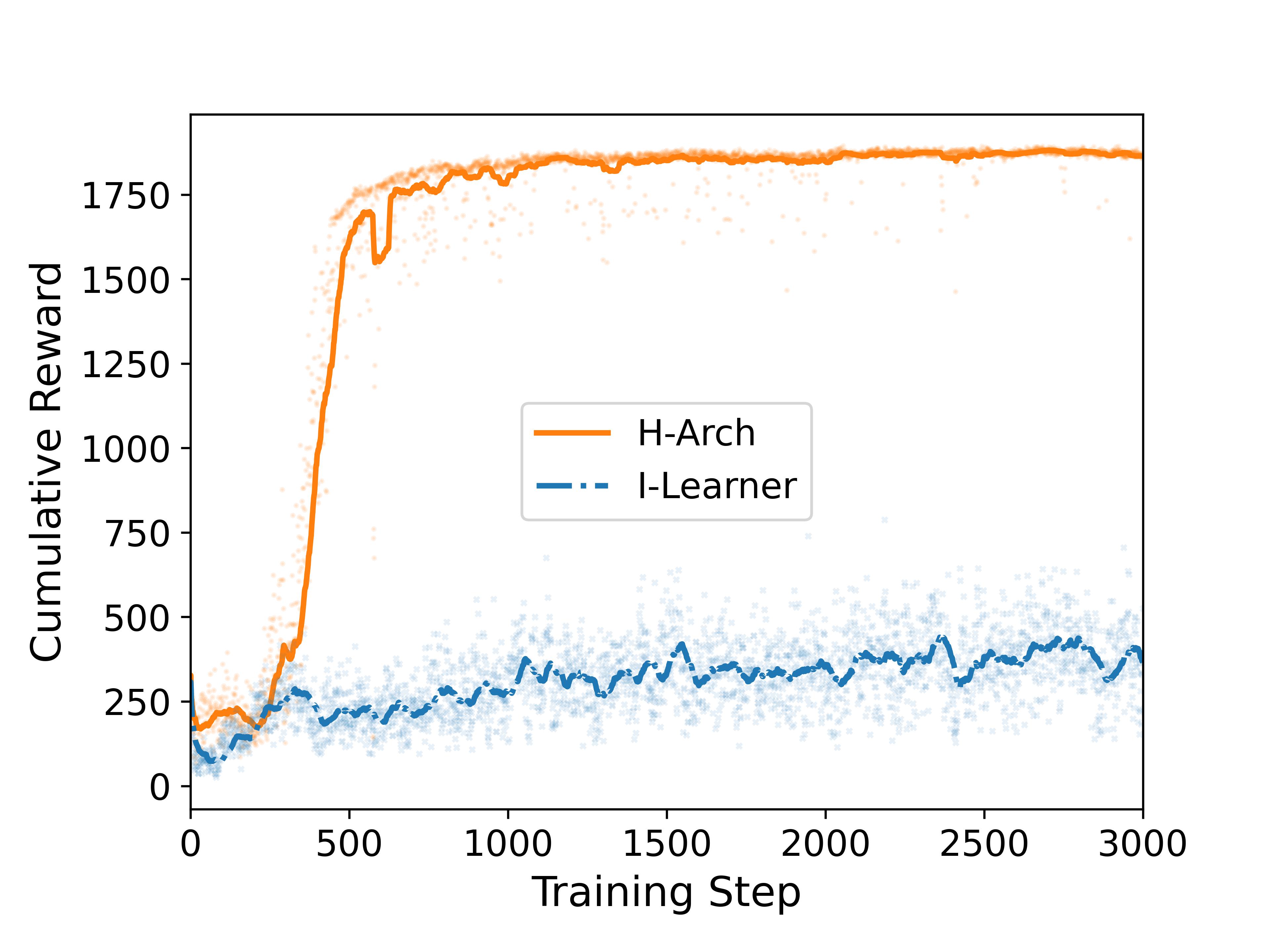}
            \caption{D-0.1 $\mathcal{R}$}
            \label{fig:6d}
        \end{subfigure} 
        
        \begin{subfigure}{0.23\textwidth}
            \includegraphics[width=\linewidth]{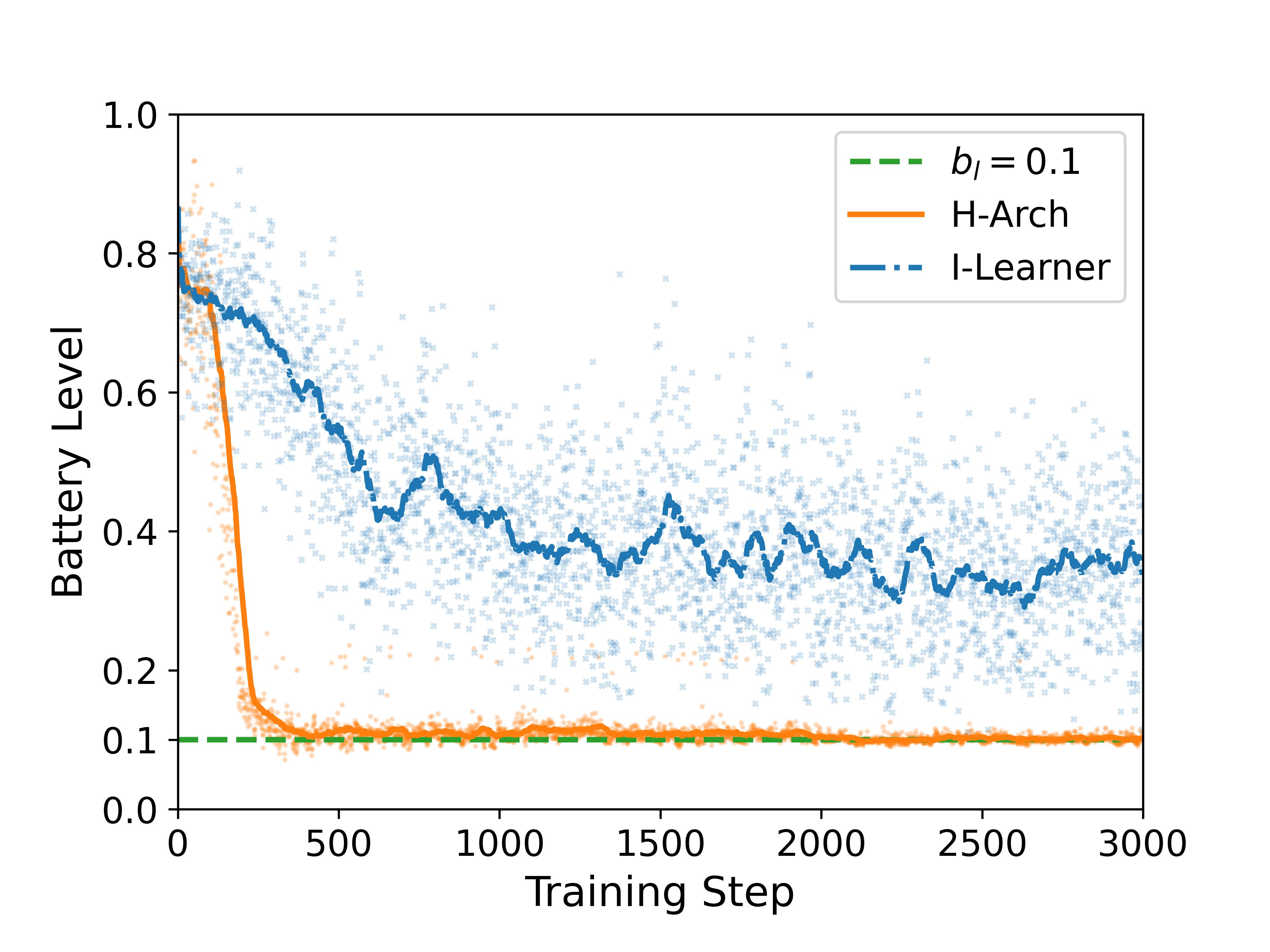}
            \caption{A-0.1 $\mathcal{B}$}
            \label{fig:6e}
        \end{subfigure} 
        \begin{subfigure}{0.23\textwidth}
            \includegraphics[width=\linewidth]{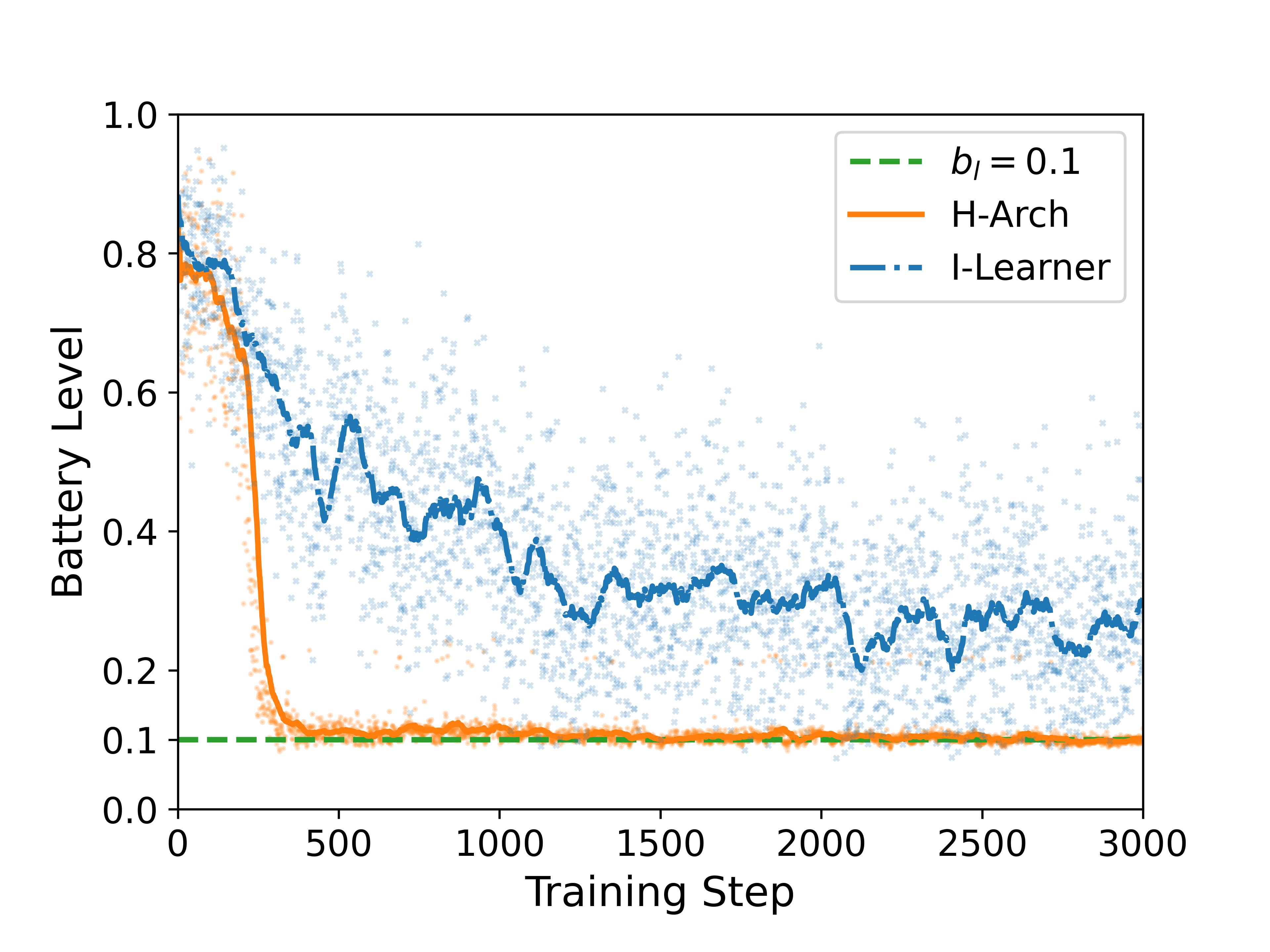}
            \caption{B-0.1 $\mathcal{B}$}
            \label{fig:6f}
        \end{subfigure} 
        \begin{subfigure}{0.23\textwidth}
            \includegraphics[width=\linewidth]{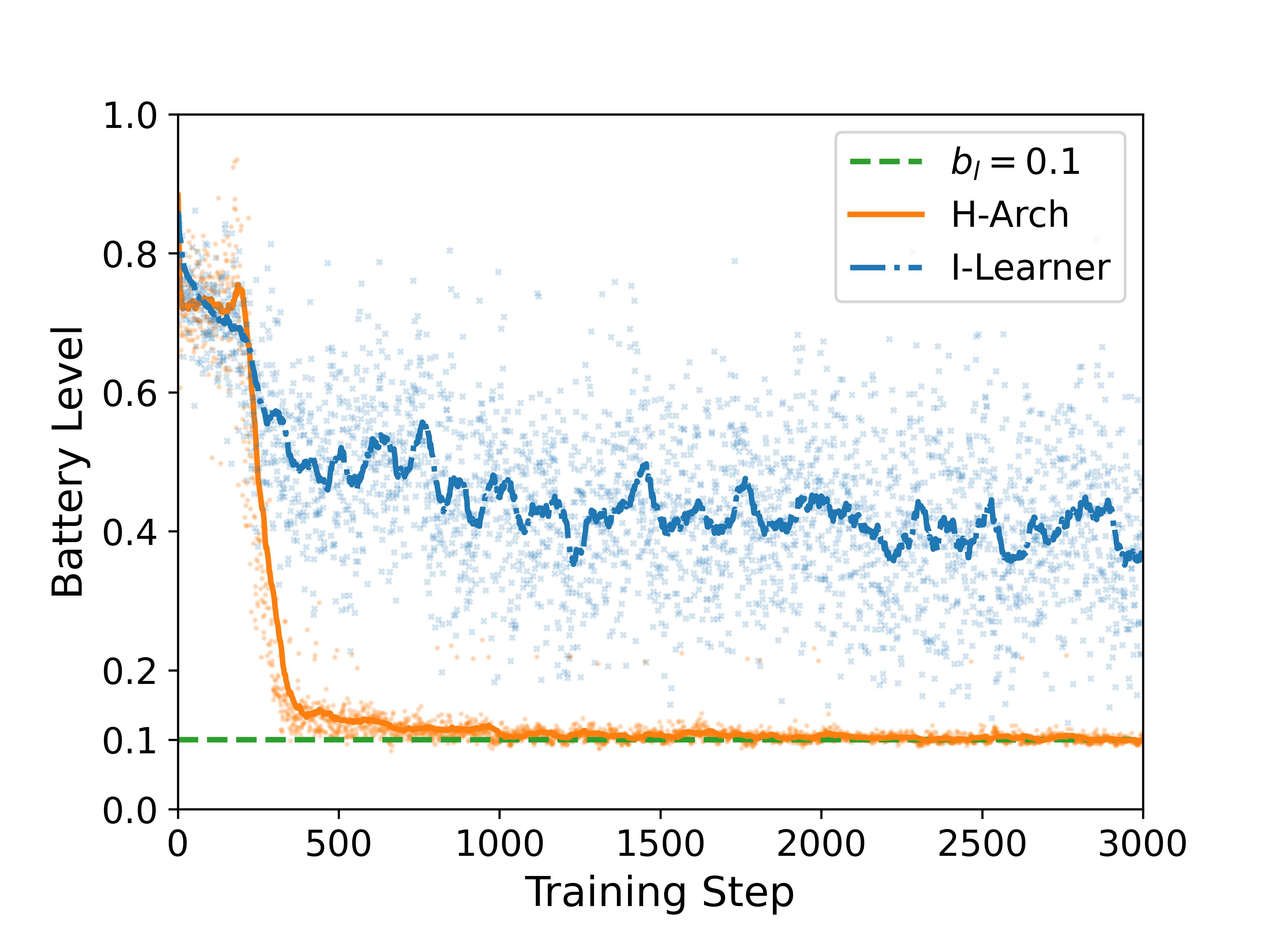}
            \caption{C-0.1 $\mathcal{B}$}
            \label{fig:6g}
        \end{subfigure} 
        \begin{subfigure}{0.23\textwidth}
            \includegraphics[width=\linewidth]{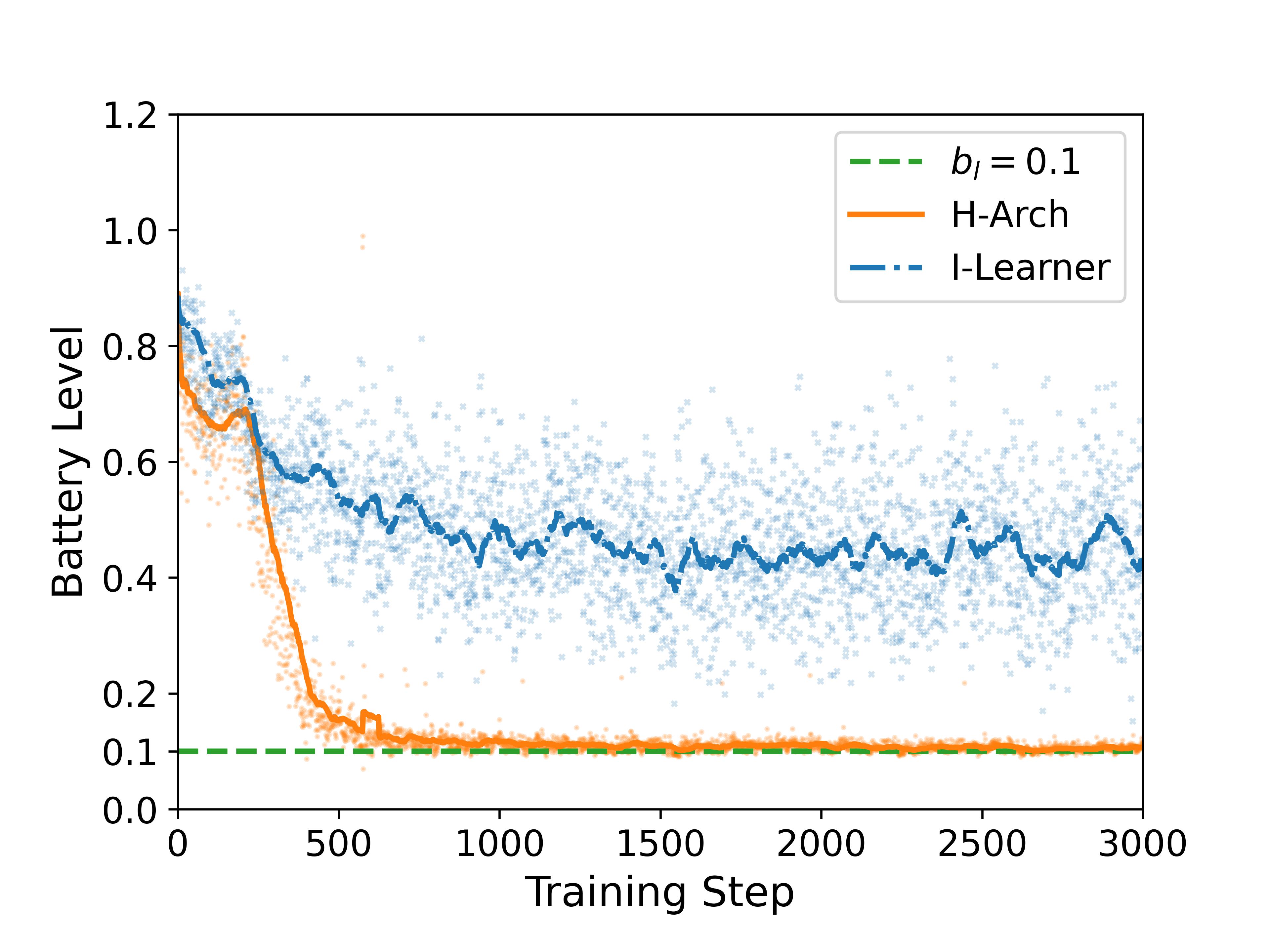}
            \caption{D-0.1 $\mathcal{B}$}
            \label{fig:6h}
        \end{subfigure} 
    
\caption{Agents' cumulative reward ($\mathcal{R}$) and battery level when recharging ($\mathcal{B}$) during training of proposed deep MARL-based models and individual learner agents on four maps with $b_l=0.1$. "H-Arch" represents the model with our proposed homogeneous MAS architecture. "I-Leaner" represents the individual learner agents. The orange points are the result of a training step of the H-Arch model. The orange solid curve shows the average of the last $50$ episodes' data of the H-Arch model. The blue crosses are the result of a training step of the I-Leaner model. The blue dash-dotted curve shows the average of the last $50$ episodes' data of the I-Leaner model.}
\label{fig:training}
\end{figure}

Fig.~\ref{fig:training} demonstrates agents' cumulative reward (Fig.~\ref{fig:6a} to Fig.~\ref{fig:6d}) and battery level when recharging (Fig.~\ref{fig:6e} to Fig.~\ref{fig:6h}) of the proposed model and individual learner agents on four maps with $b_l = 0.1$. Further results during training when $b_l = 0.15$ and $b_l = 0.2$ are included in Appendix~\ref{app:A}. If agents have not recharged in an episode, the average battery remaining when recharging in that episode will be set to $1$. 

The result indicates that the proposed MAPPO algorithm and reward function successfully train agents with homogeneous MAS architecture to pursue the maximum cumulative reward. Alongside, agents successfully learn an efficient battery recharging strategy as the cumulative reward increases, and the agents learn to recharge with a battery level that converges towards $b_l$ as the training progresses. 

When comparing the learning performance of individual learner agents and agents with the homogeneous MAS architecture, agents with the homogeneous MAS architecture learn faster and more stable than individual learner agents across all maps and $b_l$ settings. The result validates the superiority of homogeneous MAS architecture over the individual learner approach, that i) allows experience sharing between agents to improve sample efficiency, and ii) reduces the number of failure scenarios -- $4$ failure scenarios when using homogeneous MAS architecture and $31$ failure scenarios when using the individual learner approach.

In the following tests, since the individual learner agents fail to learn the battery charging strategy, the proposed model will only be compared with the CR strategy.

\subsection{Battery Recharging Performance Evaluation} \label{BRPE}

A model's battery recharging performance is evaluated based on the agents' battery failure rate, and the average battery level when agents recharge. The agent's battery failure rate is computed by the number of times that agents fail to recharge, divided by the number of times that requires the agents to recharge. The calculation is based on $10$ tests, each containing $100$ test episodes. Each test episode has a horizon of $14400$ steps, which simulates an approximate real-world 24-hour patrolling scenario. 

Table.~\ref{tab:bf-ABCD-0.1} shows the average remaining battery level when agents recharge and battery failure rate for agents patrolling on four maps with $1$ to $8$ agents and $b_l=0.1$. Further battery performance evaluation results when $b_l = 0.15$ and $b_l = 0.2$ are included in Appendix~\ref{app:B}.

The results reveal that the agents have a high success rate of battery charging ($>99.9\%$), and on average, can recharge with a remaining battery level close to the predefined $b_l$. However, agents sometimes will recharge with a battery remaining that has a variation of $\pm15\%$ from $b_l$. The variation is likely caused by the use of the linear scalarization method in the multi-objective reward function, as an agent may pursue one objective and sacrifice others, while the total reward can still be larger. For example, the agents might expend more battery than required to reduce the vertices' idleness before recharging, leading to improved patrolling performance. 

Therefore, we can conclude that the proposed methods can effectively train the agents to recharge. However, we can also conclude that the linear scalarization reward function has limitations on training agents to recharge with an \emph{exact} $b_l$ battery level remaining. 

\begin{table}[htb]
    \sisetup{round-mode=places, round-precision=5}
    \centering
    \resizebox{0.45\textwidth}{!}{%
    \begin{tabular}{lllll}%
        MAP A, $b_l = 0.1$ & & & & \\
        \sisetup{round-mode=places, round-precision=5}
        \bfseries $n$ & \bfseries $b_c$ & \bfseries $\delta_{b_c}$ & \bfseries $F$ & \bfseries $\delta_{F}$
        \begin{filecontents*}{csv/A-0.10-battery.csv}
n,bc,bcE,f,fE
1,0.10074721,0.00678492,NAN,NAN
2,0.10109227,0.00573595,1.977e-05,6.252e-05
3,0.10233,0.00504238,NAN,NAN
4,0.10325921,0.0043008,NAN,NAN
5,0.10380919,0.00366598,NAN,NAN
6,0.10428736,0.00344119,NAN,NAN
7,0.10469166,0.00328986,4.69e-06,4.19e-06
8,0.10475277,0.00299736,NAN,NAN
\end{filecontents*}
\csvreader[head to column names]{csv/A-0.10-battery.csv}{}% use head of csv as column names
        {\\\hline \n & \bc & \bcE & \f & \fE}% specify your coloumns here
    \end{tabular}
    }
    \resizebox{0.45\textwidth}{!}{%
    \begin{tabular}{lllll}%
        MAP B, $b_l = 0.1$ & & & & \\
        \sisetup{round-mode=places, round-precision=5}
        \bfseries $n$ & \bfseries $b_c$ & \bfseries $\delta_{b_c}$ & \bfseries $F$ & \bfseries $\delta_{F}$
        \begin{filecontents*}{csv/B-0.10-battery.csv}
n,bc,bcE,f,fE
1,0.09854566,0.0036235,0.00013931,7.99e-05
2,0.10225448,0.00540755,7.844e-05,0.00013682
3,0.10434858,0.00516087,3.936e-05,8.856e-05
4,0.10541307,0.00442446,3.884e-05,6.771e-05
5,0.10607322,0.00405472,7.78e-06,2.462e-05
6,0.10673131,0.00382278,1.236e-05,1.104e-05
7,0.10711379,0.00375296,1.85e-05,1.296e-05
8,0.10745261,0.00357376,6.96e-06,5.19e-06
\end{filecontents*}
\csvreader[head to column names]{csv/B-0.10-battery.csv}{}% use head of csv as column names
        {\\\hline \n & \bc & \bcE & \f & \fE}% specify your coloumns here
    \end{tabular}
    }
    \resizebox{0.45\textwidth}{!}{%
    \begin{tabular}{lllll}%
        MAP C, $b_l = 0.1$ & & & & \\
        \sisetup{round-mode=places, round-precision=5}
        \bfseries $n$ & \bfseries $b_c$ & \bfseries $\delta_{b_c}$ & \bfseries $F$ & \bfseries $\delta_{F}$
        \begin{filecontents*}{csv/C-0.10-battery.csv}
n,bc,bcE,f,fE
1,0.09590637,0.00456932,7.959e-05,0.00016778
2,0.09864255,0.00464694,0.00013856,0.00018768
3,0.1003357,0.00384435,5.245e-05,9.165e-05
4,0.10148766,0.0033111,9.84e-06,3.112e-05
5,0.10200015,0.00290944,2.349e-05,5.28e-05
6,0.10237552,0.00264264,3.107e-05,1.258e-05
7,0.10265097,0.00256302,1.081e-05,8.29e-06
8,0.10280749,0.00353097,6.81e-06,7.34e-06
\end{filecontents*}
\csvreader[head to column names]{csv/C-0.10-battery.csv}{}% use head of csv as column names
        {\\\hline \n & \bc & \bcE & \f & \fE}% specify your coloumns here
    \end{tabular}
    }
    \resizebox{0.45\textwidth}{!}{%
    \begin{tabular}{lllll}%
        MAP D, $b_l = 0.1$ & & & & \\
        \sisetup{round-mode=places, round-precision=5}
        \bfseries $n$ & \bfseries $b_c$ & \bfseries $\delta_{b_c}$ & \bfseries $F$ & \bfseries $\delta_{F}$
        \begin{filecontents*}{csv/D-0.10-battery.csv}
n,bc,bcE,f,fE
1,0.10657389,0.00572943,0.00013936,0.00012427
2,0.11009074,0.00622262,0.0001051,0.0001752
3,0.11228223,0.00587374,NAN,NAN
4,0.11320009,0.0054737,NAN,NAN
5,0.11417475,0.00525584,NAN,NAN
6,0.11421678,0.00490541,NAN,NAN
7,0.1143286,0.00458728,NAN,NAN
8,0.11408744,0.00406333,NAN,NAN
\end{filecontents*}
\csvreader[head to column names]{csv/D-0.10-battery.csv}{}% use head of csv as column names
        {\\\hline \n & \bc & \bcE & \f & \fE}% specify your coloumns here
    \end{tabular}
    }
    
    \caption{Model A/B/C/D-$0.1$'s battery recharging performance when running with $1$ to $8$ number of patrolling agents. $n$ stands for the number of patrolling agents. $b_c$ stands for the average of the remaining battery level when agents recharge, and $\delta_{b_c}$ is the corresponding standard deviation. $F$ stands for agents' battery failure rate, and $\delta_{F}$ is the corresponding standard deviation. If failures are not detected in $10$ tests with $100$ test episodes, $F$ and $\delta_{F}$ are marked as \emph{NAN}}
    \label{tab:bf-ABCD-0.1}
\end{table}

\subsection{Patrolling Performance Evaluation} \label{PPE}

\begin{figure}[hbt]
    \begin{subfigure}[b]{.45\linewidth}
        \centering
        \includegraphics[width=\linewidth]{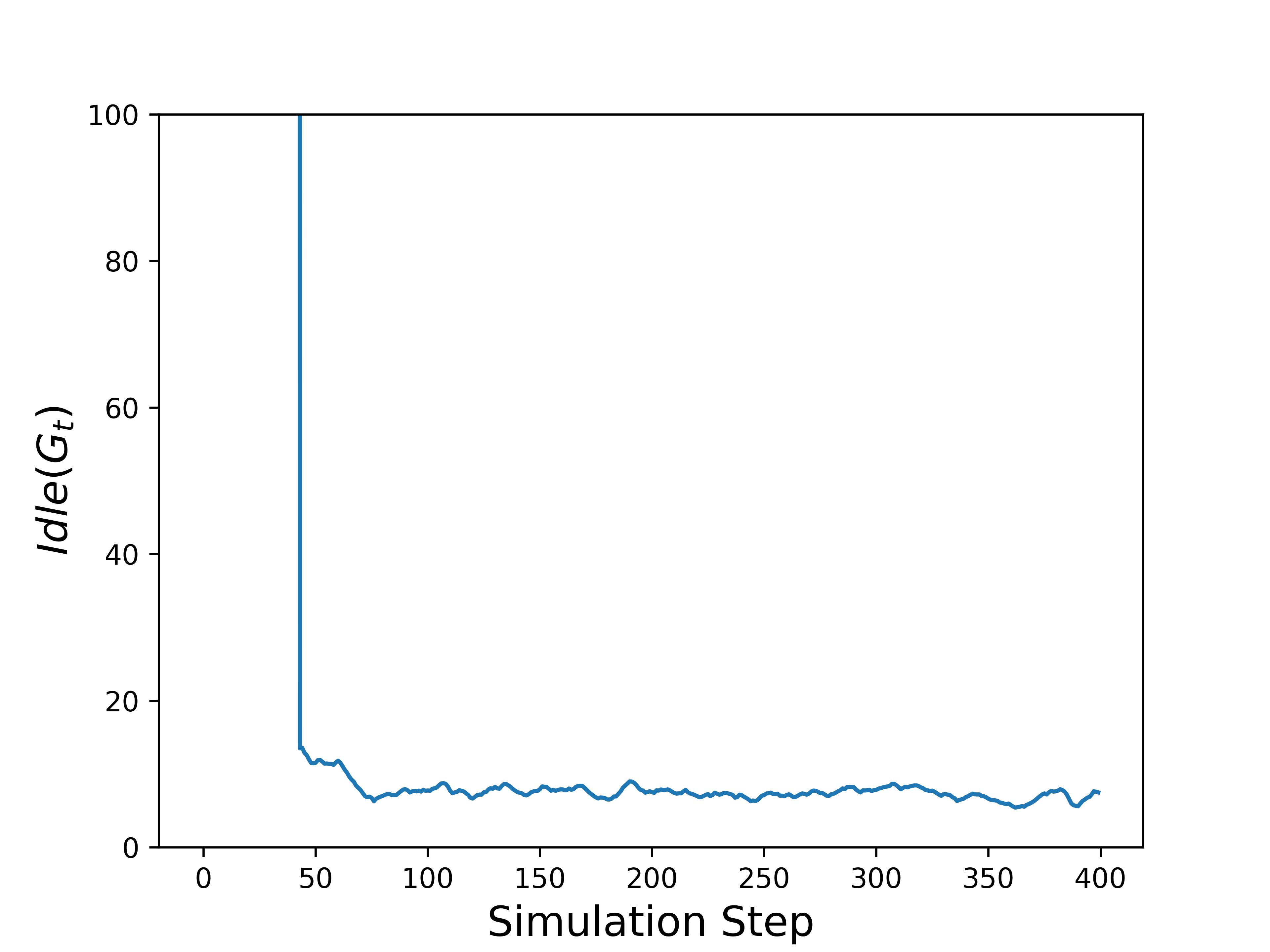}
        \caption{$max(Idle(v_t))$}
        \label{fig:7a}
    \end{subfigure}\hfill
    %%%%%
    \begin{subfigure}[b]{.45\linewidth}
        \centering
        \includegraphics[width=\linewidth]{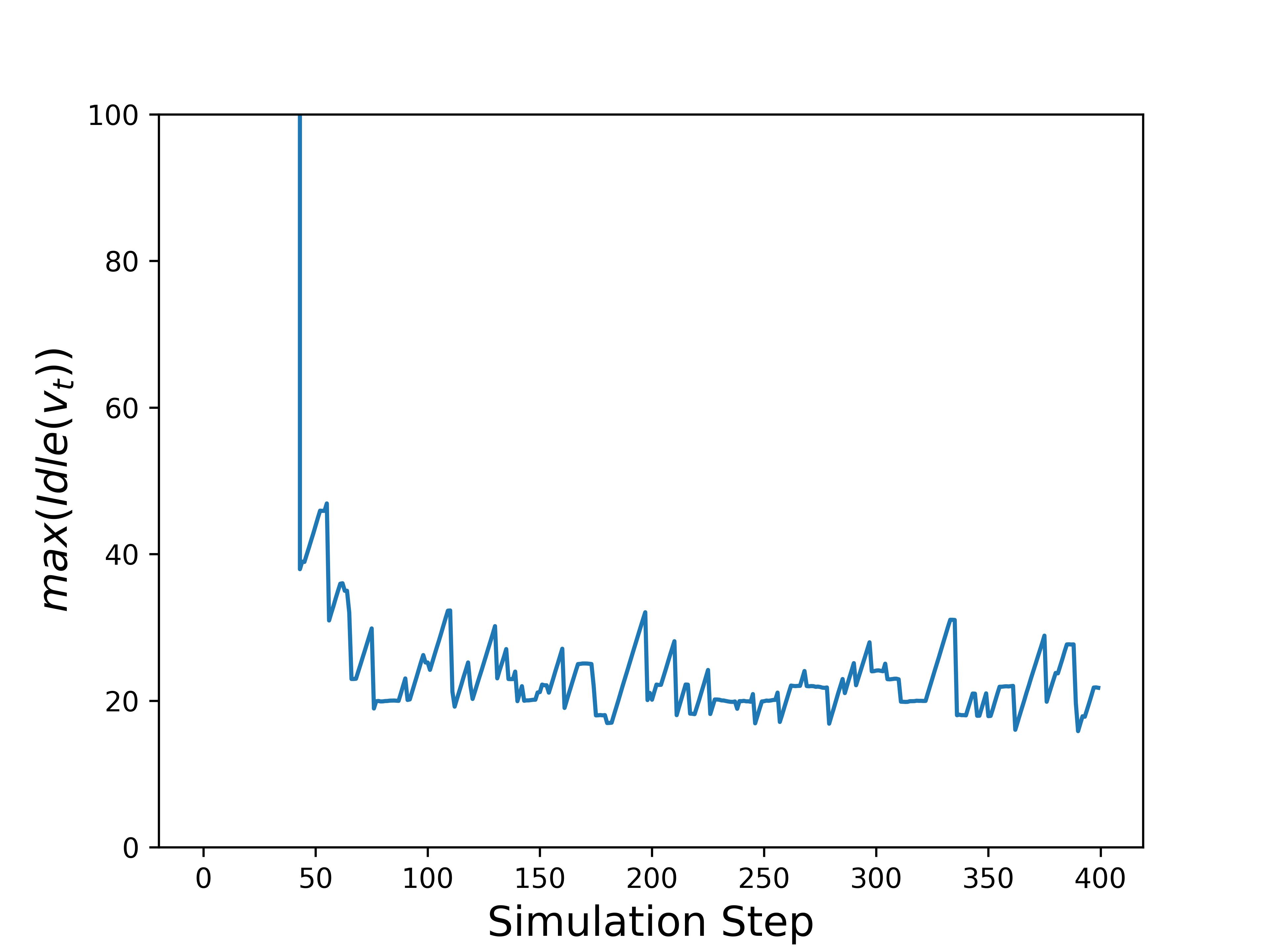}
        \caption{$Idle(G_t)$}
        \label{fig:7b}
    \end{subfigure}\hfill
    \caption{$max(Idle(v_t))$ and $Idle(G_t)$ of the first $400$ steps of a test episode run with \emph{A-$0.1$} with $5$ patrolling agents. For readability, the values of $max(Idle(v_t))$ and $Idle(G_t)$ that are larger than $100$ are not shown.}
    \label{fig:short-episode}
\end{figure}

The model's patrolling performance is assessed based on $AVG^h(G)$ and $\overline{MAX^h(G)}$, which values are computed from data collected from $100$ successfully completed episodes with no battery failures occurring. In a test episode, the $Idle(G_t)$ and $max(Idle(v_t))$ are collected after $150$ steps, when both $Idle(G_t)$ and $max(Idle(v_t))$ have stabilised, as depicted in Fig.~\ref{fig:short-episode}. It is worth noting that deep MARL-based agents do not recharge exactly with $b_l$ battery remaining. Therefore, for a fair comparison, the CR agents and deep MARL-based agents are set to recharge with approximately the same amount of battery level remaining.

\begin{figure}[htb]

        \begin{subfigure}{0.24\textwidth}
            \includegraphics[width=\linewidth]{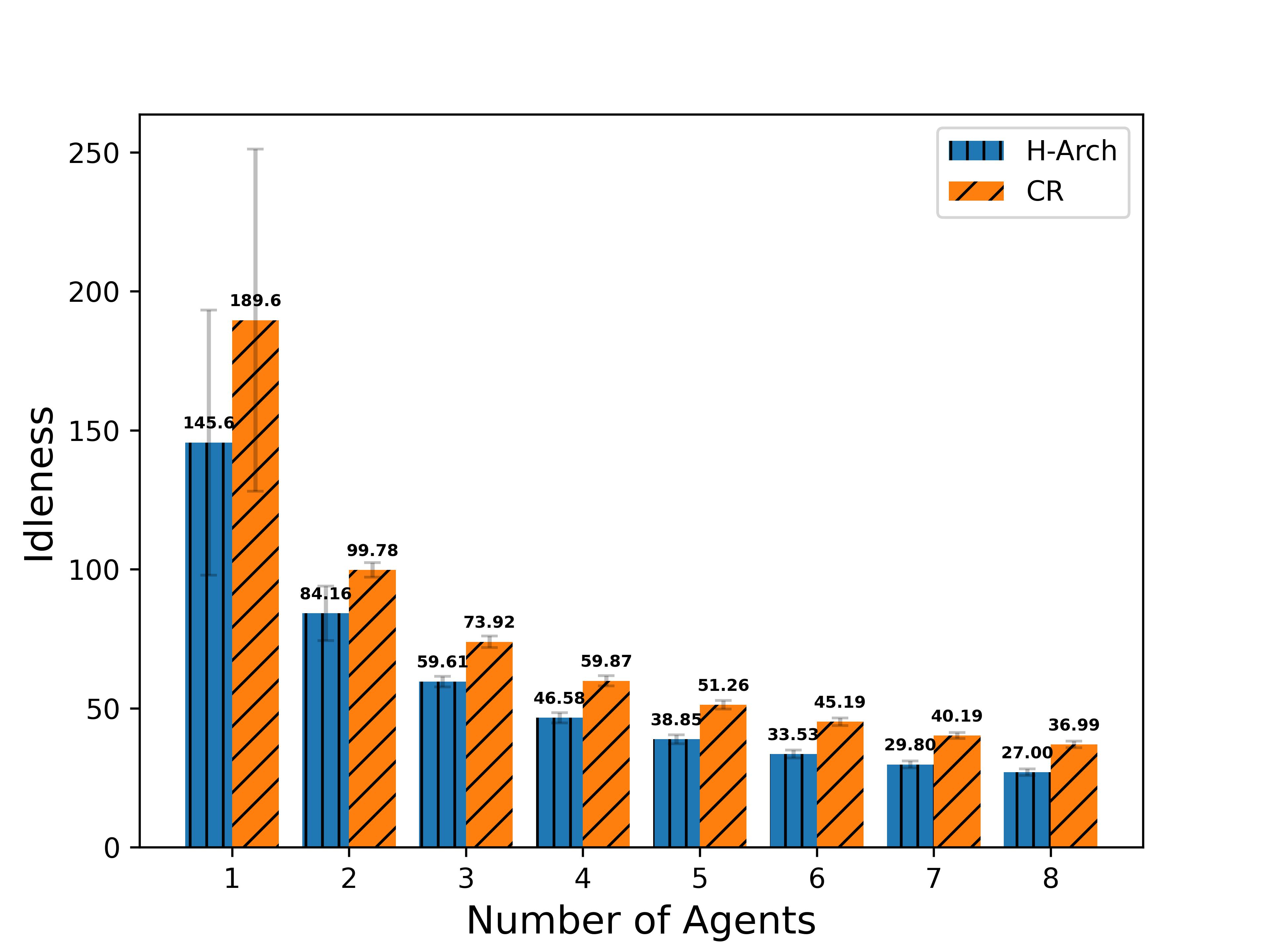}
            \caption{A-0.1 $\overline{MAX^h(G)}$}
            \label{fig:8a}
        \end{subfigure} \hfill
        \begin{subfigure}{0.24\textwidth}
            \includegraphics[width=\linewidth]{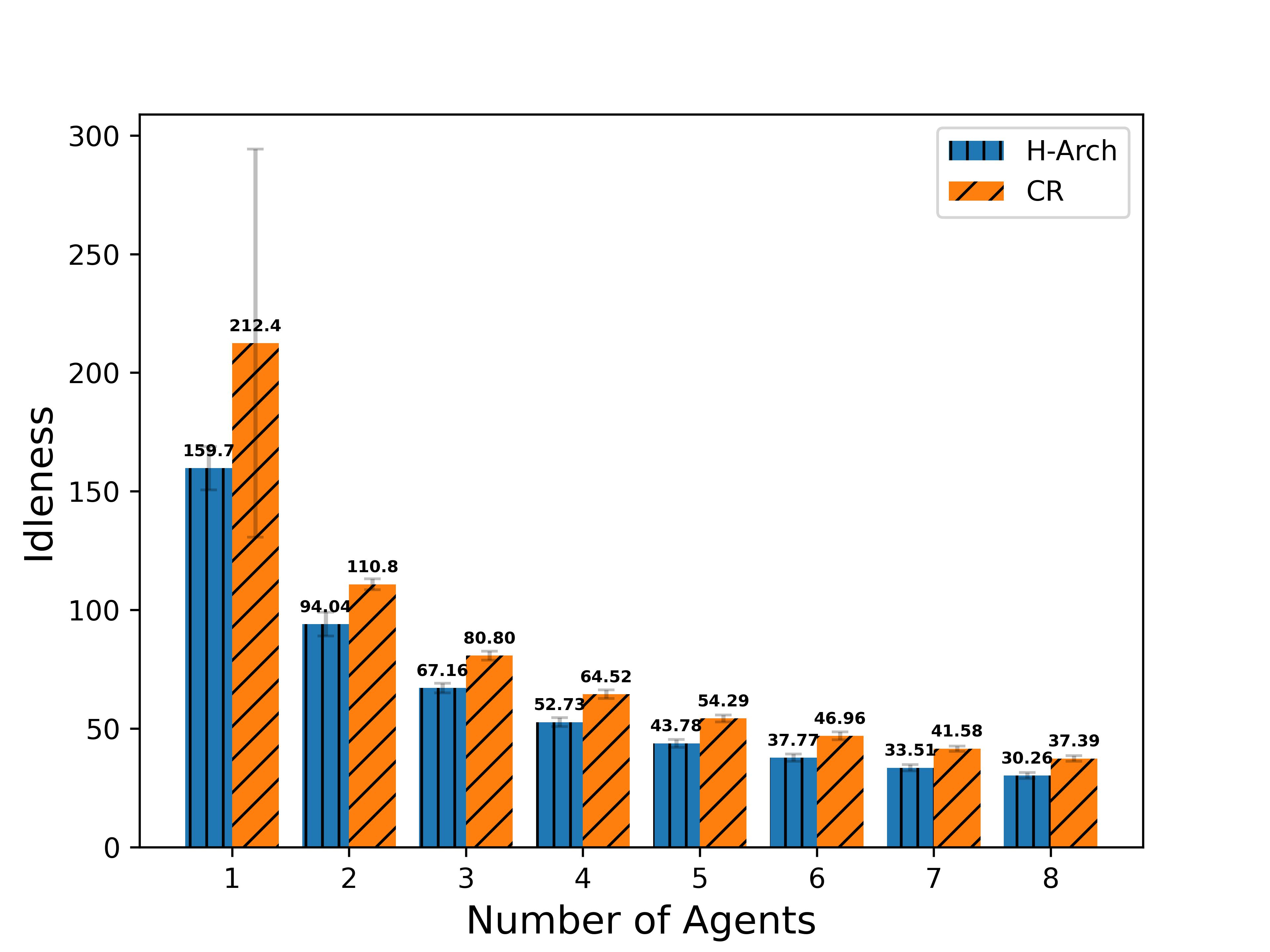}
            \caption{B-0.1 $\overline{MAX^h(G)}$}
            \label{fig:8b}
        \end{subfigure} \hfill
        \begin{subfigure}{0.24\textwidth}
            \includegraphics[width=\linewidth]{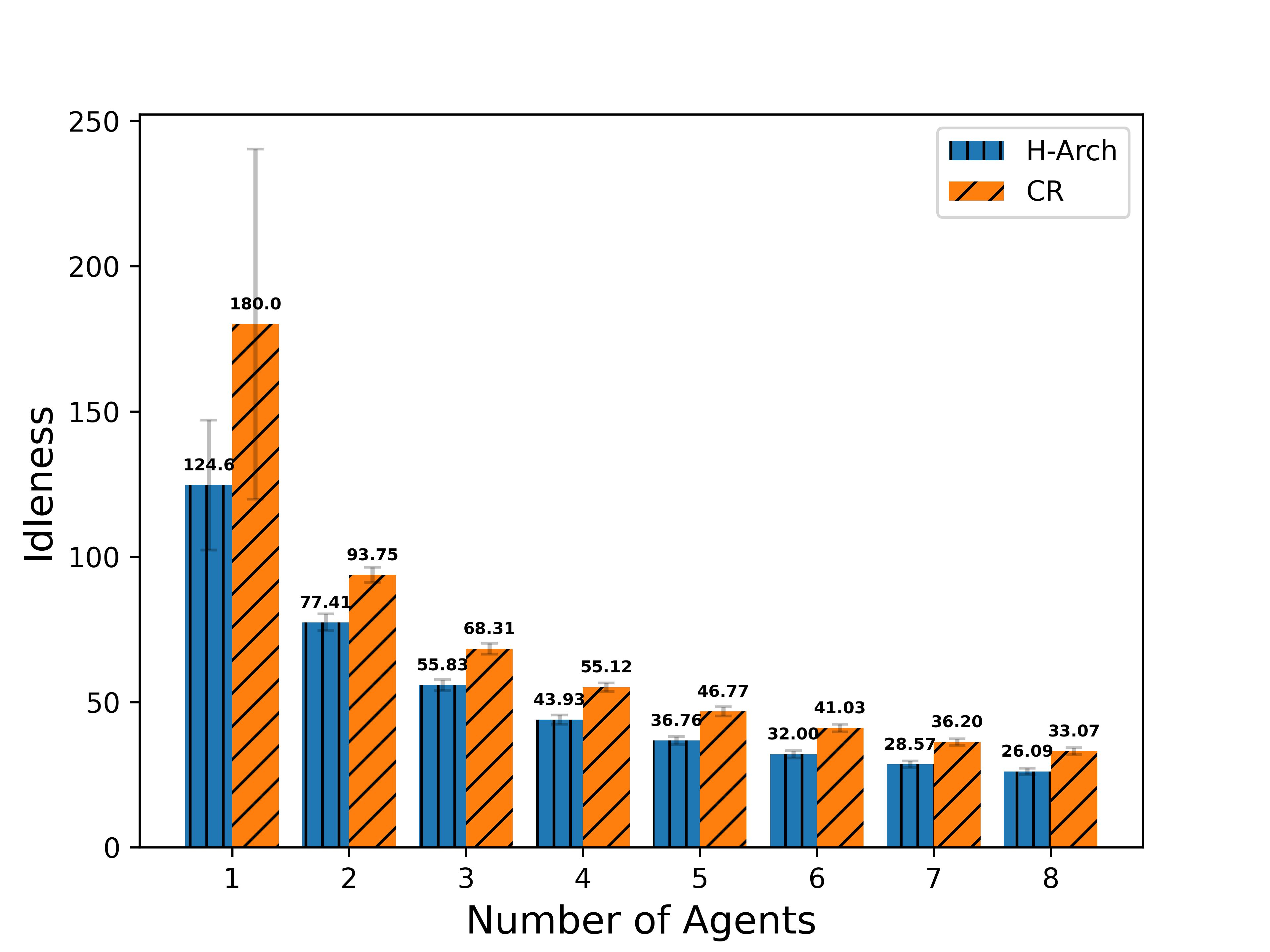}
            \caption{C-0.1 $\overline{MAX^h(G)}$}
            \label{fig:8c}
        \end{subfigure} \hfill
        \begin{subfigure}{0.24\textwidth}
            \includegraphics[width=\linewidth]{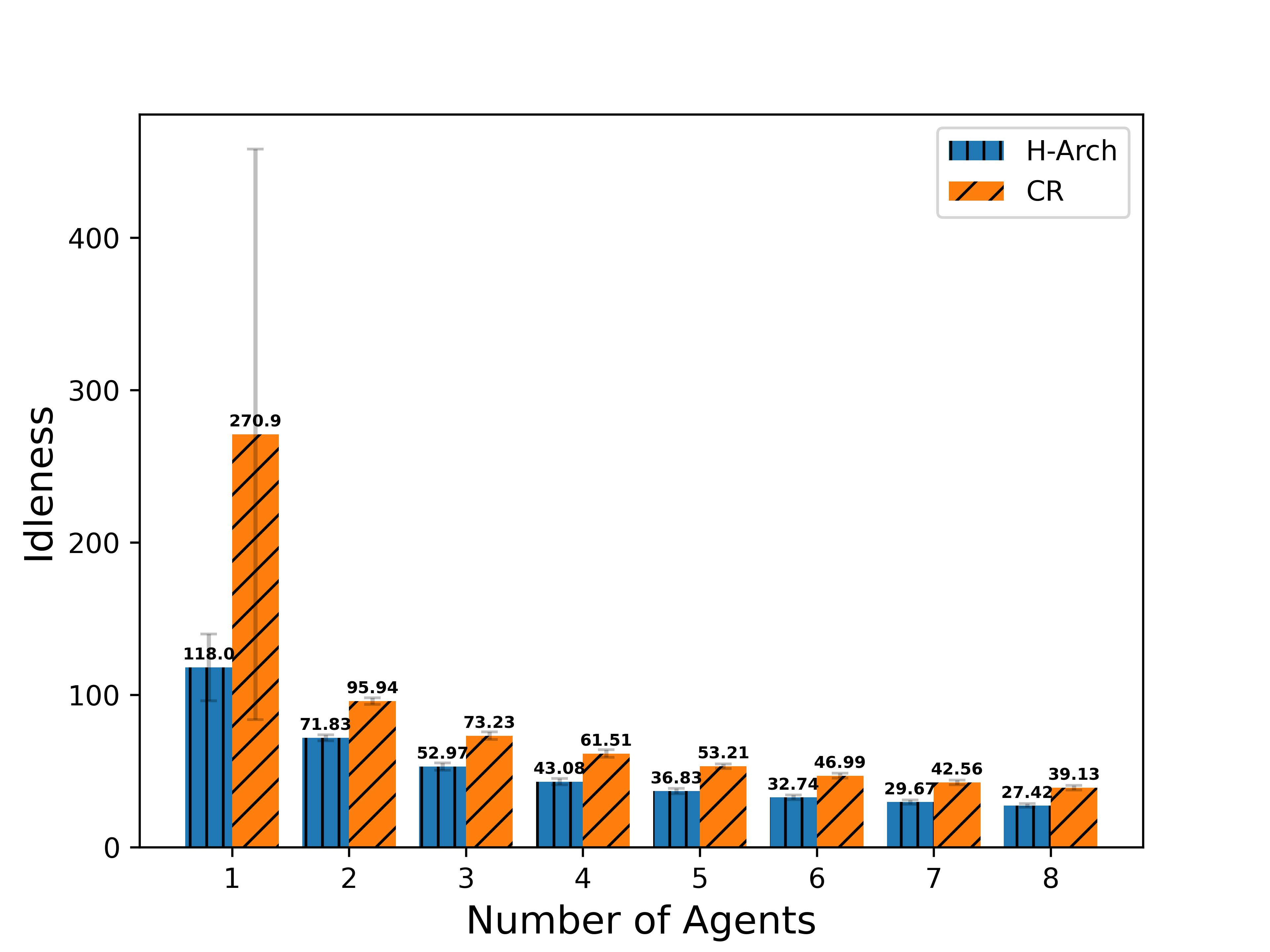}
            \caption{D-0.1 $\overline{MAX^h(G)}$}
            \label{fig:8d}
        \end{subfigure} \hfill
        
        \begin{subfigure}{0.24\textwidth}
            \includegraphics[width=\linewidth]{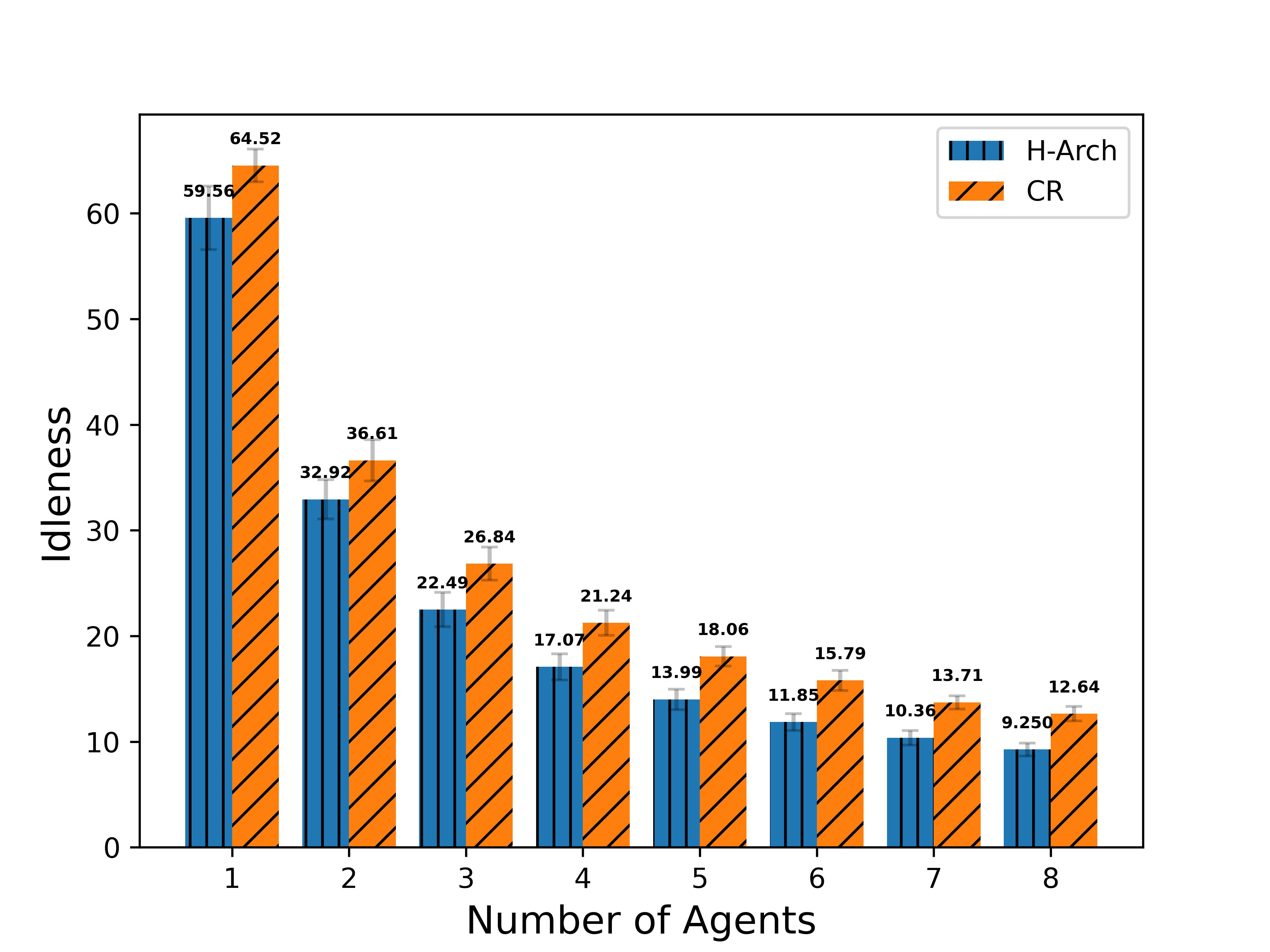}
            \caption{A-0.1 $AVG^h(G)$}
            \label{fig:8e}
        \end{subfigure} \hfill
        \begin{subfigure}{0.24\textwidth}
            \includegraphics[width=\linewidth]{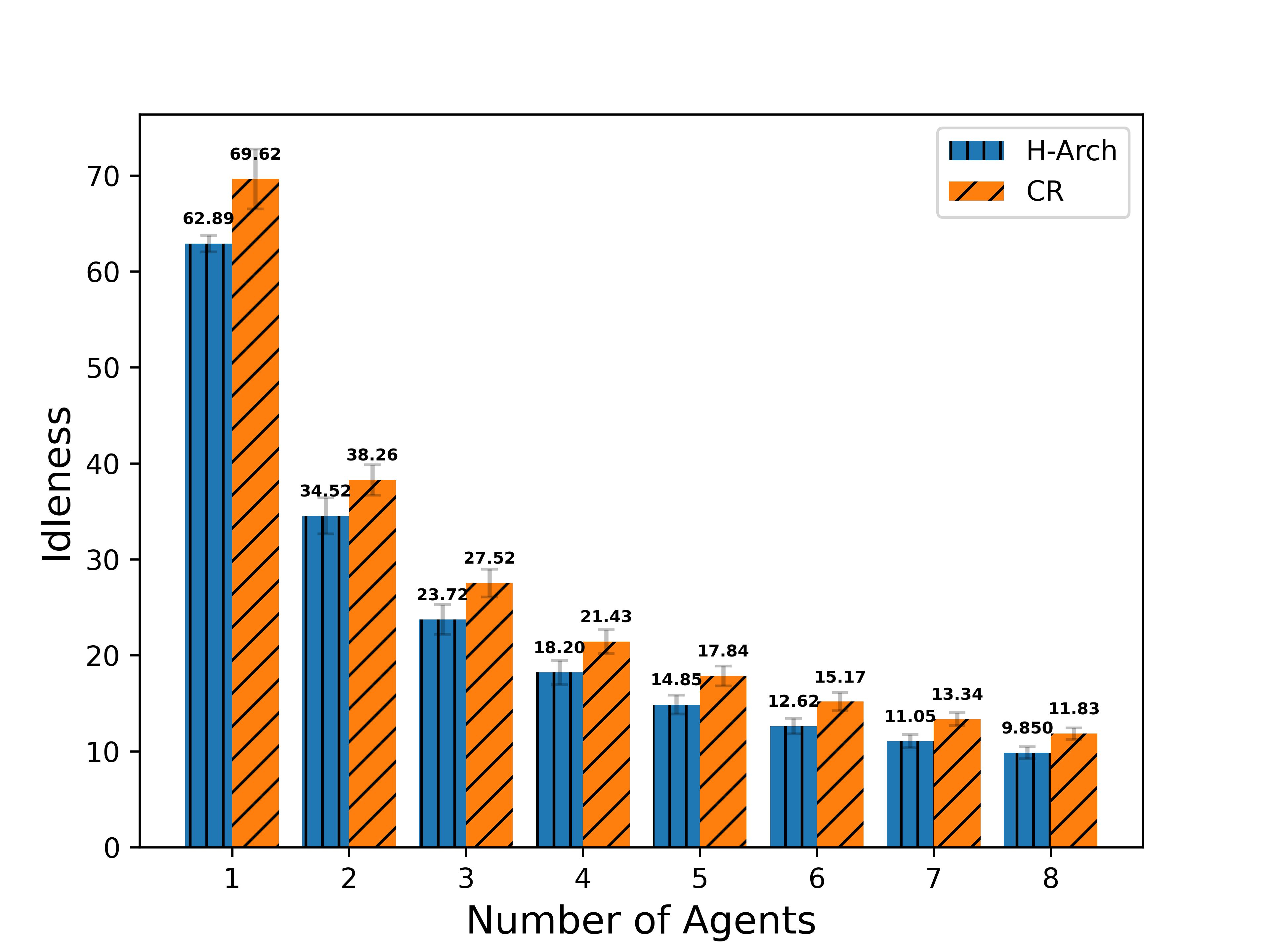}
            \caption{B-0.1 $AVG^h(G)$}
            \label{fig:8f}
        \end{subfigure} \hfill
        \begin{subfigure}{0.24\textwidth}
            \includegraphics[width=\linewidth]{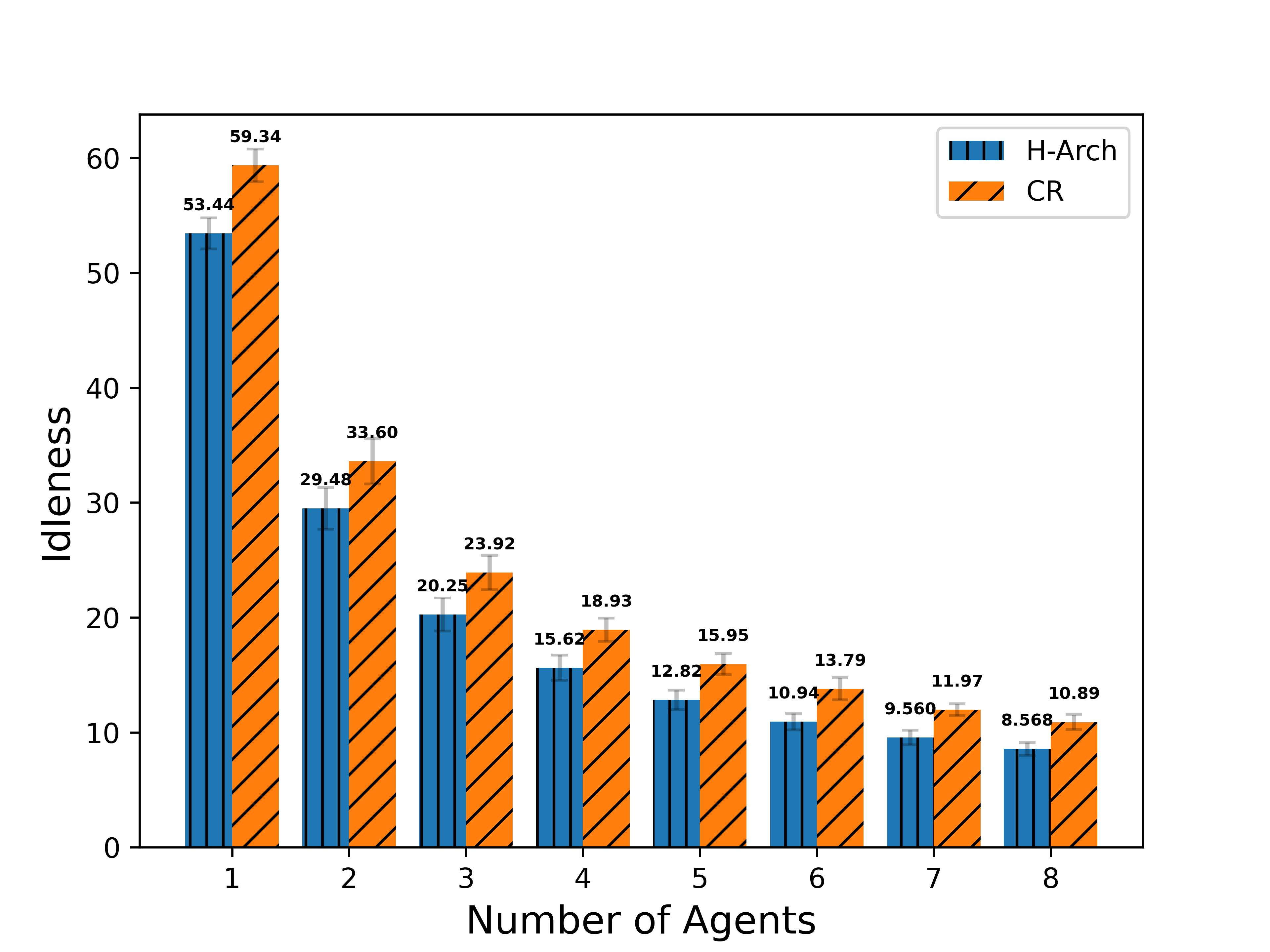}
            \caption{C-0.1 $AVG^h(G)$}
            \label{fig:8g}
        \end{subfigure} \hfill
        \begin{subfigure}{0.24\textwidth}
            \includegraphics[width=\linewidth]{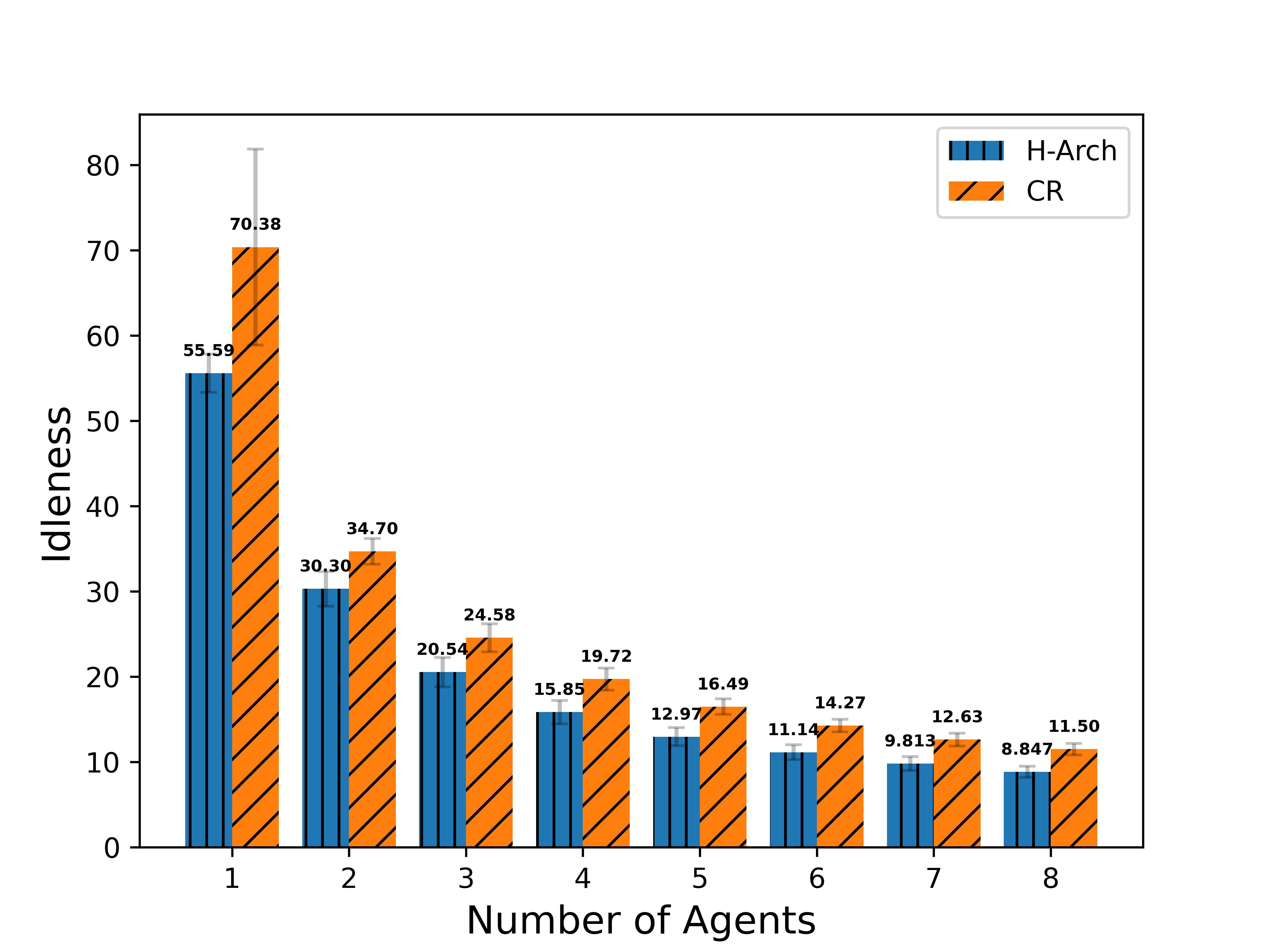}
            \caption{D-0.1 $AVG^h(G)$}
            \label{fig:8h}
        \end{subfigure} \hfill
    
\caption{Patrolling performance result (evaluated by $AVG^h(G)$ and $\overline{MAX^h(G)}$) of the proposed deep MARL-based model (H-Arch) and Cognitive Reactive strategy (CR) on four maps running with $1$ to $8$ number of patrolling agents and $b_l=0.1$. The unit of idleness is the length of a discrete timestep (0.1 minutes). The lower the $AVG^h(G)$ and $\overline{MAX^h(G)}$ are, the better the model performs.}
\label{fig:pp-eval}
\end{figure}

Fig.~\ref{fig:pp-eval} demonstrate the patrolling performance results of the proposed MARL-based model and CR strategy on four maps running with $1$ to $8$ number of patrolling agents and $b_l=0.1$. Further patrolling performance results when $b_l=0.15$ and $b_l=0.2$ are demonstrated in Appendix ~\ref{app:C}. 

The results reveal that the MARL-based model outperforms the CR strategy on all maps with a different number of patrolling agents and different $b_l$ settings. This validates that the proposed learning algorithm and reward function can effectively train agents to optimise both $AVG^h(G)$ and $\overline{MAX^h(G)}$ criteria.

\subsection{Fault tolerance and ability to cooperate with supplementary agents}

%In this set of experiments we evaluate the ability of the system to perform in the presence of failures for maps A-D with bl=0-.1...

We evaluate the fault tolerance of the agents and their ability to cooperate with supplementary agents by running test episodes with models on four maps with $b_l = 0.1$. Each test episode simulates approximately a $200$ days patrolling scenario. 

Since agents have a low battery failure rate (Table ~\ref{tab:bf-ABCD-0.1} and Appendix ~\ref{app:B}), it could take a long time to observe a battery failure scenario. Therefore, we deliberately fail a random number of agents every $10$ days in the simulation, leaving at least one active patrolling agent. In addition, a random number ($\geq0$) of supplementary agents will be introduced every $10$ days, with a maximum $8$ number of patrolling agents in the system.  At the beginning of a test episode, there will be a random number of patrolling agents, between $1$ and $8$, in the system.

% The patrolling system begins with $8$ patrolling agents, and we fail two agents every $30$ days in the simulation until only $1$ agent is left; after that, two supplementary agents are added to the system every $30$ days until there are $8$ patrolling agents. The battery failure rate is shown in Table ~\ref{tab:bf-ABCD-0.1} and Appendix ~\ref{app:B}.
% The $AVG^h(G)$, $\overline{MAX^h(G)}$ arecalculated based on the average of $max(Idle(v_t))$ and $Idle(G_t)$ collected in each day in the simulation. Similarly, the remaining battery level when recharging is calculated based on the average of the agents' recharging data in each day.
% %the patrolling and battery recharging performance are averaged over a day...

\begin{figure}[hb]

        \begin{subfigure}[t]{0.32\textwidth}
            \includegraphics[width=\linewidth]{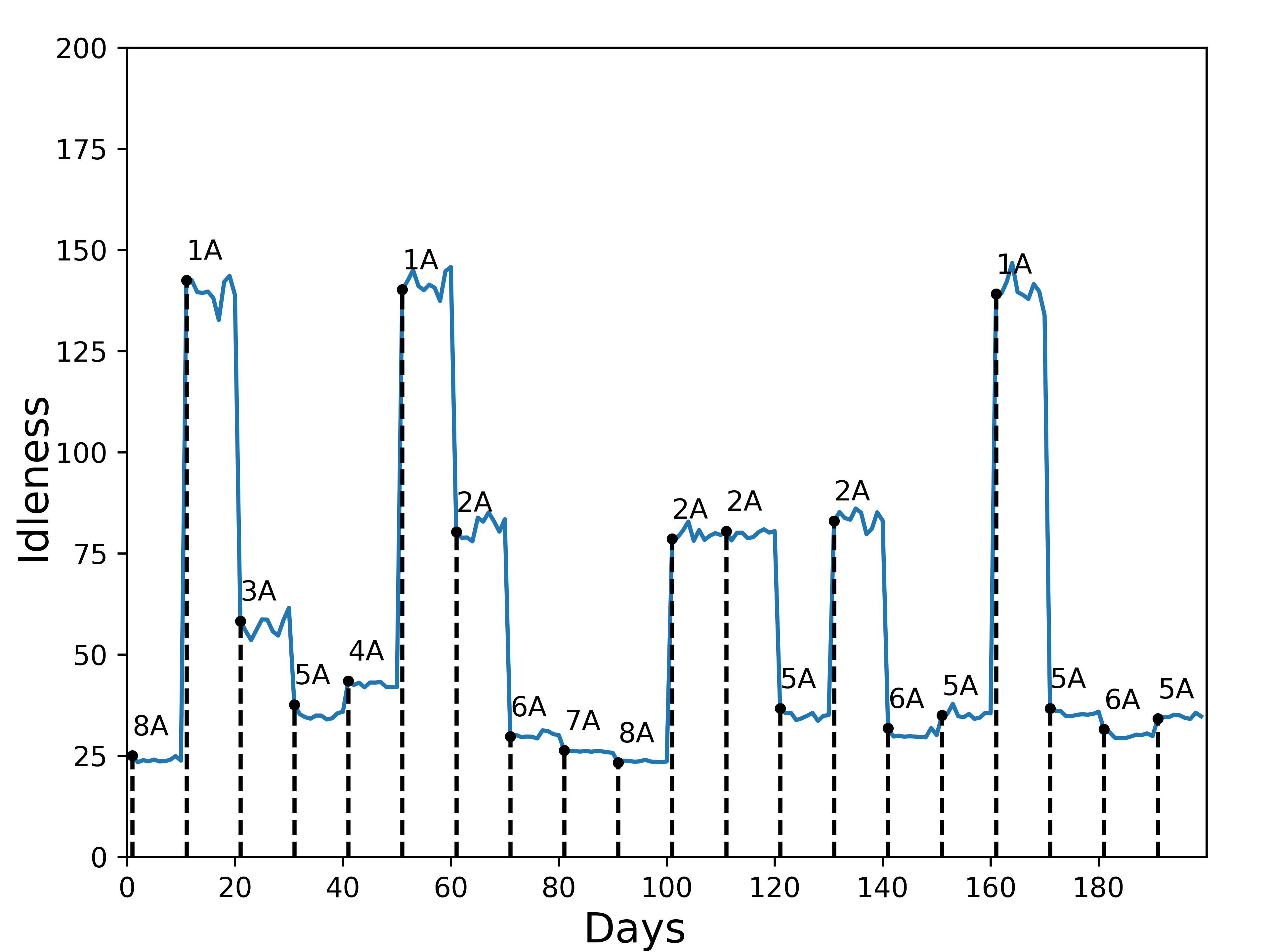}
            \caption{$\overline{MAX^h(G)}$ of Model A-$0.1$}
        \end{subfigure} \hfill
        \begin{subfigure}[t]{0.32\textwidth}
            \includegraphics[width=\linewidth]{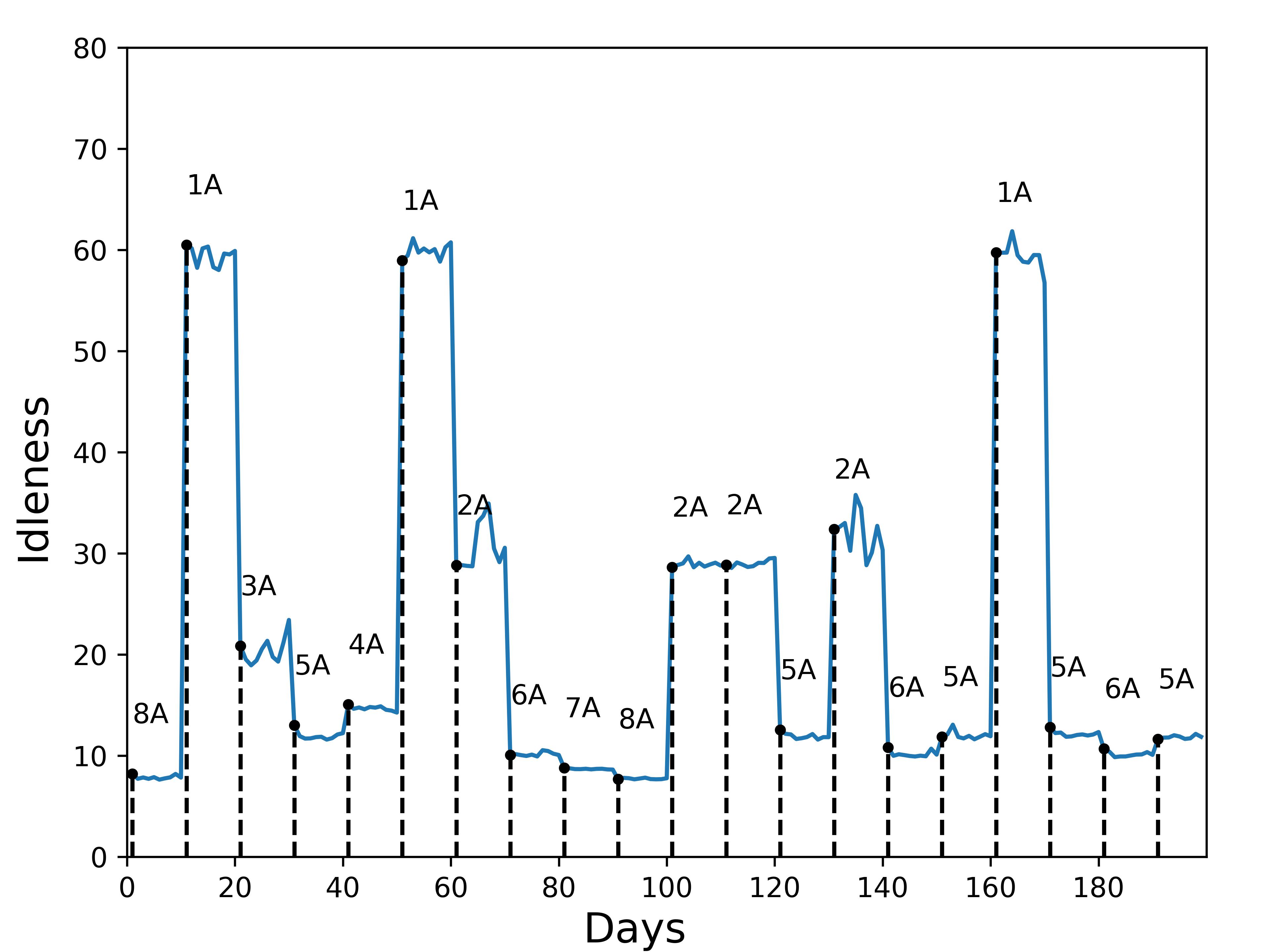}
            \caption{$AVG^h(G)$ of Model A-$0.1$}
        \end{subfigure} \hfill
        \begin{subfigure}[t]{0.32\textwidth}
            \includegraphics[width=\linewidth]{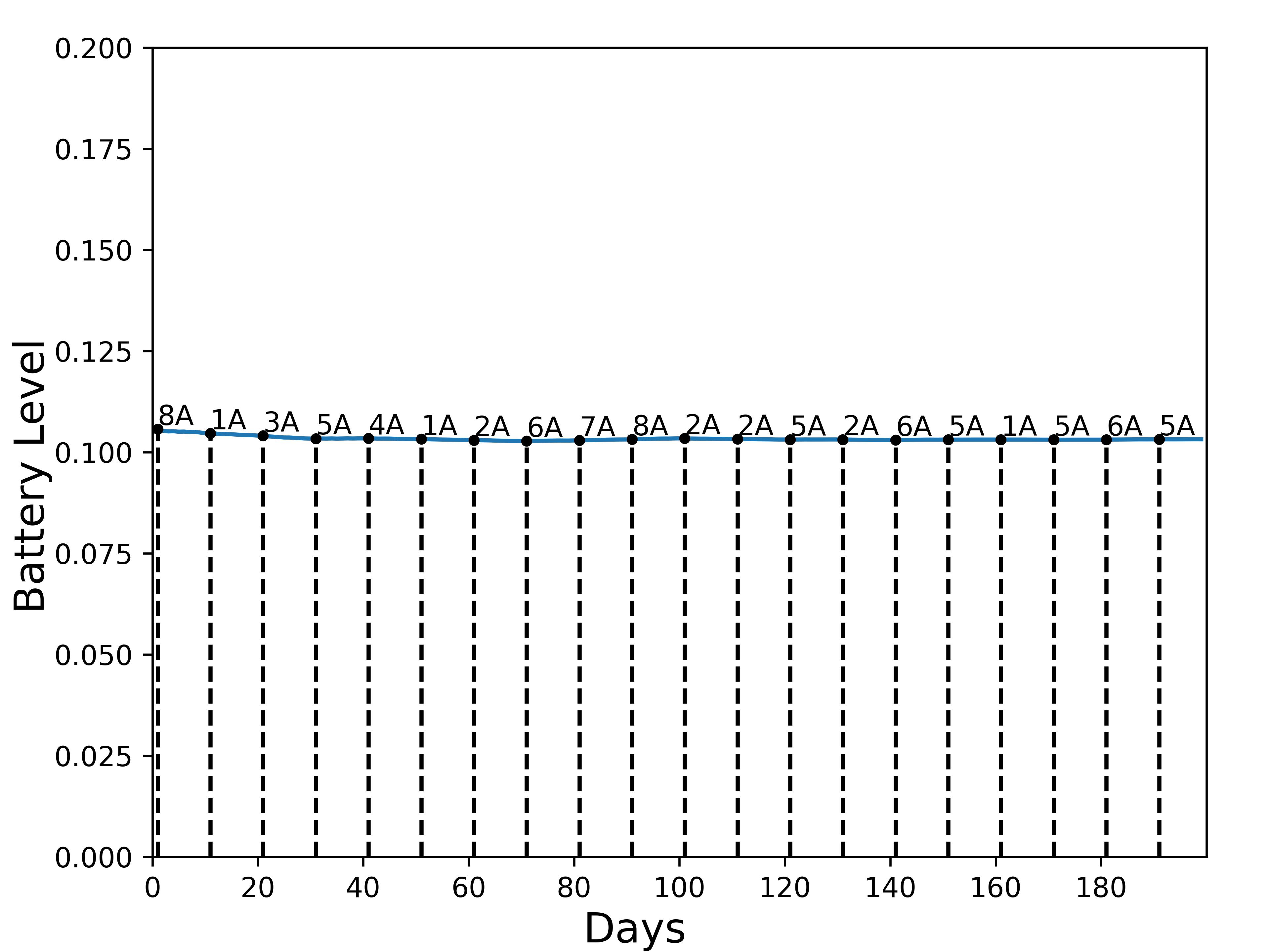}
            \caption{Battery remaining when recharging of Model A-$0.1$}
        \end{subfigure}
        
\caption{$\overline{MAX^h(G)}$), $AVG^h(G)$, and Battery remaining when recharging of model A-$0.1$ when patrolling with a varying number of agents in test episodes with a horizon of $200$ days. The data are measured daily based. The vertical dash line marks the time step when agent failures occur and when supplementary agents are introduced. The label $nA$ over the solid line indicates the number of patrolling agents during the time between two vertical dash lines. For example, $3A$ represents there are $3$ patrolling agents. }
\label{fig:long}
\end{figure}

Fig.~\ref{fig:long} shows the $\overline{MAX^h(G)}$, $AVG^h(G)$, and the battery remaining when agents recharge, of the model A-$0.1$. The performance result of models B/C/D-$0.1$ are shown in Appendix ~\ref{app:long}. From the result, we can see that when agent failures occur, or when supplementary agents are introduced, the agents will patrol with the expected patrolling performance as shown in Fig. ~\ref{fig:pp-eval}, and recharge with the expected battery remaining as shown in Table ~\ref{tab:bf-ABCD-0.1} and Appendix ~\ref{app:B}. Therefore, the system's patrolling performance when agent failure occurs and when supplementary agents are introduced is demonstrated.

%% file: 07_appendix.tex
\appendix
\section{Training result of proposed deep MARL-based models and models using individual learner approach on four maps with $b_l=0.15$ and $b_l=0.2$} \label{app:A}

\begin{figure}[htb]
    \centering 
        \begin{subfigure}{0.23\textwidth}
            \includegraphics[width=\linewidth]{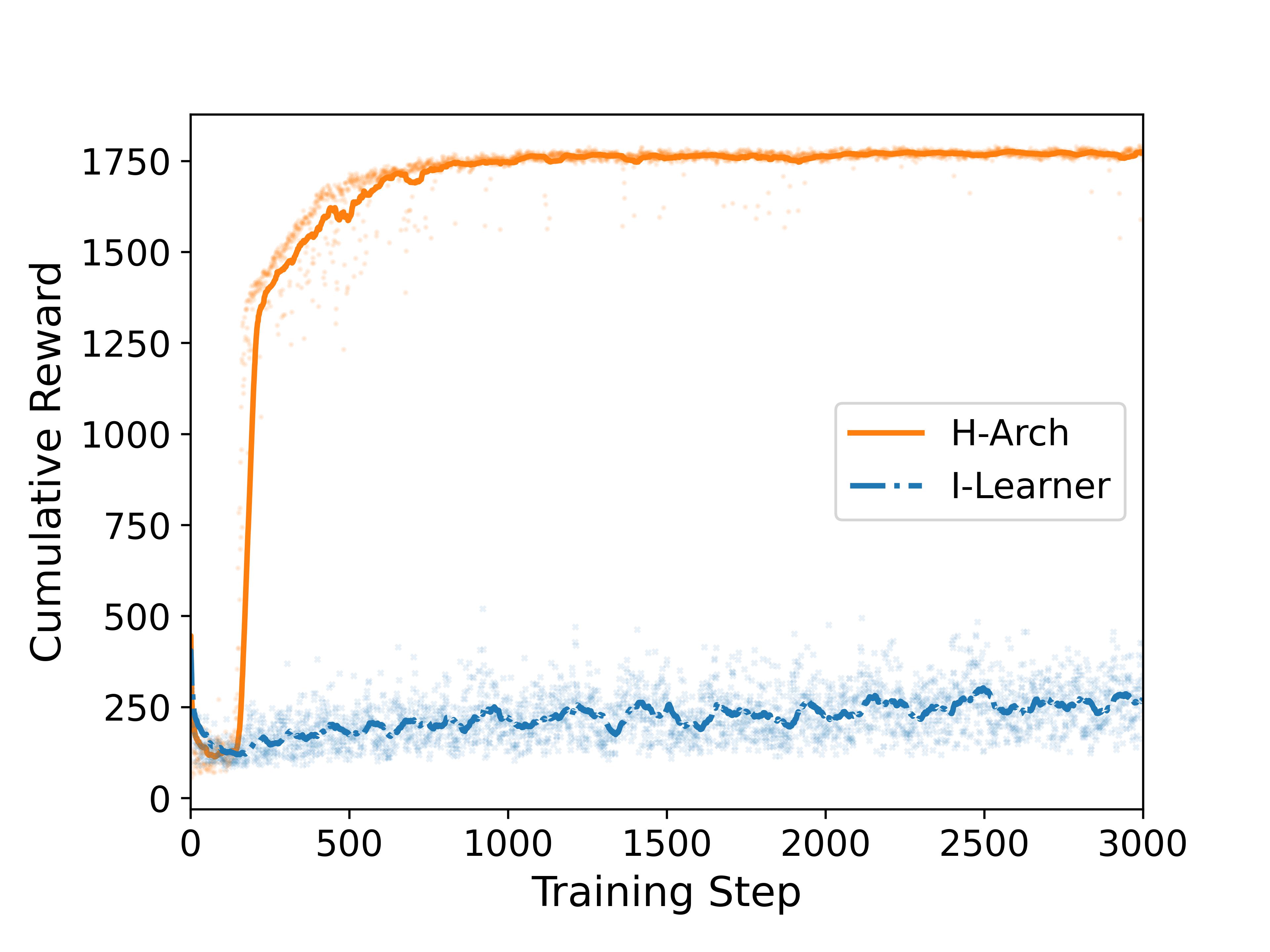}
            \caption{A-0.15 $\mathcal{R}$}
        \end{subfigure} 
        \begin{subfigure}{0.23\textwidth}
            \includegraphics[width=\linewidth]{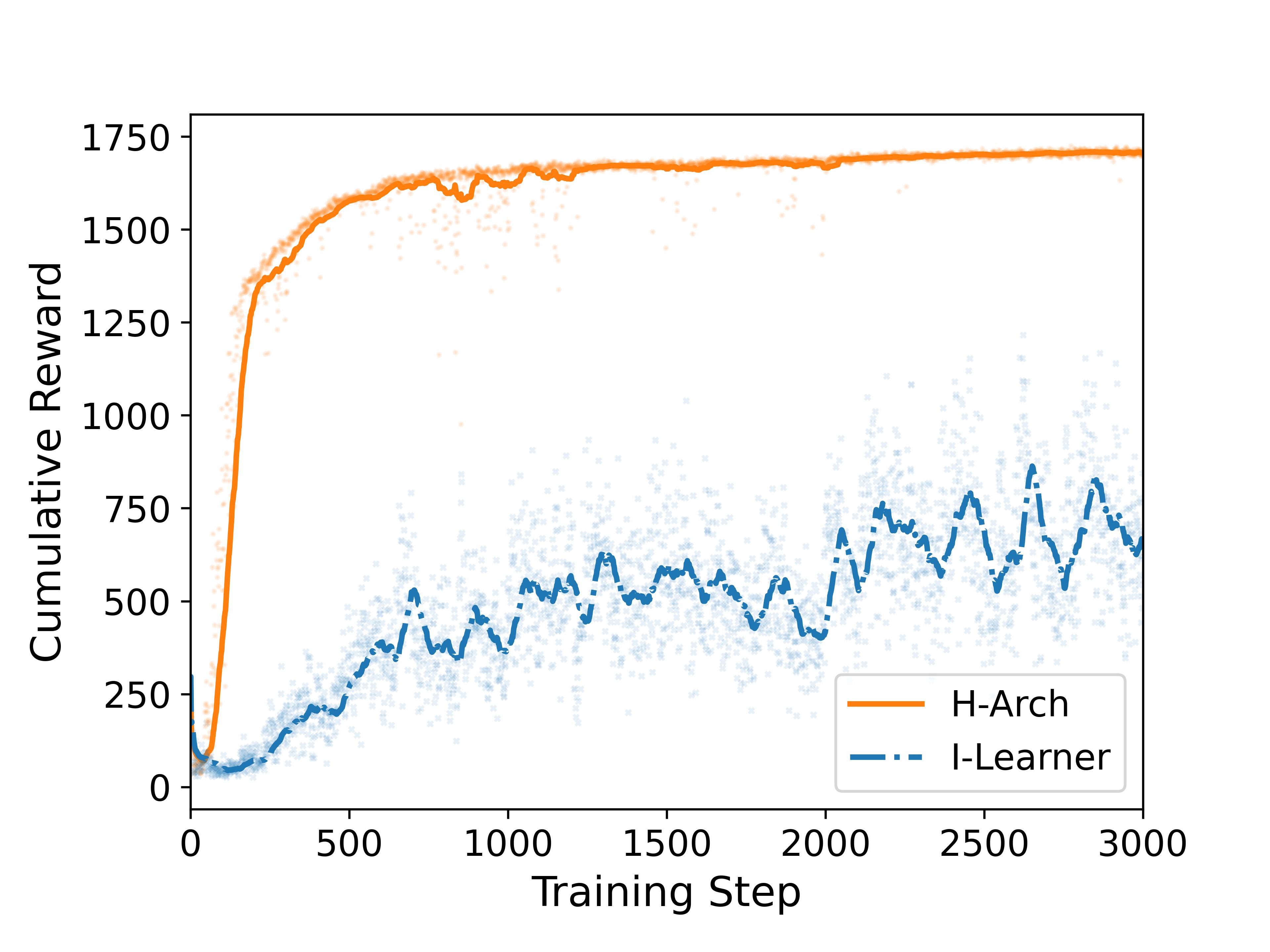}
            \caption{B-0.15 $\mathcal{R}$}
        \end{subfigure} 
        \begin{subfigure}{0.23\textwidth}
            \includegraphics[width=\linewidth]{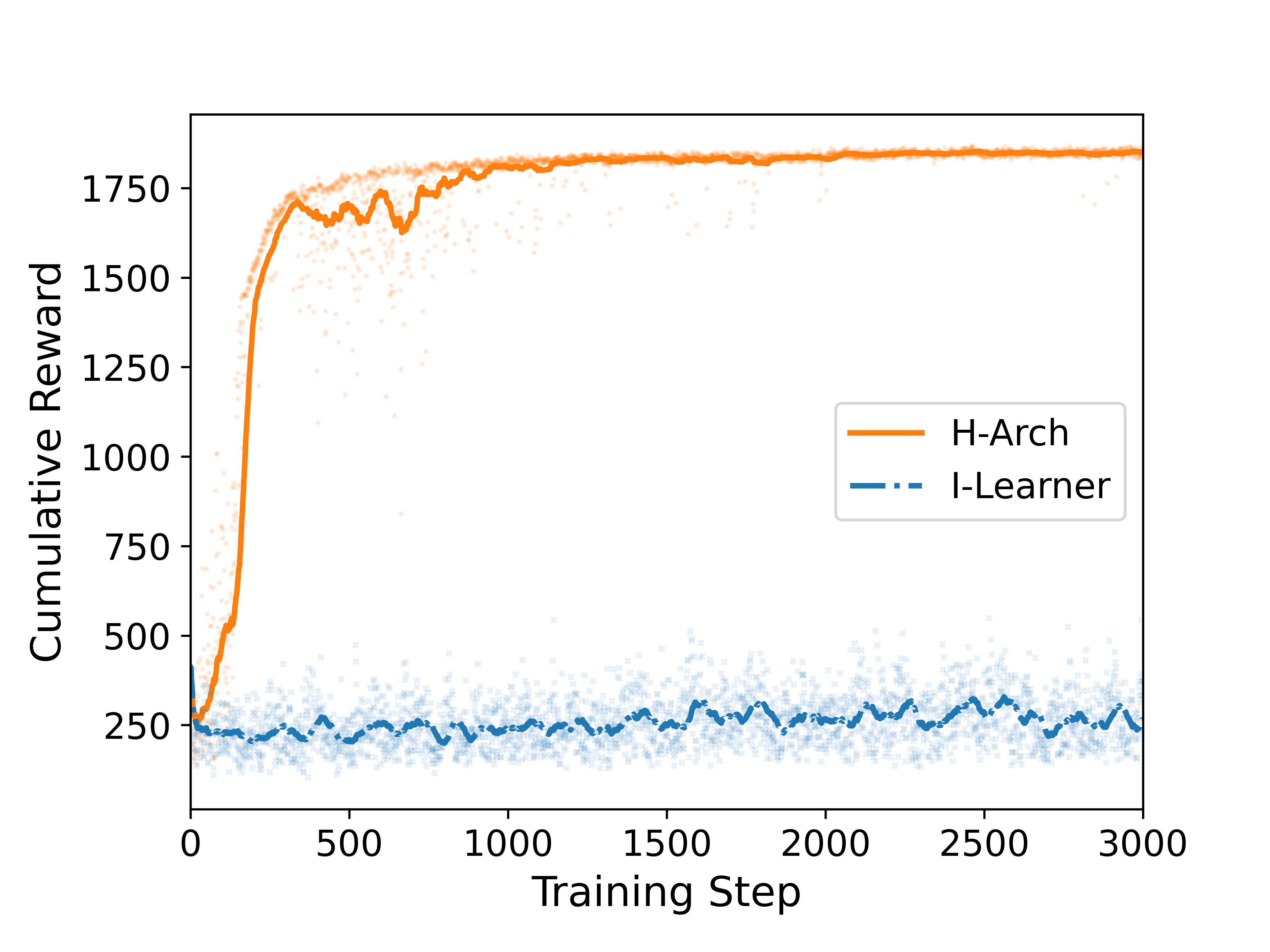}
            \caption{C-0.15 $\mathcal{R}$}
        \end{subfigure} 
        \begin{subfigure}{0.23\textwidth}
            \includegraphics[width=\linewidth]{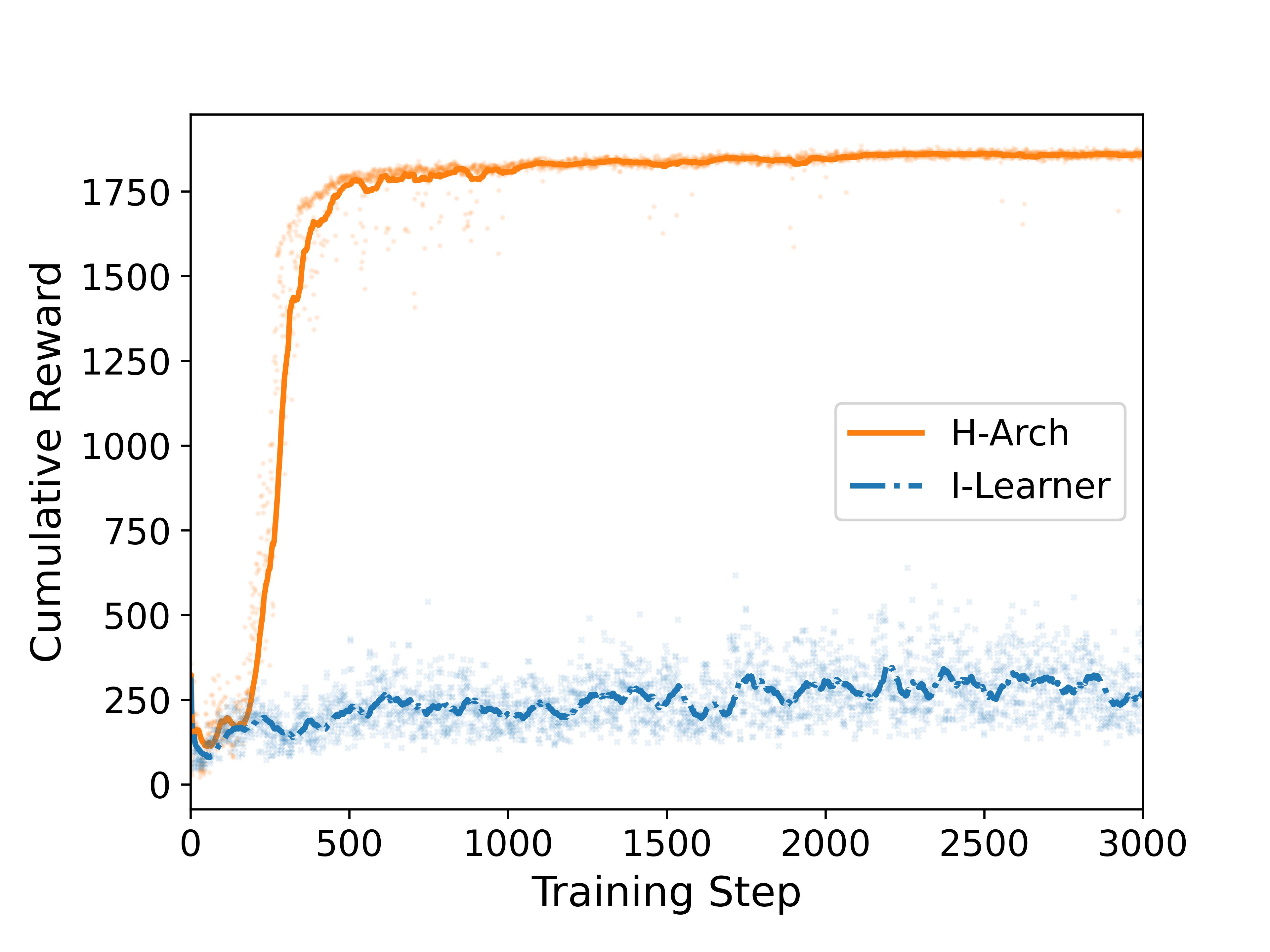}
            \caption{D-0.15 $\mathcal{R}$}
        \end{subfigure} 
        
        \hfill
        
        \begin{subfigure}{0.23\textwidth}
            \includegraphics[width=\linewidth]{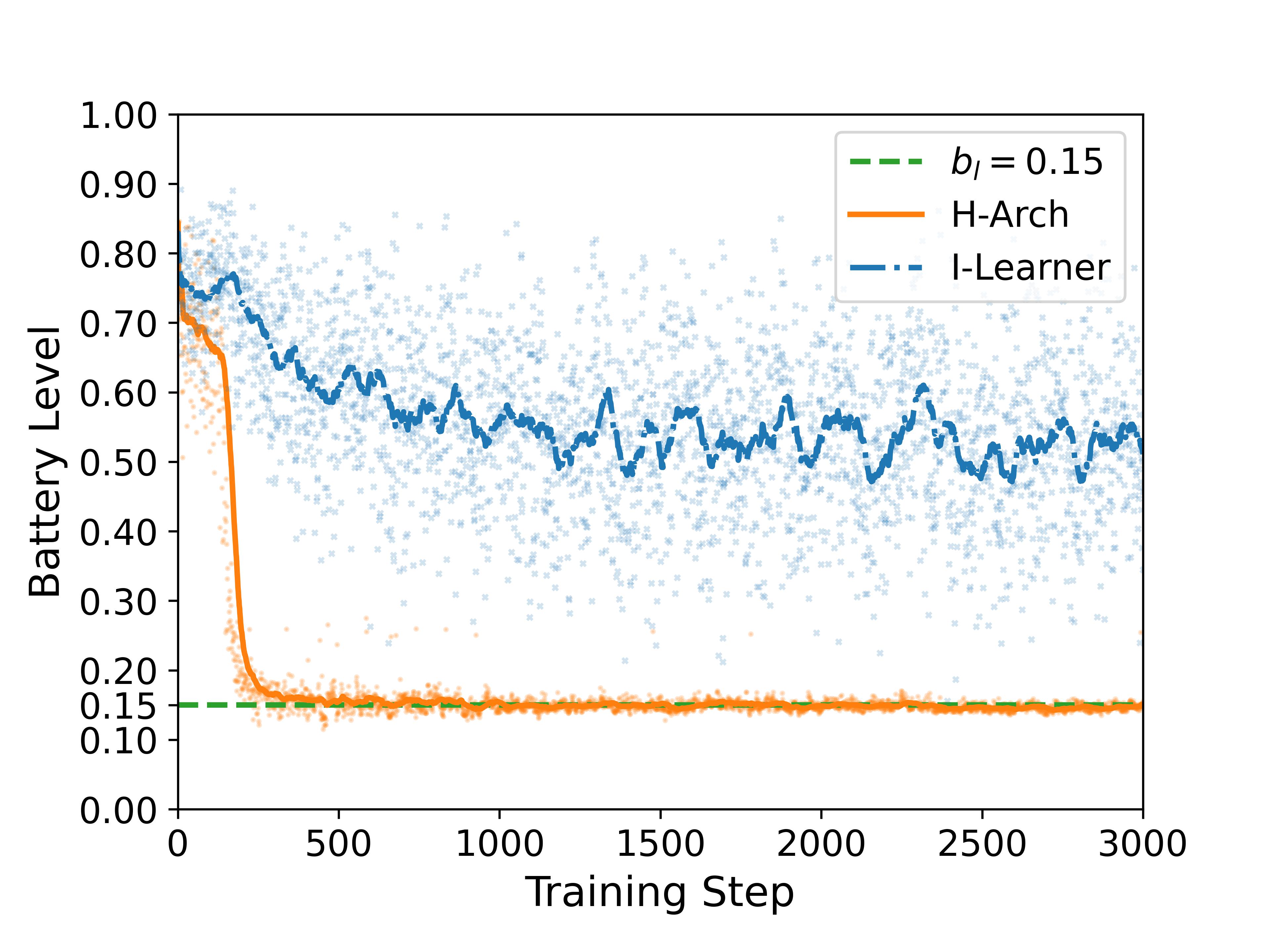}
            \caption{A-0.15 $\mathcal{B}$}
        \end{subfigure} 
        \begin{subfigure}{0.23\textwidth}
            \includegraphics[width=\linewidth]{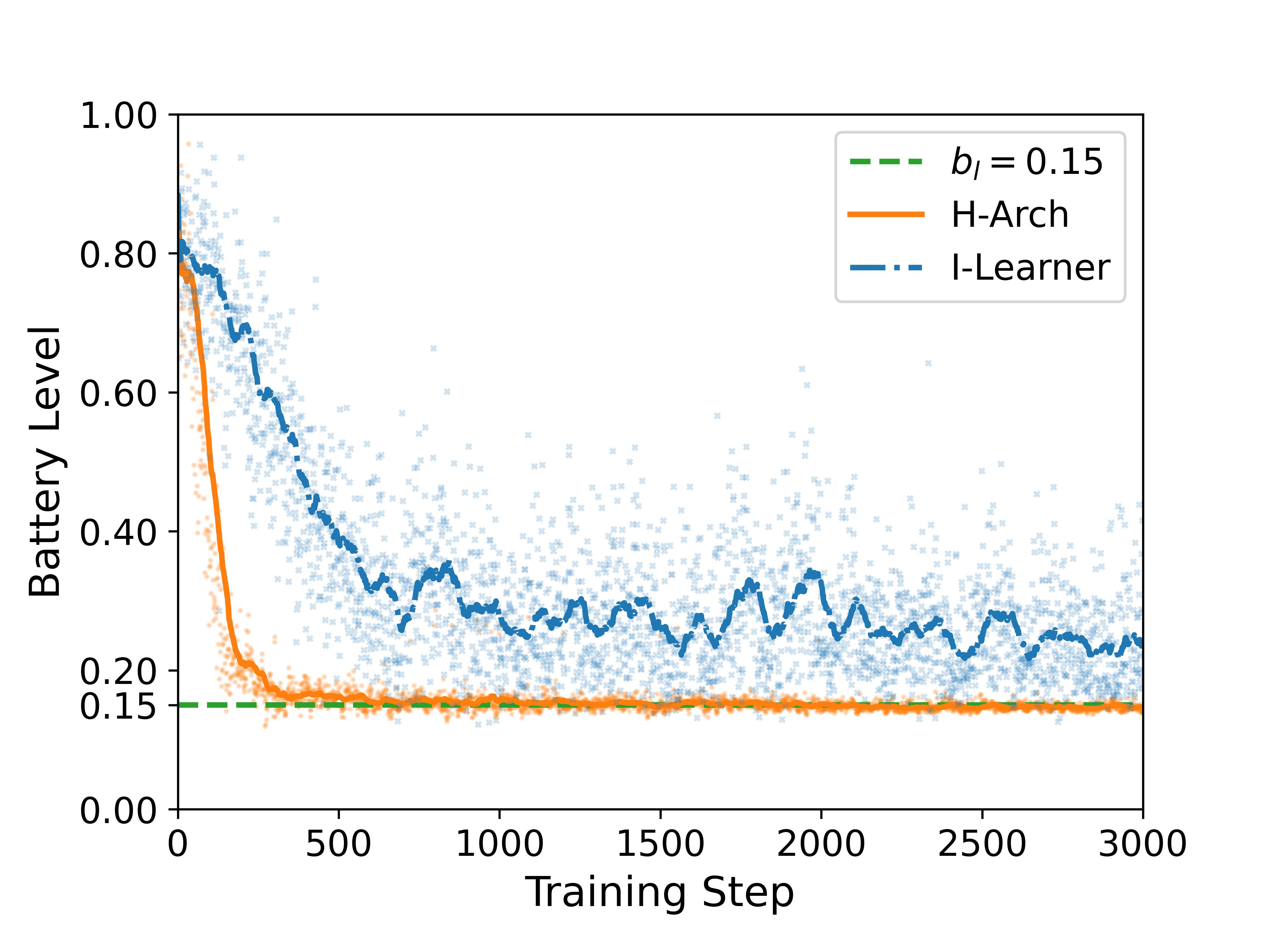}
            \caption{B-0.15 $\mathcal{B}$}
        \end{subfigure} 
        \begin{subfigure}{0.23\textwidth}
            \includegraphics[width=\linewidth]{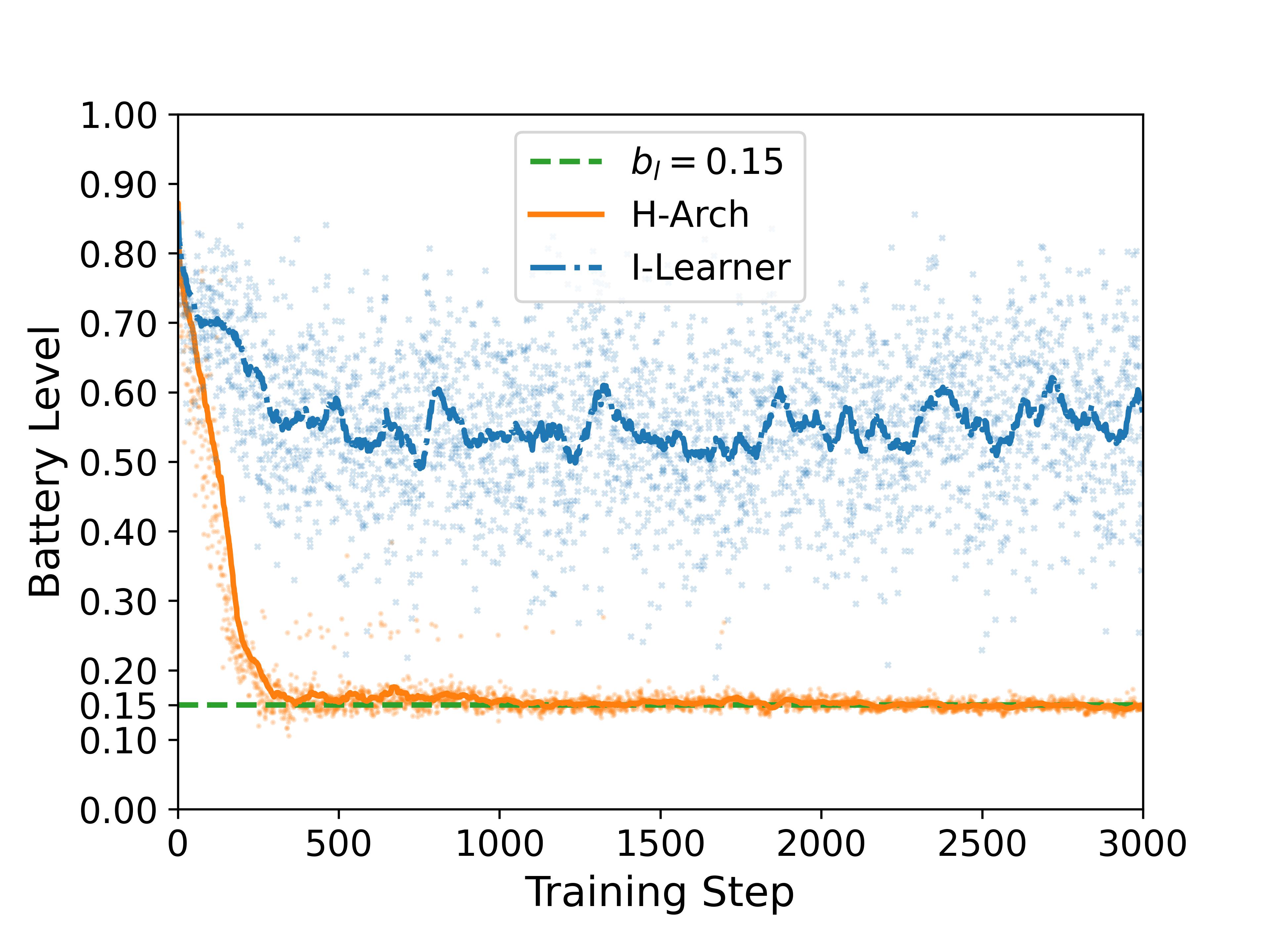}
            \caption{C-0.15 $\mathcal{B}$}
        \end{subfigure} 
        \begin{subfigure}{0.23\textwidth}
            \includegraphics[width=\linewidth]{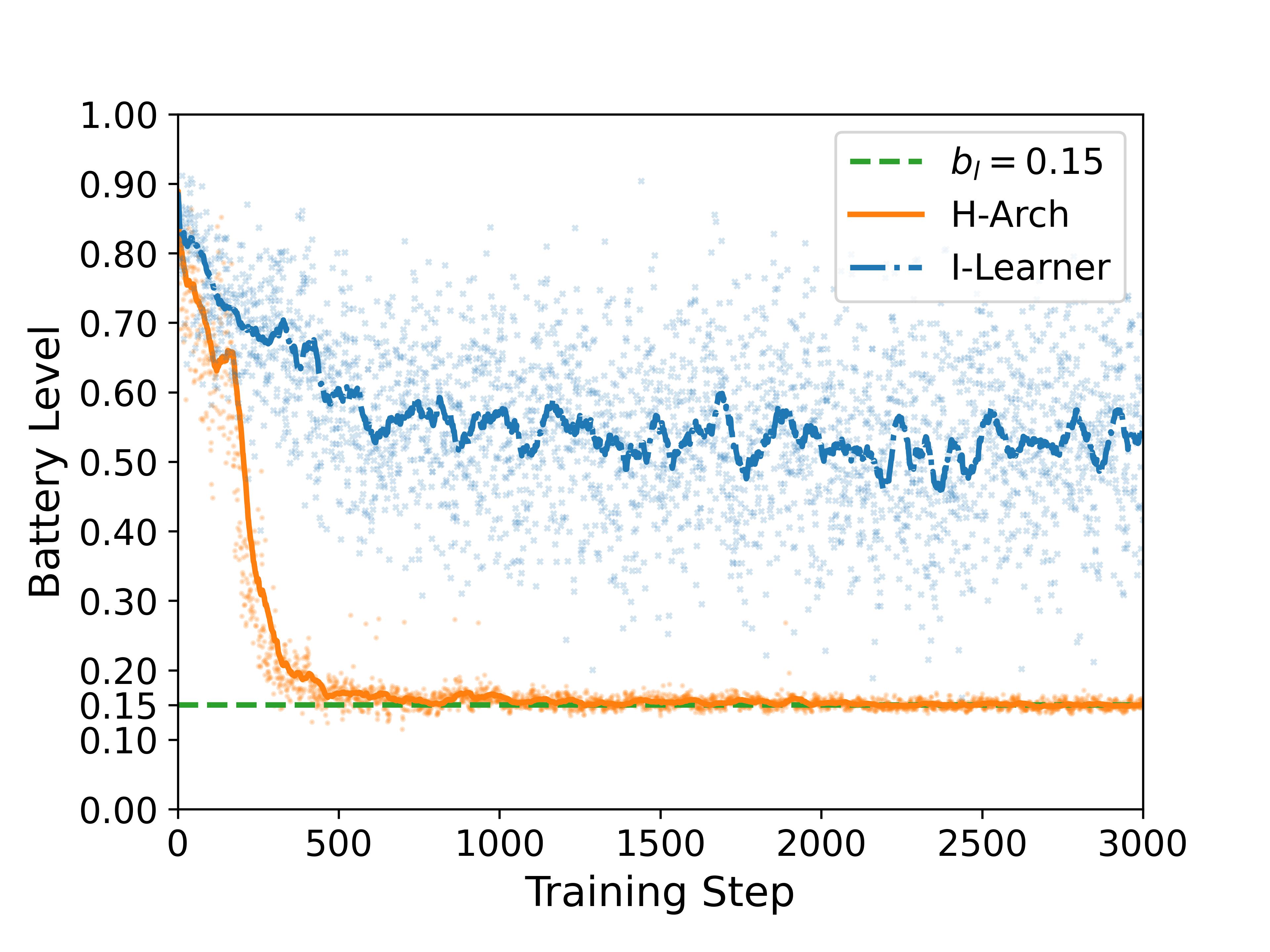}
            \caption{D-0.15 $\mathcal{B}$}
        \end{subfigure} 

\caption{Training result of proposed deep MARL-based models and models using individual learner approach on four maps with $b_l=0.15$}
\label{fig:training-r-0.15}
\end{figure}

\begin{figure} [htb]
        \centering
        \begin{subfigure}{0.23\textwidth}
            \includegraphics[width=\linewidth]{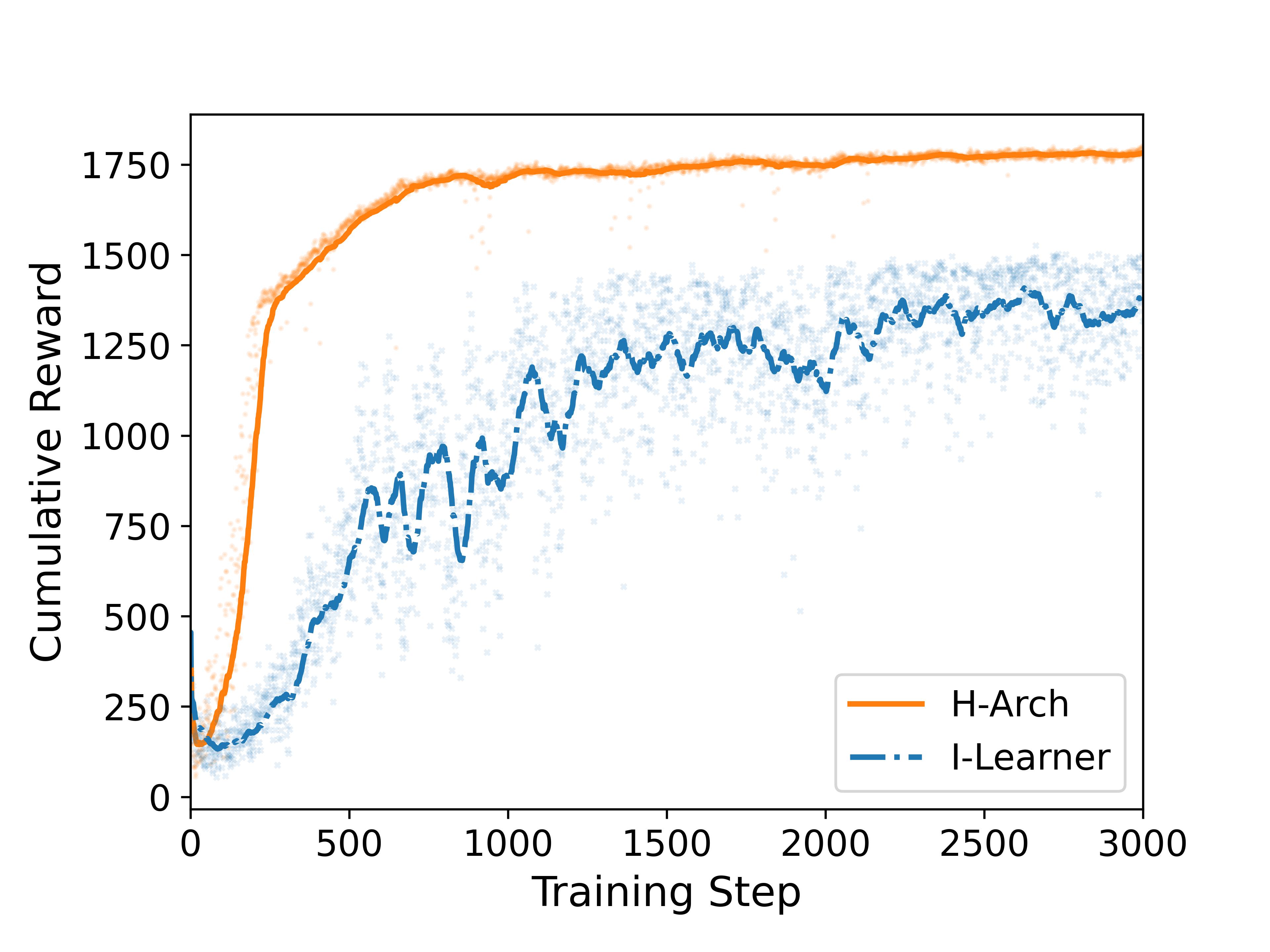}
            \caption{A-0.20 $\mathcal{R}$}
        \end{subfigure} 
        \begin{subfigure}{0.23\textwidth}
            \includegraphics[width=\linewidth]{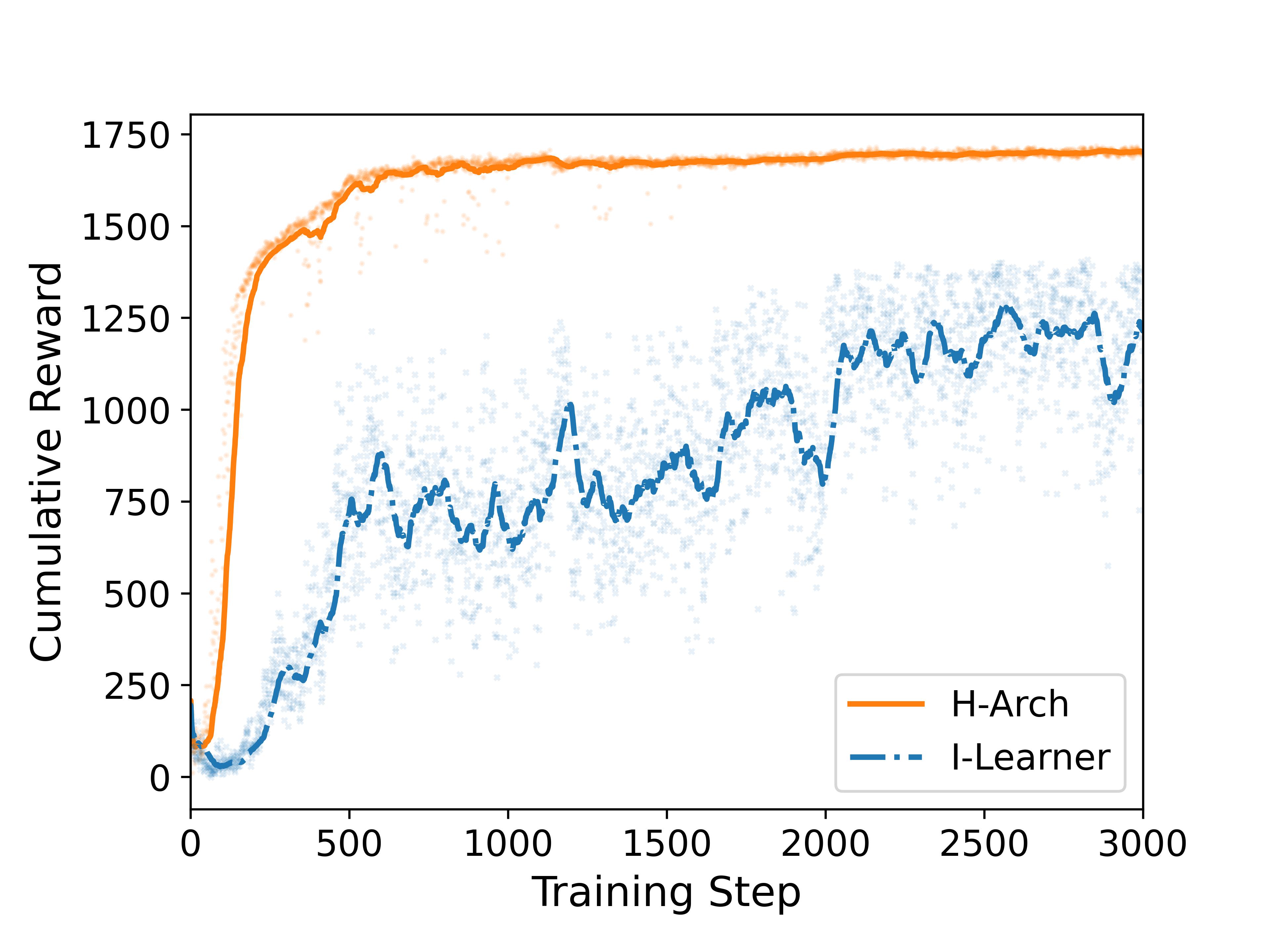}
            \caption{B-0.20 $\mathcal{R}$}
        \end{subfigure} 
        \begin{subfigure}{0.23\textwidth}
            \includegraphics[width=\linewidth]{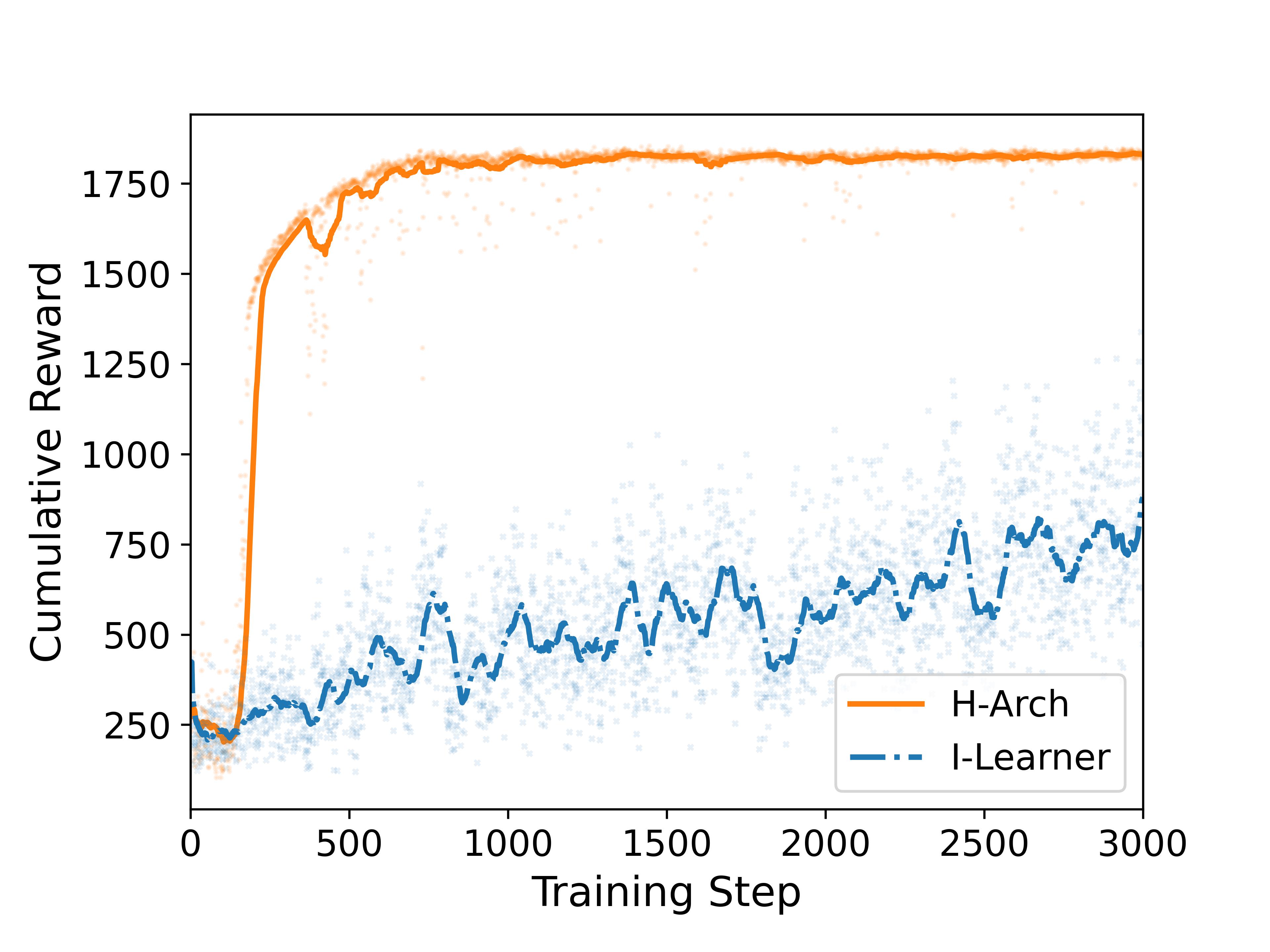}
            \caption{C-0.20 $\mathcal{R}$}
        \end{subfigure} 
        \begin{subfigure}{0.23\textwidth}
            \includegraphics[width=\linewidth]{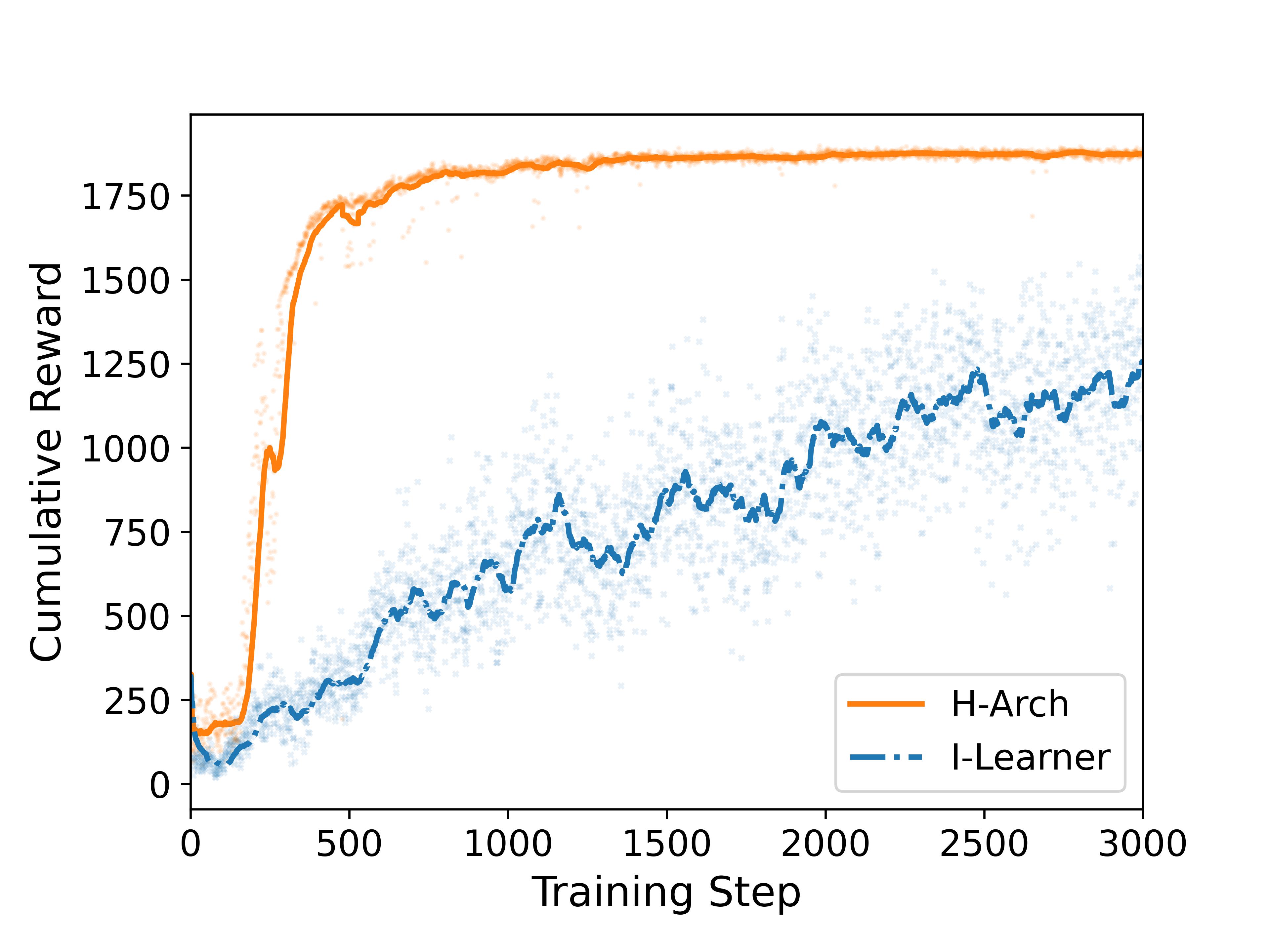}
            \caption{D-0.20 $\mathcal{R}$}
        \end{subfigure} 

        \hfill
        
        \begin{subfigure}{0.23\textwidth}
            \includegraphics[width=\linewidth]{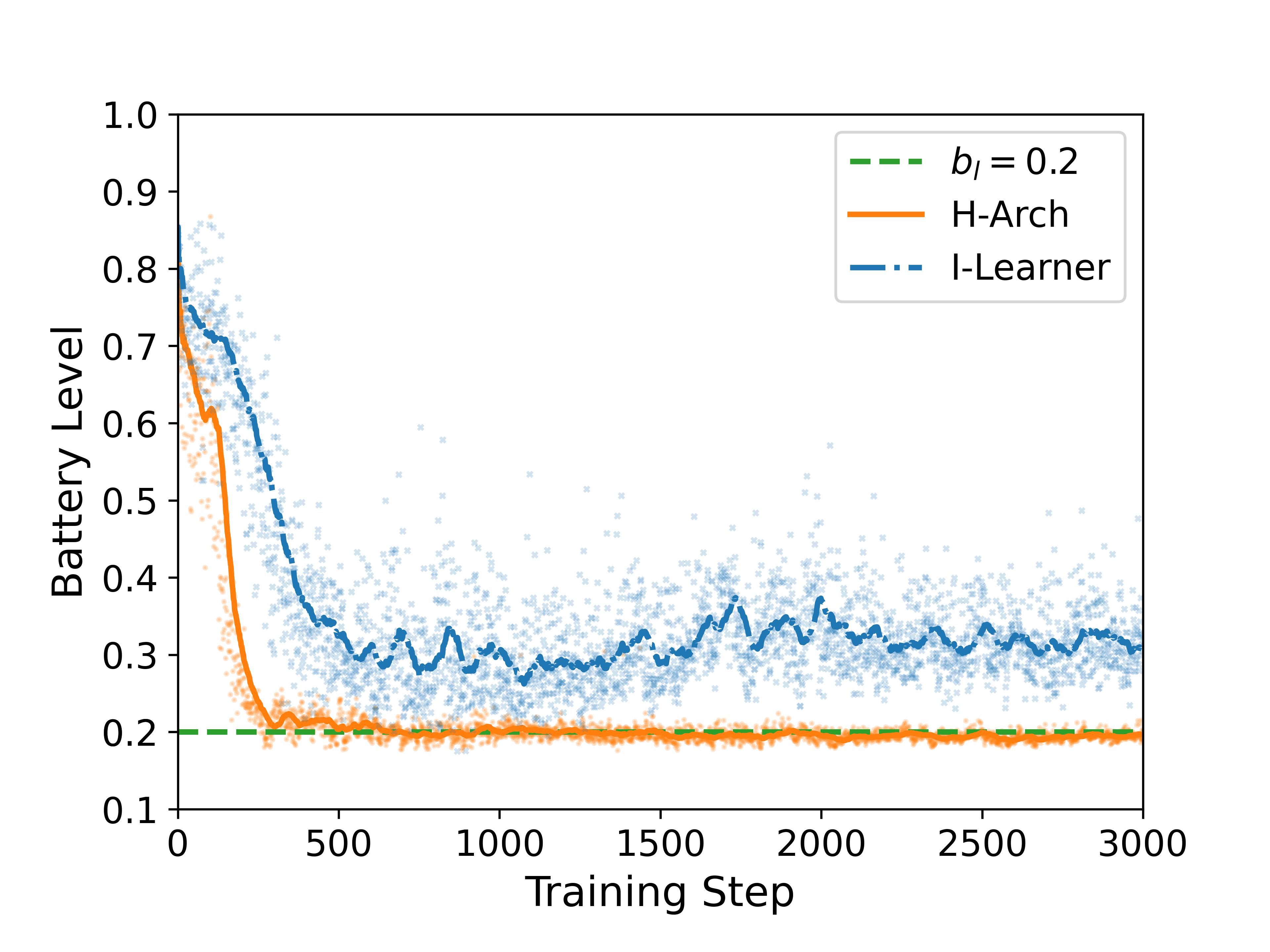}
            \caption{A-0.20 $\mathcal{B}$}
        \end{subfigure} 
        \begin{subfigure}{0.23\textwidth}
            \includegraphics[width=\linewidth]{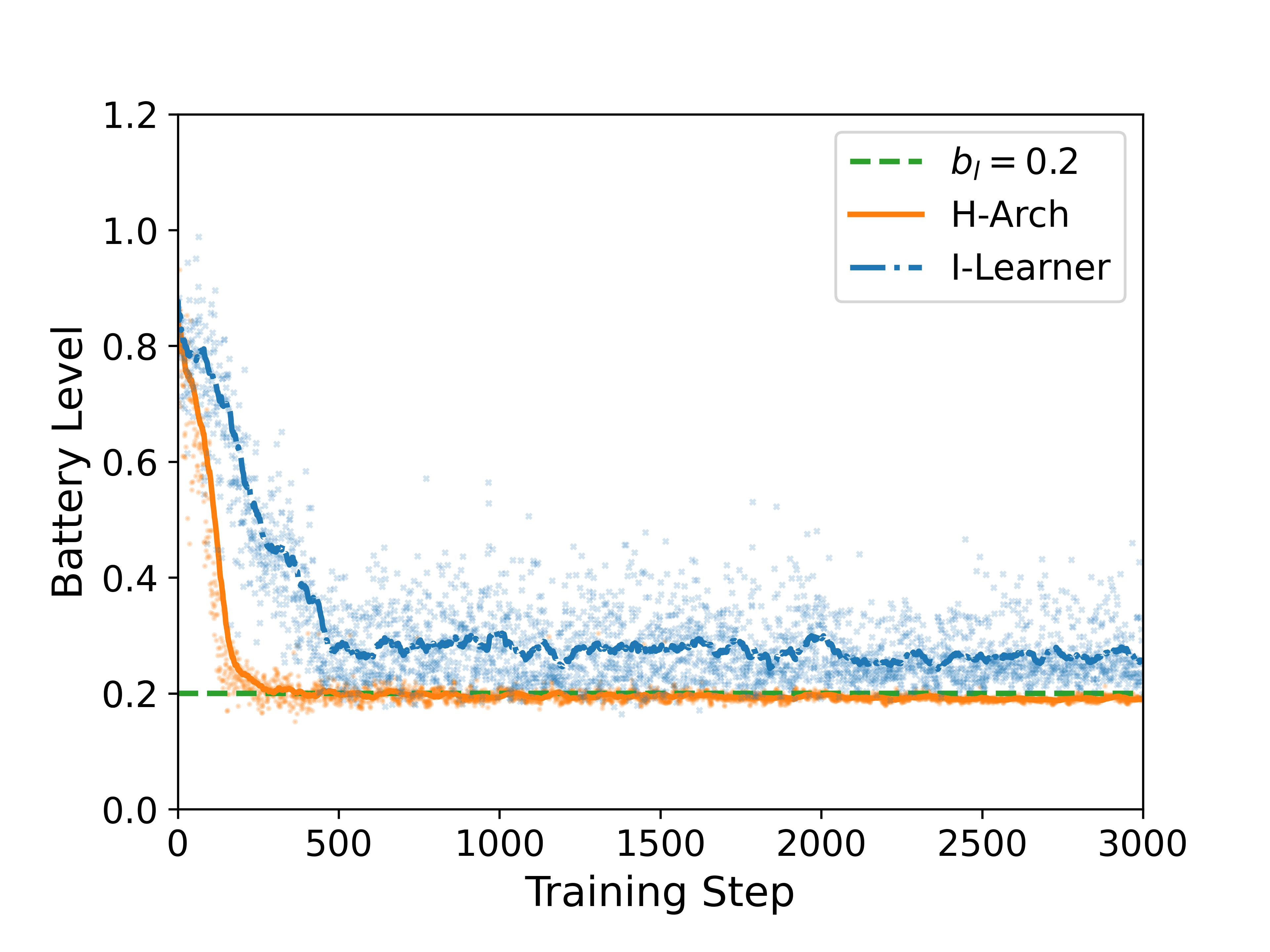}
            \caption{B-0.20 $\mathcal{B}$}
        \end{subfigure} 
        \begin{subfigure}{0.23\textwidth}
            \includegraphics[width=\linewidth]{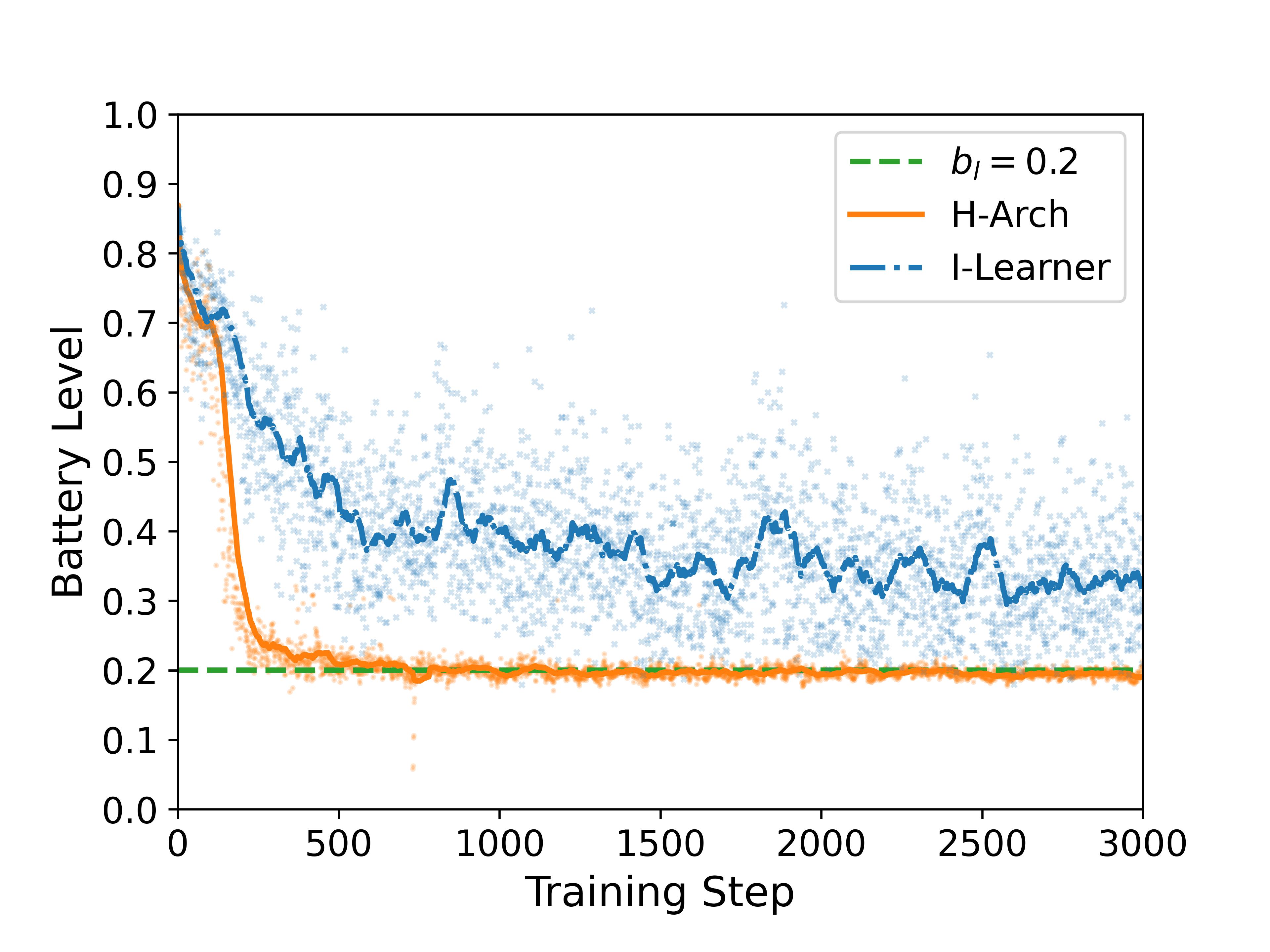}
            \caption{C-0.20 $\mathcal{B}$}
        \end{subfigure} 
        \begin{subfigure}{0.23\textwidth}
            \includegraphics[width=\linewidth]{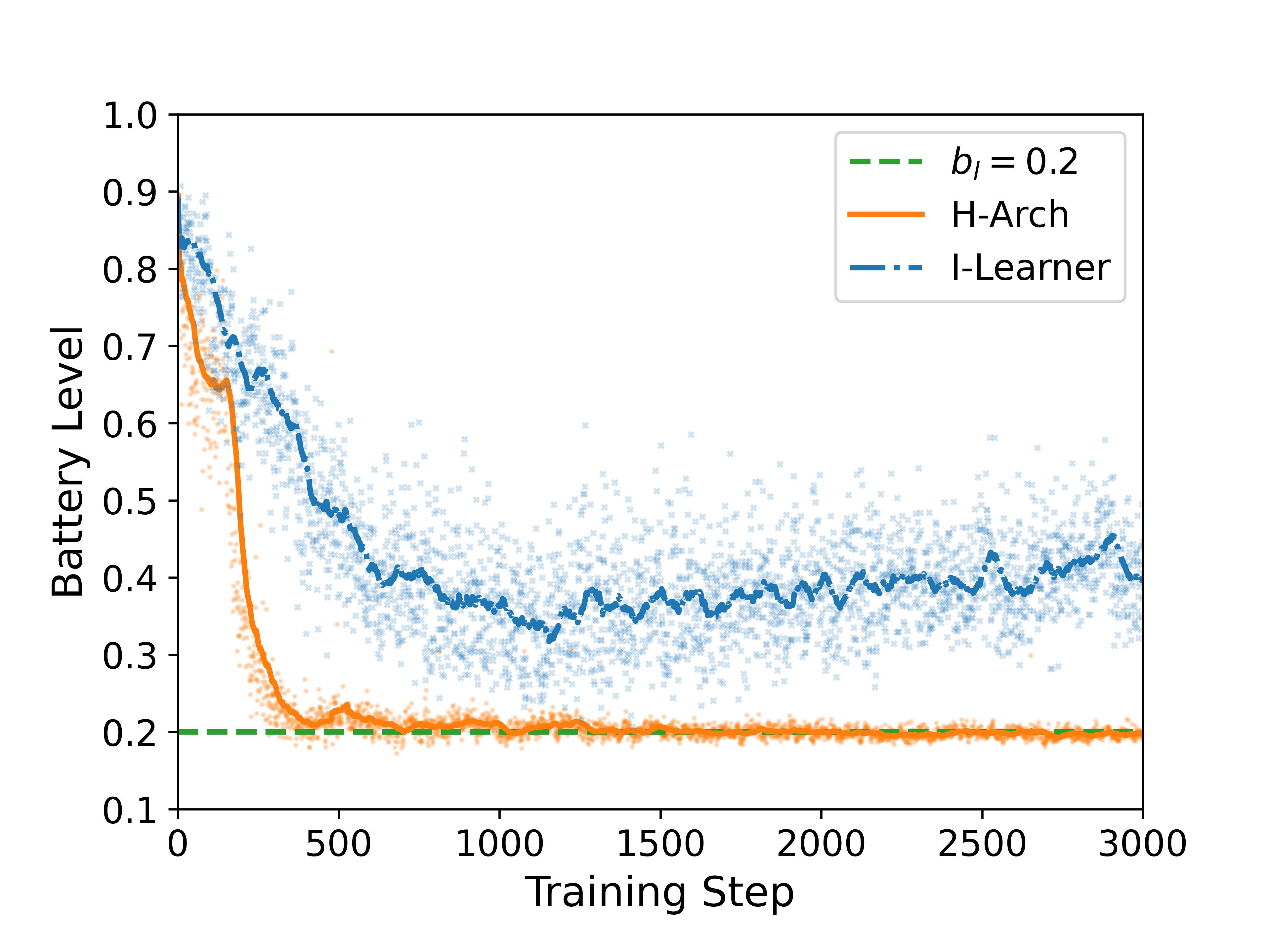}
            \caption{D-0.20 $\mathcal{B}$}
        \end{subfigure} 
    
\caption{Training result of proposed deep MARL-based models and models using individual learner approach on four maps with $b_l=0.2$}
\label{fig:training-r-0.2}
\end{figure}

\clearpage

\section{Battery recharging performance (agents' battery remaining when recharging and agents' battery failure rate) of models when $b_l=0.15$ and $b_l=0.2$} \label{app:B}

\begin{table}[htb]
    \sisetup{round-mode=places, round-precision=5}
    \centering
    \resizebox{0.45\textwidth}{!}{%
    \begin{tabular}{lllll}%
        MAP A, $b_l = 0.15$ & & & & \\
        \sisetup{round-mode=places, round-precision=5}
        \bfseries $n$ & \bfseries $b_c$ & \bfseries $\delta_{b_c}$ & \bfseries $F$ & \bfseries $\delta_{F}$
        \begin{filecontents*}{csv/A-0.15-battery.csv}
n,bc,bcE,f,fE
1,0.14428443,0.00561691,0.00033546,0.00047694
2,0.1463736,0.00556257,1.873e-05,5.924e-05
3,0.14790591,0.00442916,1.243e-05,3.93e-05
4,0.1488956,0.00355883,NAN,NAN
5,0.14965263,0.00323127,7.44e-06,2.354e-05
6,0.15014068,0.0026922,NAN,NAN
7,0.15058575,0.00237612,NAN,NAN
8,0.15091651,0.00196652,NAN,NAN
\end{filecontents*}
\csvreader[head to column names]{csv/A-0.15-battery.csv}{}% use head of csv as column names
        {\\\hline \n & \bc & \bcE & \f & \fE}% specify your coloumns here
    \end{tabular}
    }
    \resizebox{0.45\textwidth}{!}{%
    \begin{tabular}{lllll}%
        MAP B, $b_l = 0.15$ & & & & \\
        \sisetup{round-mode=places, round-precision=5}
        \bfseries $n$ & \bfseries $b_c$ & \bfseries $\delta_{b_c}$ & \bfseries $F$ & \bfseries $\delta_{F}$
        \begin{filecontents*}{csv/B-0.15-battery.csv}
n,bc,bcE,f,fE
1,0.13731706,0.00652479,0.00011394,0.00018346
2,0.14217063,0.00621512,NAN,NAN
3,0.14522715,0.00523774,NAN,NAN
4,0.14771809,0.00490473,NAN,NAN
5,0.1491448,0.00405235,NAN,NAN
6,0.1503471,0.00364652,3.3e-07,1.05e-06
7,0.15114583,0.00333696,2.24e-06,3.03e-06
8,0.15185933,0.00319985,NAN,NAN
\end{filecontents*}
\csvreader[head to column names]{csv/B-0.15-battery.csv}{}% use head of csv as column names
        {\\\hline \n & \bc & \bcE & \f & \fE}% specify your coloumns here
    \end{tabular}
    }
    \resizebox{0.45\textwidth}{!}{%
    \begin{tabular}{lllll}%
        MAP C, $b_l = 0.15$ & & & & \\
        \sisetup{round-mode=places, round-precision=5}
        \bfseries $n$ & \bfseries $b_c$ & \bfseries $\delta_{b_c}$ & \bfseries $F$ & \bfseries $\delta_{F}$
        \begin{filecontents*}{csv/C-0.15-battery.csv}
n,bc,bcE,f,fE
1,0.13781114,0.00610972,3.81e-05,0.00012047
2,0.14520451,0.00588039,NAN,NAN
3,0.14801488,0.00512705,NAN,NAN
4,0.15003441,0.00495857,NAN,NAN
5,0.15052294,0.00434752,NAN,NAN
6,0.15145571,0.00408324,NAN,NAN
7,0.15136983,0.0037477,NAN,NAN
8,0.15181206,0.00356054,NAN,NAN
\end{filecontents*}
\csvreader[head to column names]{csv/C-0.15-battery.csv}{}% use head of csv as column names
        {\\\hline \n & \bc & \bcE & \f & \fE}% specify your coloumns here
    \end{tabular}
    }
    \resizebox{0.45\textwidth}{!}{%
    \begin{tabular}{lllll}%
        MAP D, $b_l = 0.15$ & & & & \\
        \sisetup{round-mode=places, round-precision=5}
        \bfseries $n$ & \bfseries $b_c$ & \bfseries $\delta_{b_c}$ & \bfseries $F$ & \bfseries $\delta_{F}$
        \begin{filecontents*}{csv/D-0.15-battery.csv}
n,bc,bcE,f,fE
1,0.14393187,0.00436944,NAN,NAN
2,0.14866106,0.00525744,7.448e-05,0.0001301
3,0.15254658,0.00514717,NAN,NAN
4,0.15511891,0.0048661,9.28e-06,2.935e-05
5,0.15678917,0.00433544,NAN,NAN
6,0.15753978,0.00401064,NAN,NAN
7,0.15837369,0.00402351,NAN,NAN
8,0.15887332,0.00359275,NAN,NAN
\end{filecontents*}
\csvreader[head to column names]{csv/D-0.15-battery.csv}{}% use head of csv as column names
        {\\\hline \n & \bc & \bcE & \f & \fE}% specify your coloumns here
    \end{tabular}
    }
    
    \caption{Model A/B/C/D-$0.15$'s battery recharging performance when running with $1$ to $8$ number of patrolling agents.}
    \label{tab:bf-ABCD-0.15}
\end{table}

\begin{table}[htb]
    \sisetup{round-mode=places, round-precision=5}
    \centering
    \resizebox{0.45\textwidth}{!}{%
    \begin{tabular}{lllll}%
        MAP A, $b_l = 0.2$ & & & & \\
        \sisetup{round-mode=places, round-precision=5}
        \bfseries $n$ & \bfseries $b_c$ & \bfseries $\delta_{b_c}$ & \bfseries $F$ & \bfseries $\delta_{F}$
        \begin{filecontents*}{csv/A-0.20-battery.csv}
n,bc,bcE,f,fE
1,0.18669421,0.00487885,NAN,NAN
2,0.19442098,0.00559147,NAN,NAN
3,0.19820384,0.00438953,NAN,NAN
4,0.20148169,0.00421036,NAN,NAN
5,0.20355802,0.0035248,NAN,NAN
6,0.20495822,0.00320135,NAN,NAN
7,0.20337918,0.00370833,NAN,NAN
8,0.2074599,0.00310022,NAN,NAN
\end{filecontents*}
\csvreader[head to column names]{csv/A-0.20-battery.csv}{}% use head of csv as column names
        {\\\hline \n & \bc & \bcE & \f & \fE}% specify your coloumns here
    \end{tabular}
    }
    \resizebox{0.45\textwidth}{!}{%
    \begin{tabular}{lllll}%
        MAP B, $b_l = 0.2$ & & & & \\
        \sisetup{round-mode=places, round-precision=5}
        \bfseries $n$ & \bfseries $b_c$ & \bfseries $\delta_{b_c}$ & \bfseries $F$ & \bfseries $\delta_{F}$
        \begin{filecontents*}{csv/B-0.20-battery.csv}
n,bc,bcE,f,fE
1,0.18653785,0.00375892,NAN,NAN
2,0.19317545,0.00416678,NAN,NAN
3,0.19792613,0.00349916,NAN,NAN
4,0.201145,0.00311074,NAN,NAN
5,0.20341452,0.00292715,NAN,NAN
6,0.20500273,0.00293716,NAN,NAN
7,0.20593749,0.00236546,NAN,NAN
8,0.20679314,0.00235254,NAN,NAN
\end{filecontents*}
\csvreader[head to column names]{csv/B-0.20-battery.csv}{}% use head of csv as column names
        {\\\hline \n & \bc & \bcE & \f & \fE}% specify your coloumns here
    \end{tabular}
    }
    \resizebox{0.45\textwidth}{!}{%
    \begin{tabular}{lllll}%
        MAP C, $b_l = 0.2$ & & & & \\
        \sisetup{round-mode=places, round-precision=5}
        \bfseries $n$ & \bfseries $b_c$ & \bfseries $\delta_{b_c}$ & \bfseries $F$ & \bfseries $\delta_{F}$
        \begin{filecontents*}{csv/C-0.20-battery.csv}
n,bc,bcE,f,fE
1,0.1920172,0.00465746,0.00014245,0.0001839
2,0.19602972,0.00439958,3.553e-05,7.49e-05
3,0.19888329,0.00391839,2.368e-05,4.993e-05
4,0.20086008,0.00342715,8.9e-06,2.814e-05
5,0.20197806,0.0033666,1.418e-05,2.989e-05
6,0.20274285,0.00301271,1.375e-05,1.145e-05
7,0.20299493,0.00267754,5.45e-06,4.11e-06
8,0.20307488,0.00257913,5.94e-06,6.4e-06
\end{filecontents*}
\csvreader[head to column names]{csv/C-0.20-battery.csv}{}% use head of csv as column names
        {\\\hline \n & \bc & \bcE & \f & \fE}% specify your coloumns here
    \end{tabular}
    }
    \resizebox{0.45\textwidth}{!}{%
    \begin{tabular}{lllll}%
        MAP D, $b_l = 0.2$ & & & & \\
        \sisetup{round-mode=places, round-precision=5}
        \bfseries $n$ & \bfseries $b_c$ & \bfseries $\delta_{b_c}$ & \bfseries $F$ & \bfseries $\delta_{F}$
        \begin{filecontents*}{csv/D-0.20-battery.csv}
n,bc,bcE,f,fE
1,0.1868343,0.0191314,7.152e-05,0.00015078
2,0.19494794,0.0051707,NAN,NAN
3,0.19914397,0.00491285,NAN,NAN
4,0.20139663,0.00438635,NAN,NAN
5,0.2024746,0.00428778,NAN,NAN
6,0.20293806,0.00382805,NAN,NAN
7,0.20337918,0.00370833,NAN,NAN
8,0.20329535,0.00347145,NAN,NAN
\end{filecontents*}
\csvreader[head to column names]{csv/D-0.20-battery.csv}{}% use head of csv as column names
        {\\\hline \n & \bc & \bcE & \f & \fE}% specify your coloumns here
    \end{tabular}
    }
    
    \caption{Model A/B/C/D-$0.2$'s battery recharging performance when running with $1$ to $8$ number of patrolling agents.}
    \label{tab:bf-ABCD-0.2}
\end{table}

\clearpage

\section{Patrolling performance result (evaluated by $AVG^h(G)$ and $\overline{MAX^h(G)}$) of proposed deep MARL-based models and CR strategy on four maps with $b_l=0.15$ and $b_l=0.2$} \label{app:C}

\begin{figure}[htb]
        \begin{subfigure}{0.48\textwidth}
            \includegraphics[width=\linewidth]{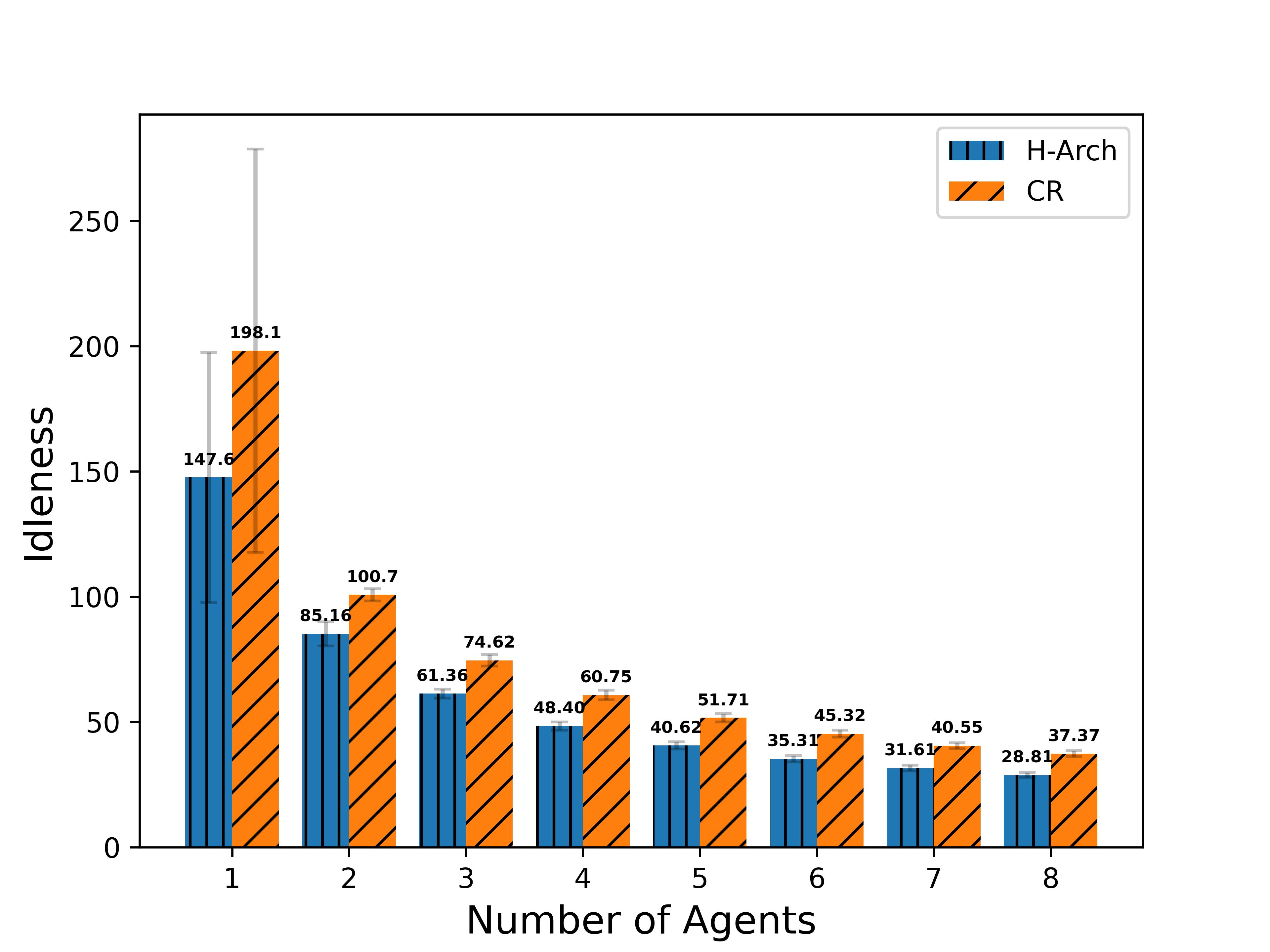}
            \caption{A-0.15 $\overline{MAX^h(G)}$}
        \end{subfigure} \hfill
        \begin{subfigure}{0.48\textwidth}
            \includegraphics[width=\linewidth]{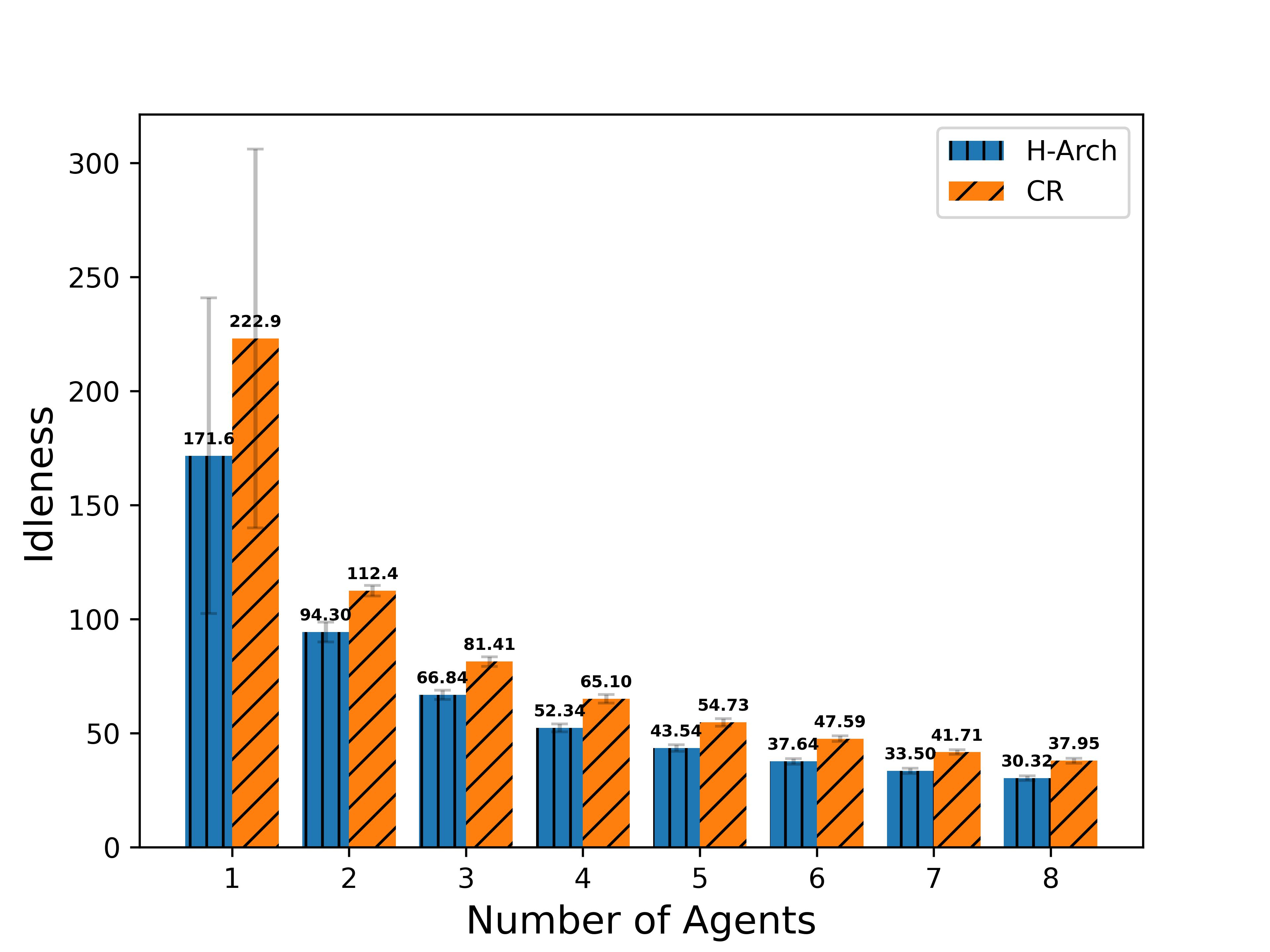}
            \caption{B-0.15 $\overline{MAX^h(G)}$}
        \end{subfigure} \hfill
        
        \begin{subfigure}{0.48\textwidth}
            \includegraphics[width=\linewidth]{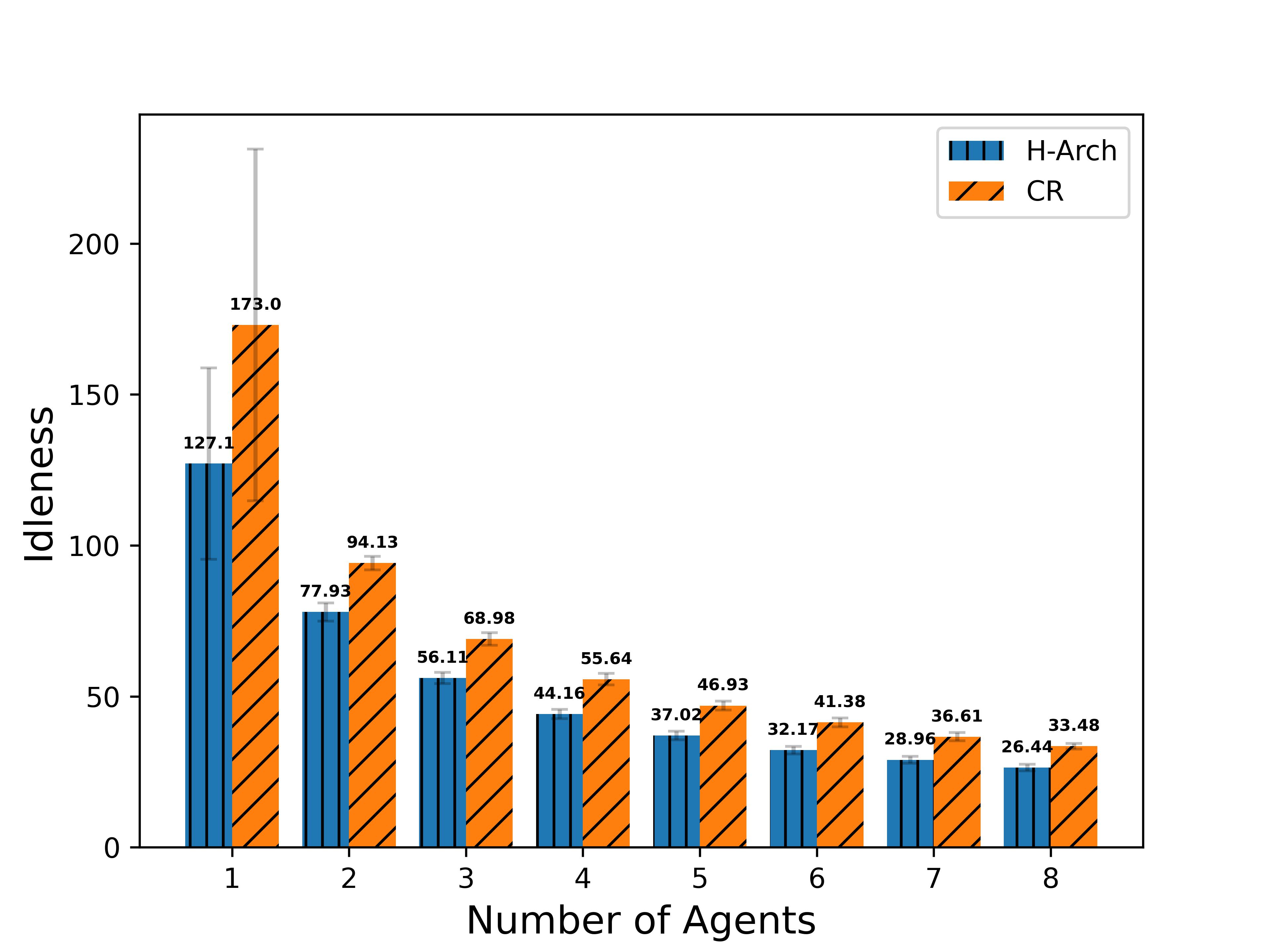}
            \caption{C-0.15 $\overline{MAX^h(G)}$}
        \end{subfigure} \hfill
        \begin{subfigure}{0.48\textwidth}
            \includegraphics[width=\linewidth]{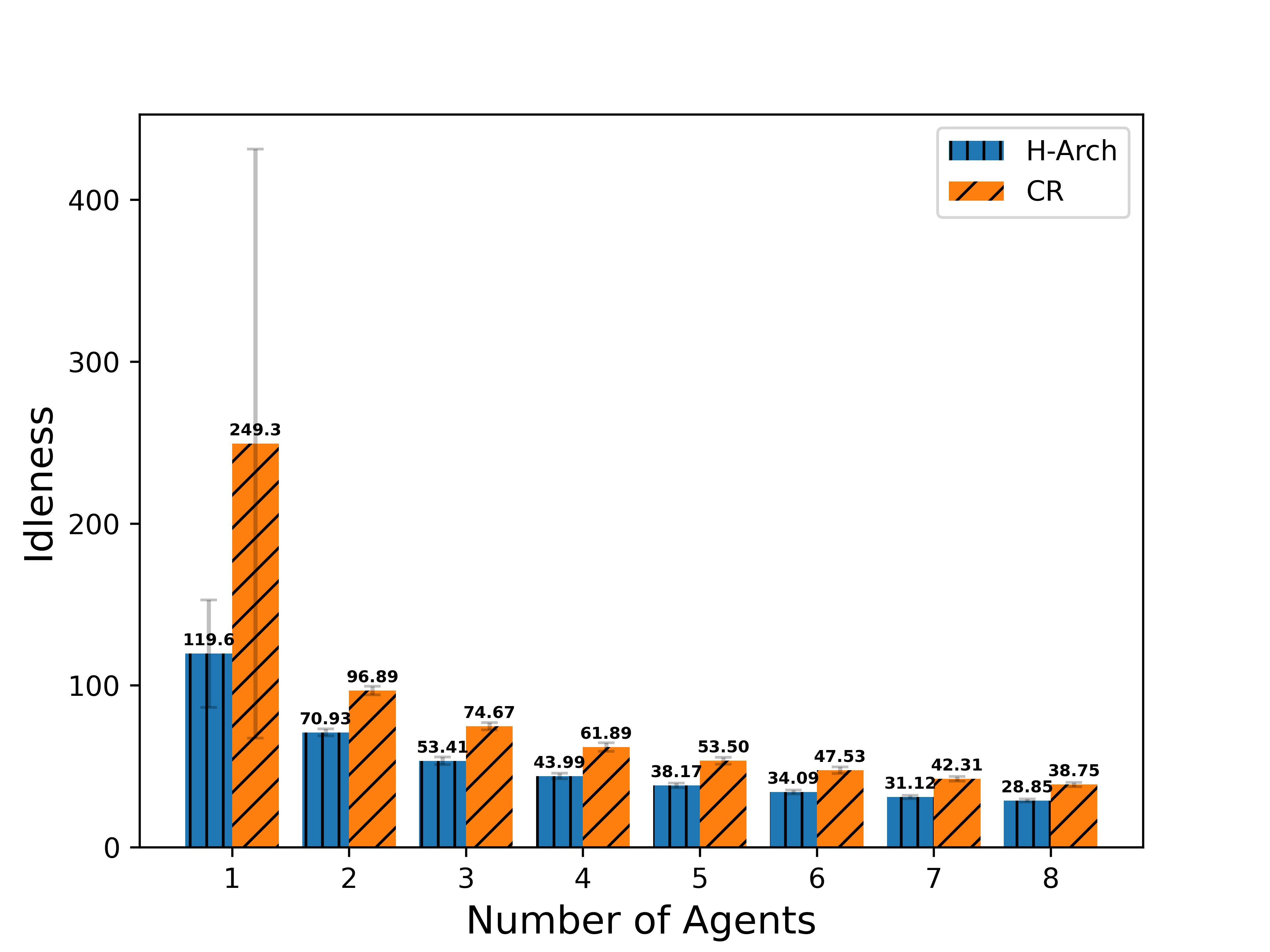}
            \caption{D-0.15 $\overline{MAX^h(G)}$}
        \end{subfigure} \hfill
        
\caption{Patrolling performance result ($\overline{MAX^h(G)}$) of proposed deep MARL-based models and CR strategy on four maps with $b_l=0.15$}
\label{fig:pp-eval-avg-0.15}
\end{figure}

\begin{figure}[htb]
        \begin{subfigure}{0.49\textwidth}
            \includegraphics[width=\linewidth]{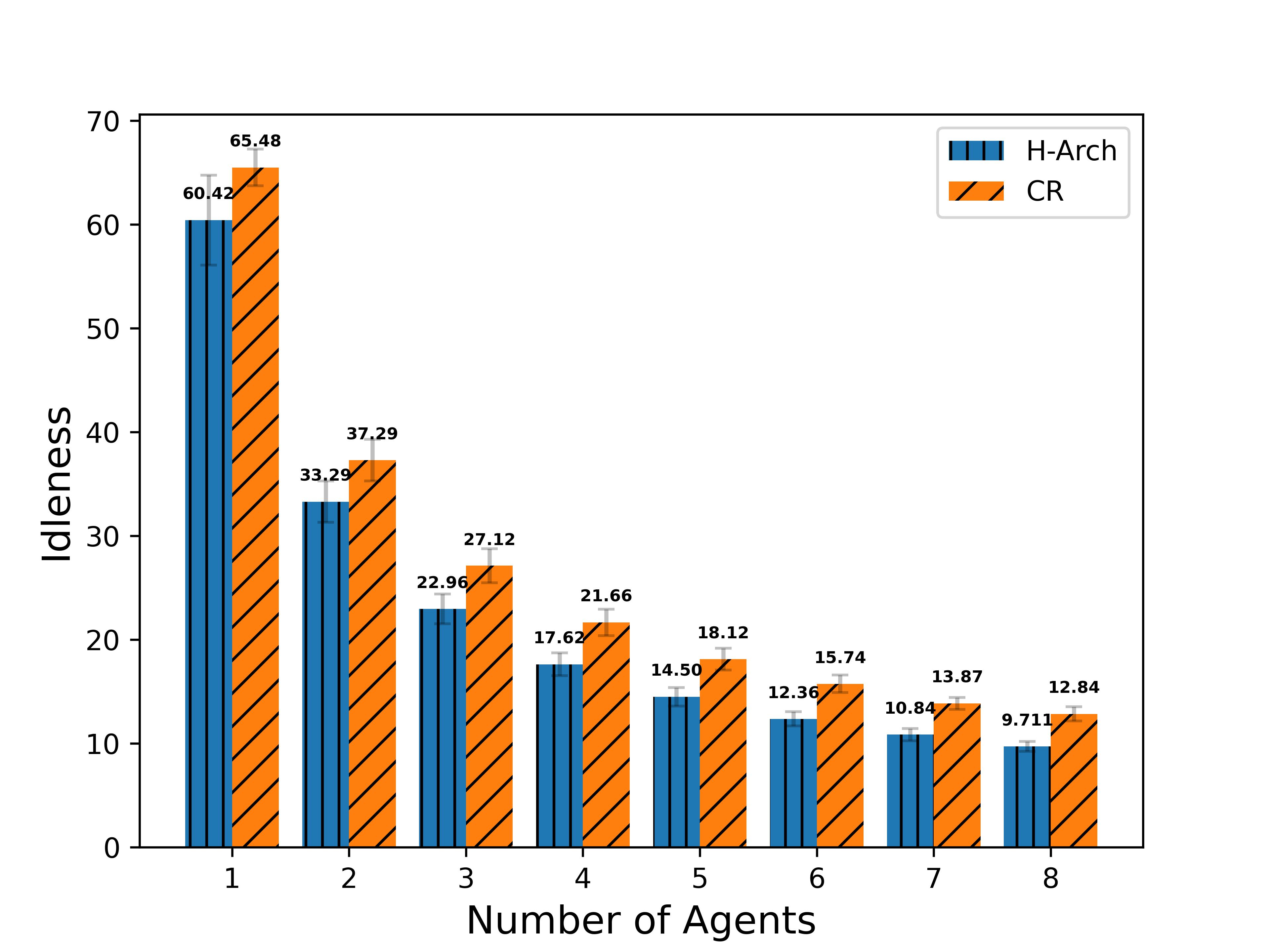}
            \caption{A-0.15 $AVG^h(G)$}
        \end{subfigure} \hfill
        \begin{subfigure}{0.49\textwidth}
            \includegraphics[width=\linewidth]{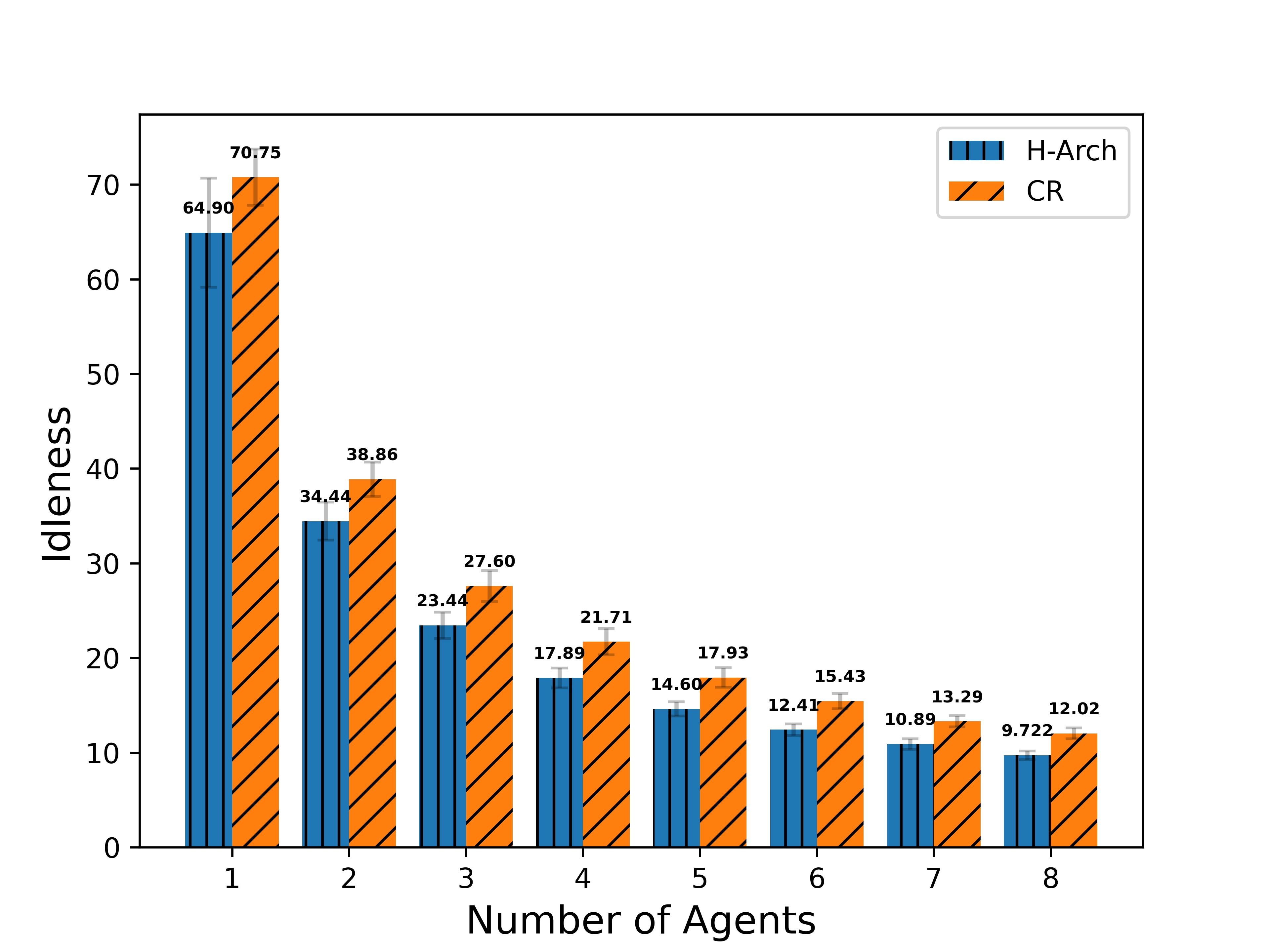}
            \caption{B-0.15 $AVG^h(G)$}
        \end{subfigure} \hfill
        
        \begin{subfigure}{0.49\textwidth}
            \includegraphics[width=\linewidth]{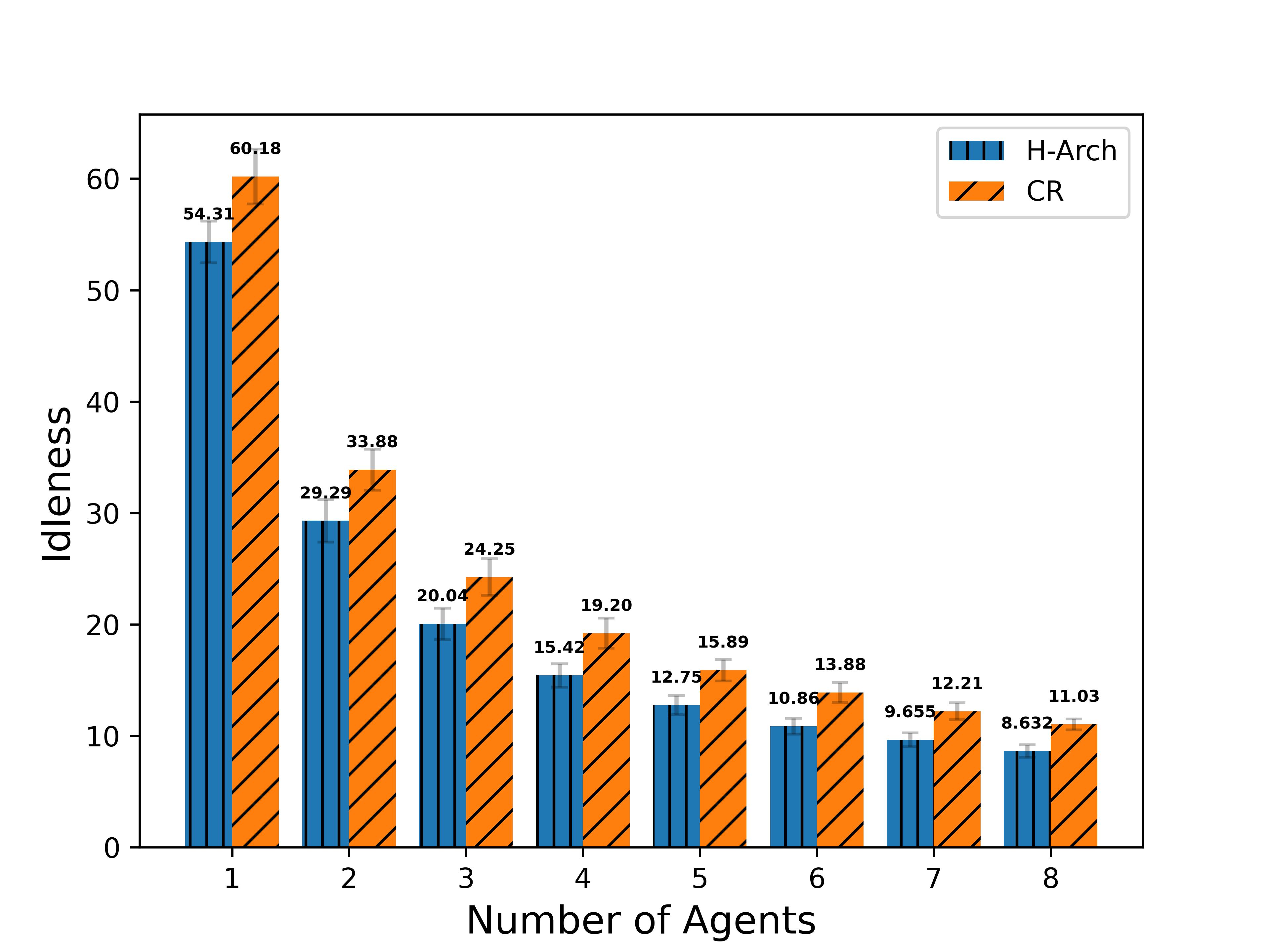}
            \caption{C-0.15 $AVG^h(G)$}
        \end{subfigure} \hfill
        \begin{subfigure}{0.49\textwidth}
            \includegraphics[width=\linewidth]{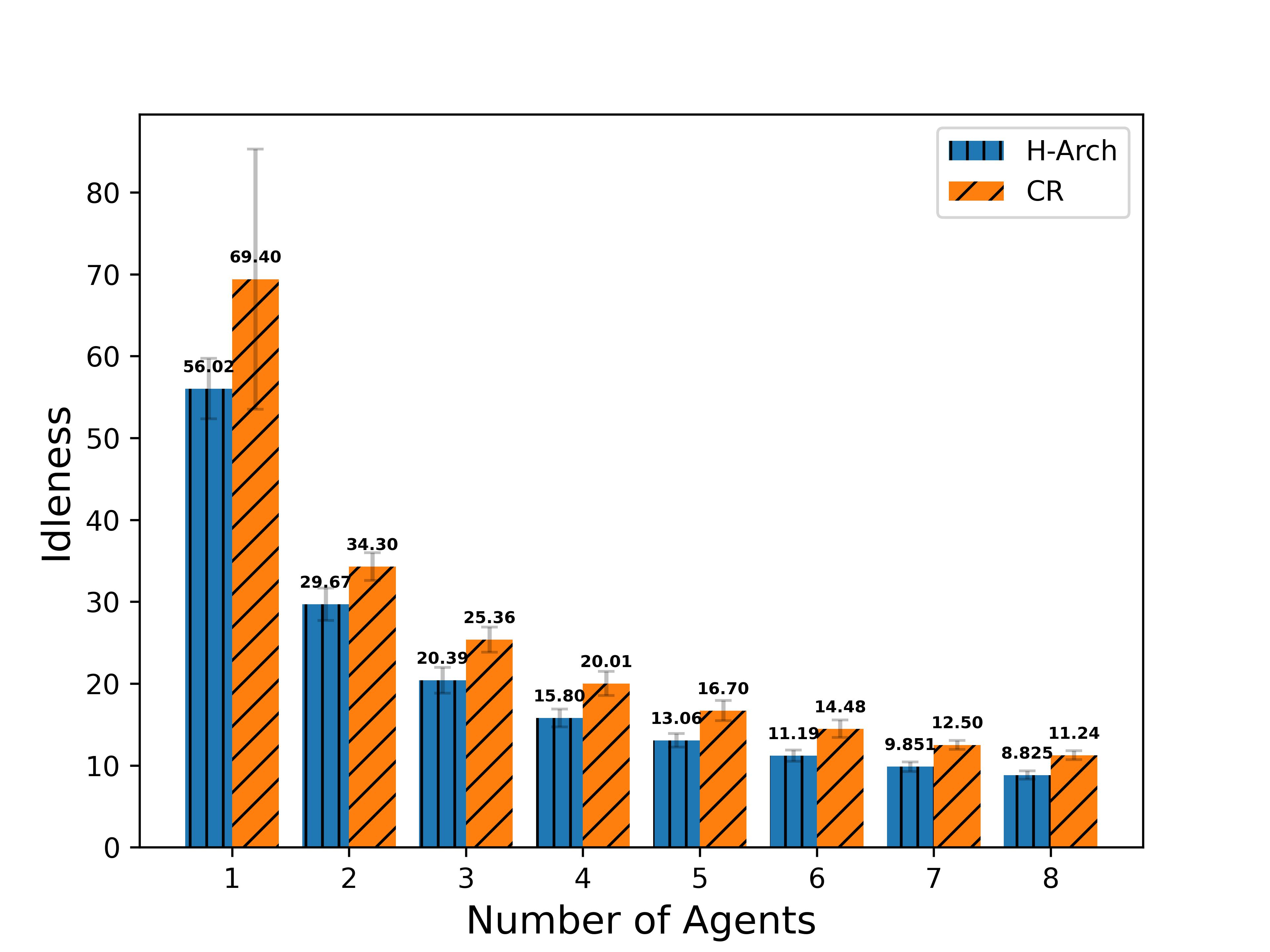}
            \caption{D-0.15 $AVG^h(G)$}
        \end{subfigure} \hfill

\caption{Patrolling performance result ($AVG^h(G)$) of proposed deep MARL-based models and CR strategy on four maps with $b_l=0.15$}
\label{fig:pp-eval-max-0.15}
\end{figure}

\begin{figure}[htb]

        \begin{subfigure}{0.49\textwidth}
            \includegraphics[width=\linewidth]{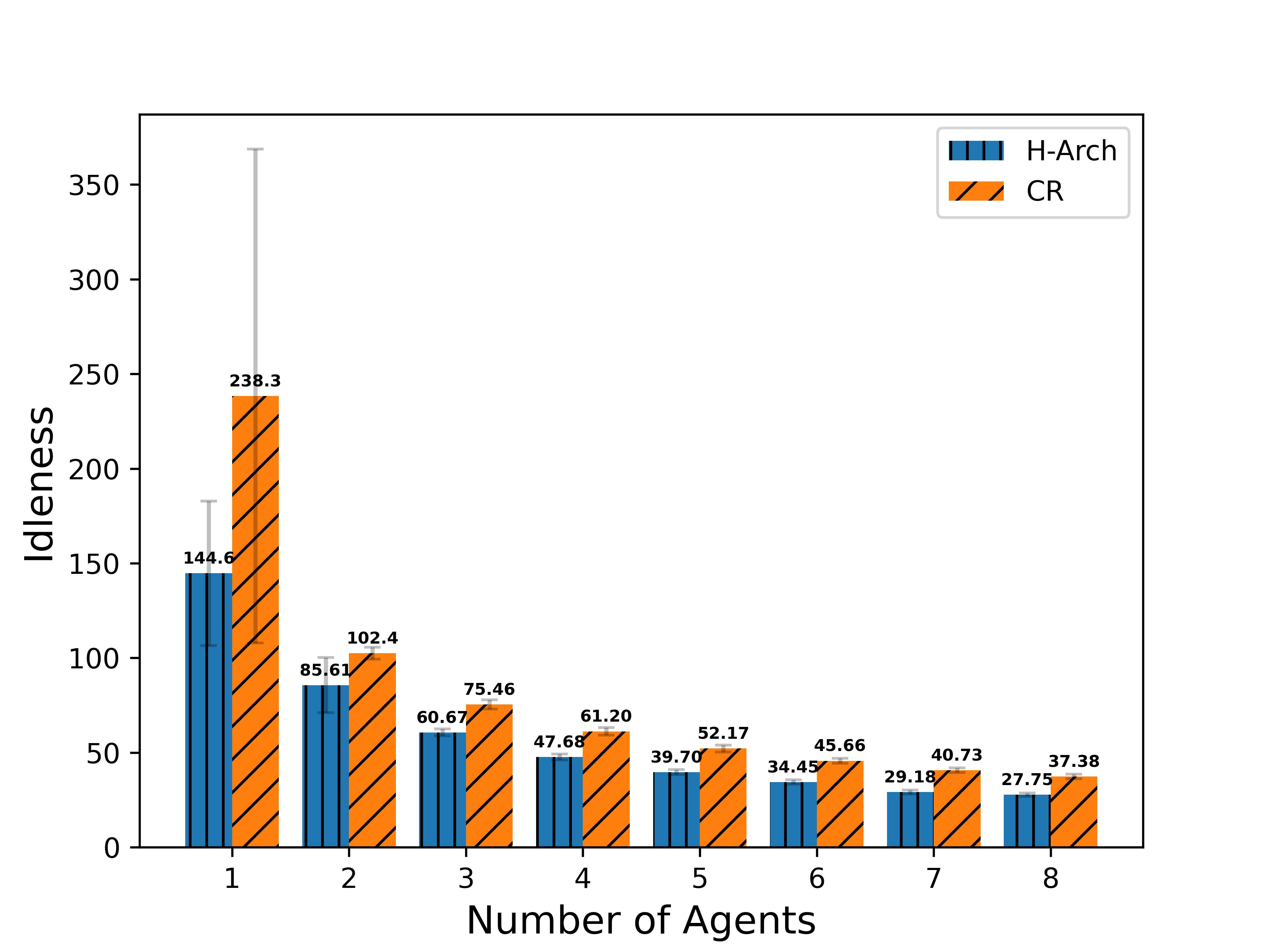}
            \caption{A-0.2 $\overline{MAX^h(G)}$}
        \end{subfigure} \hfill
        \begin{subfigure}{0.49\textwidth}
            \includegraphics[width=\linewidth]{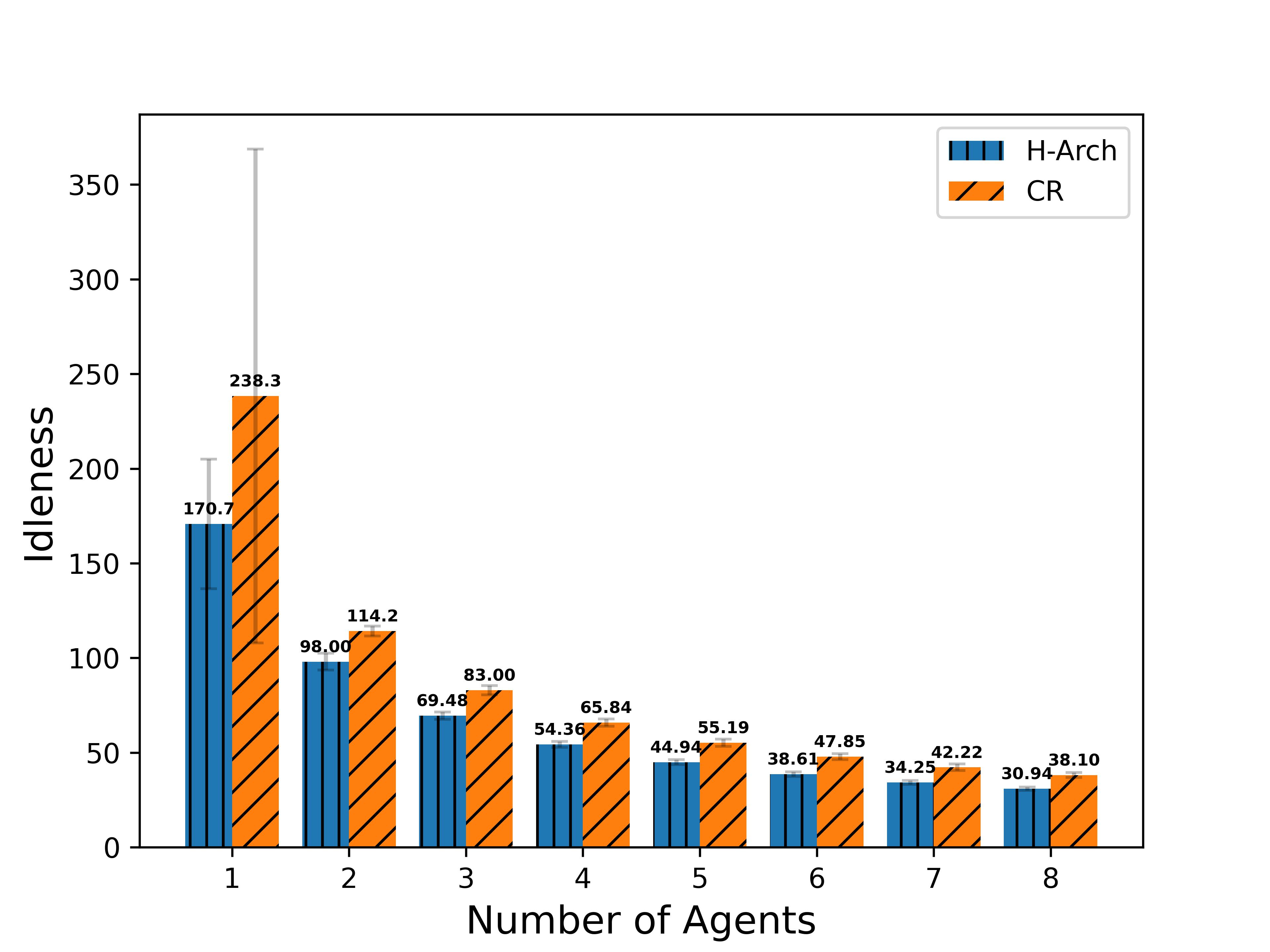}
            \caption{B-0.2 $\overline{MAX^h(G)}$}
        \end{subfigure} \hfill
        
        \begin{subfigure}{0.49\textwidth}
            \includegraphics[width=\linewidth]{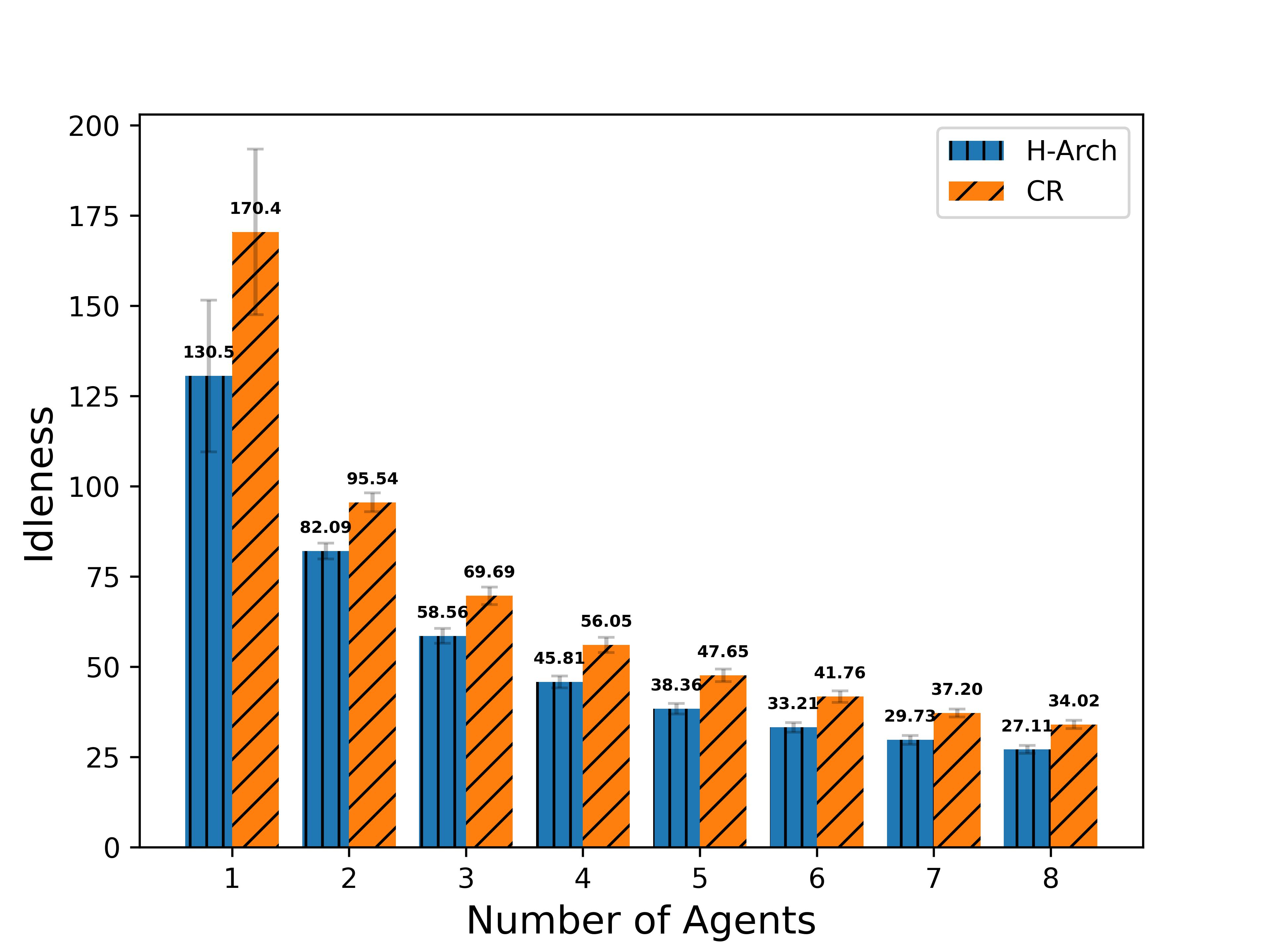}
            \caption{C-0.2 $\overline{MAX^h(G)}$}
        \end{subfigure} \hfill
        \begin{subfigure}{0.49\textwidth}
            \includegraphics[width=\linewidth]{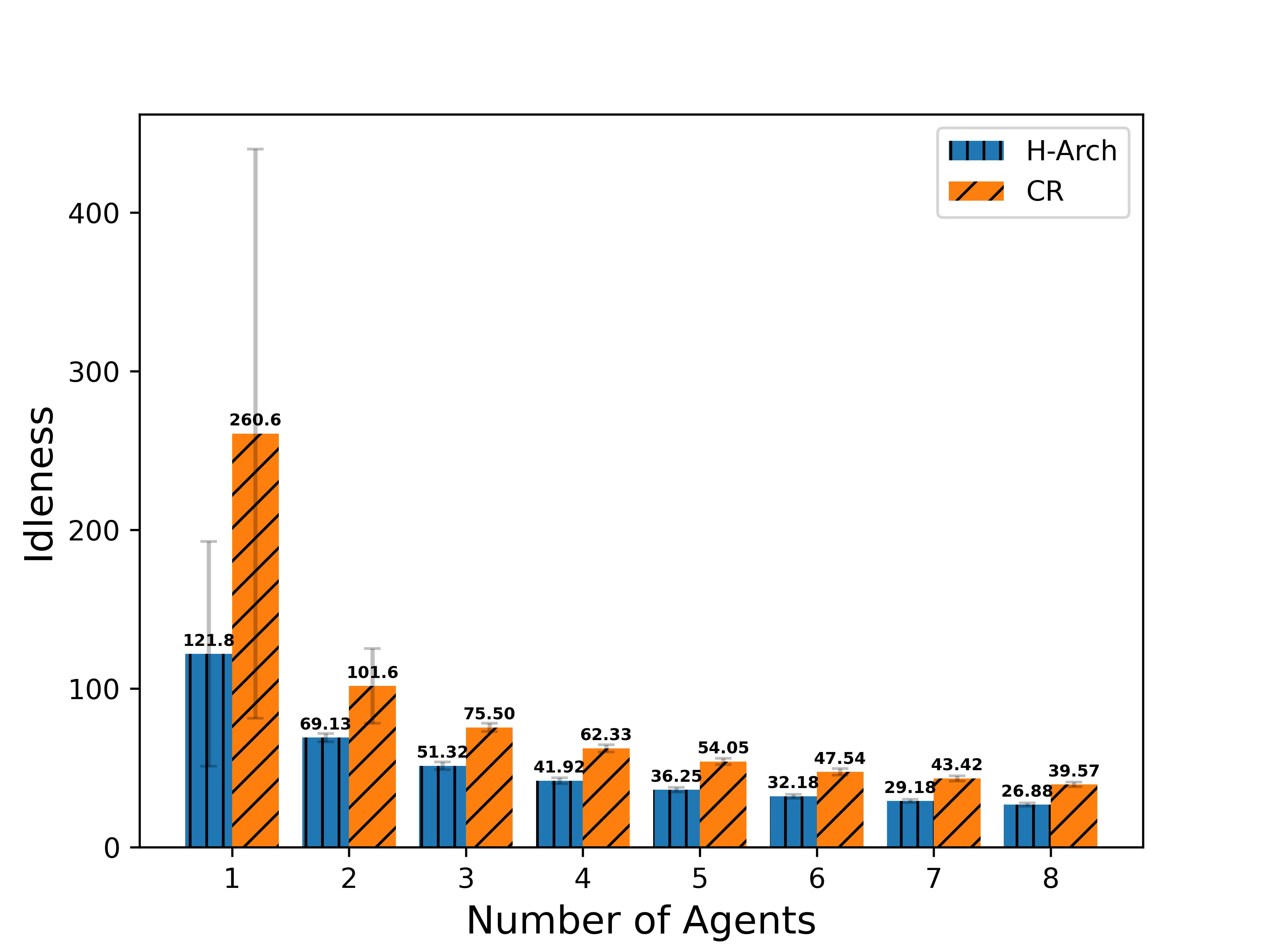}
            \caption{D-0.2 $\overline{MAX^h(G)}$}
        \end{subfigure} \hfill

\caption{Patrolling performance result ($\overline{MAX^h(G)}$) of proposed deep MARL-based models and CR strategy on four maps with $b_l=0.2$}
\label{fig:pp-eval-avg-0.2}
\end{figure}

\begin{figure}[htb]
        
        \begin{subfigure}{0.49\textwidth}
            \includegraphics[width=\linewidth]{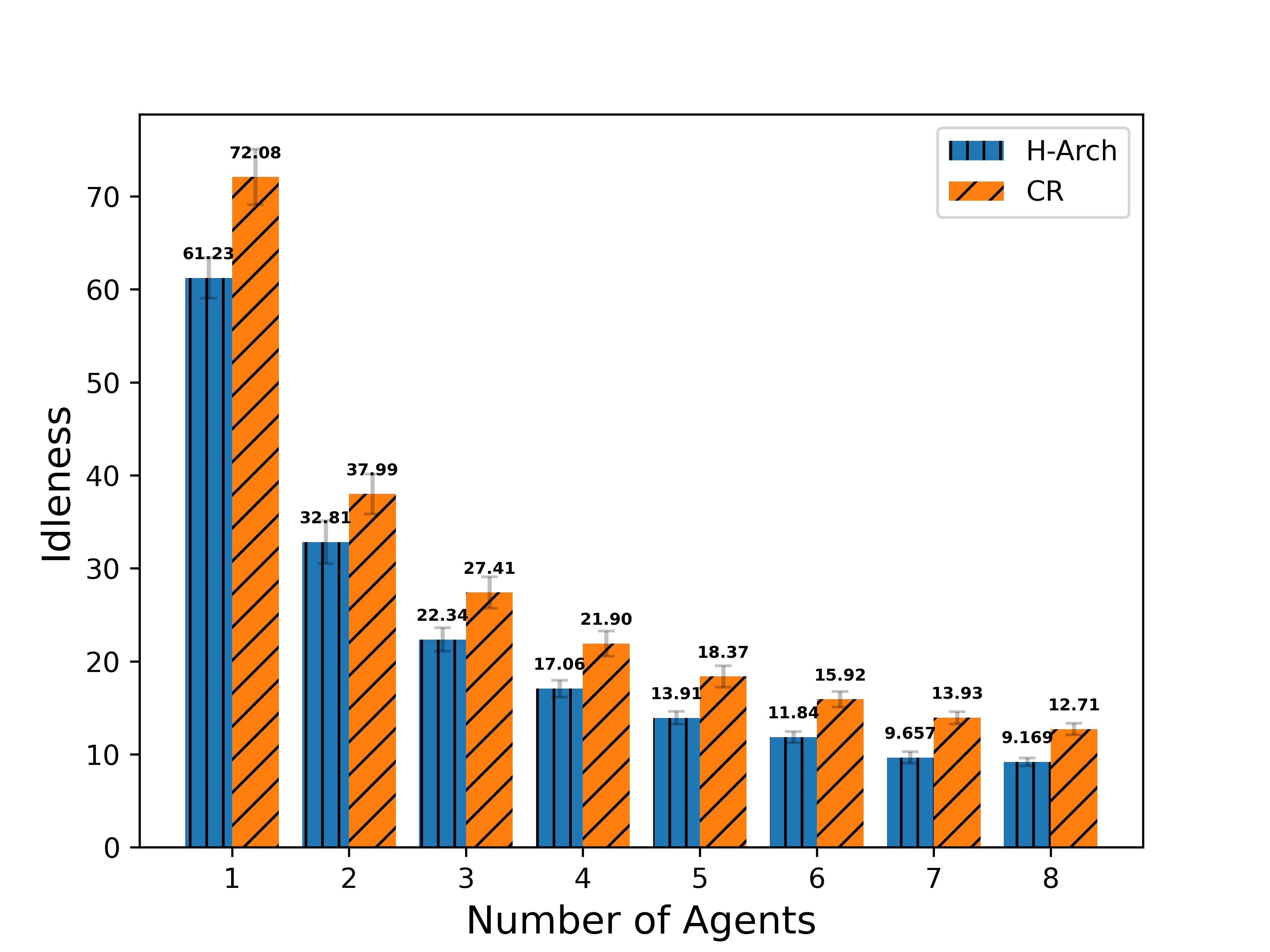}
            \caption{A-0.2 $AVG^h(G)$}
        \end{subfigure} \hfill
        \begin{subfigure}{0.49\textwidth}
            \includegraphics[width=\linewidth]{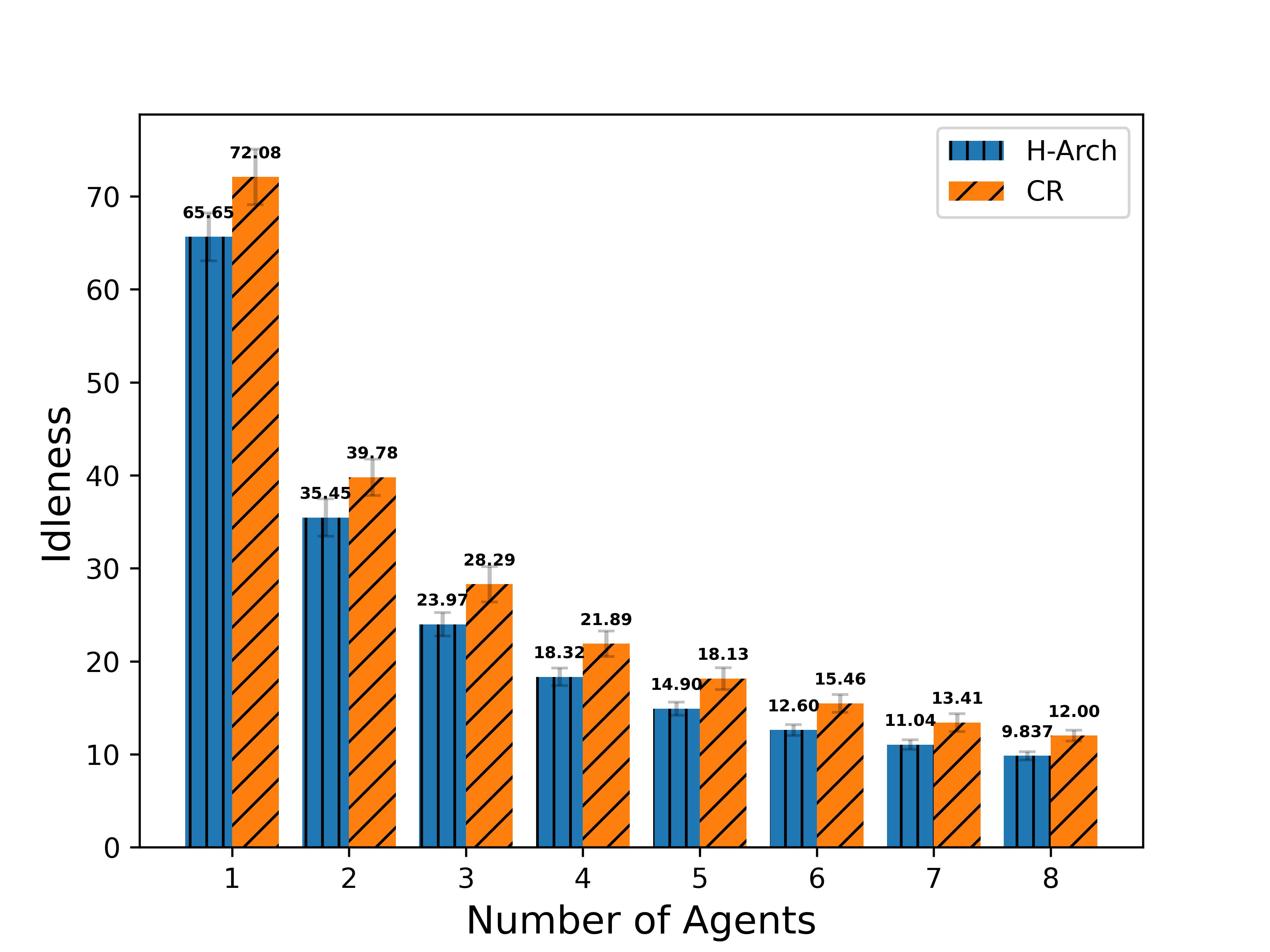}
            \caption{B-0.2 $AVG^h(G)$}
        \end{subfigure} \hfill
        
        \begin{subfigure}{0.49\textwidth}
            \includegraphics[width=\linewidth]{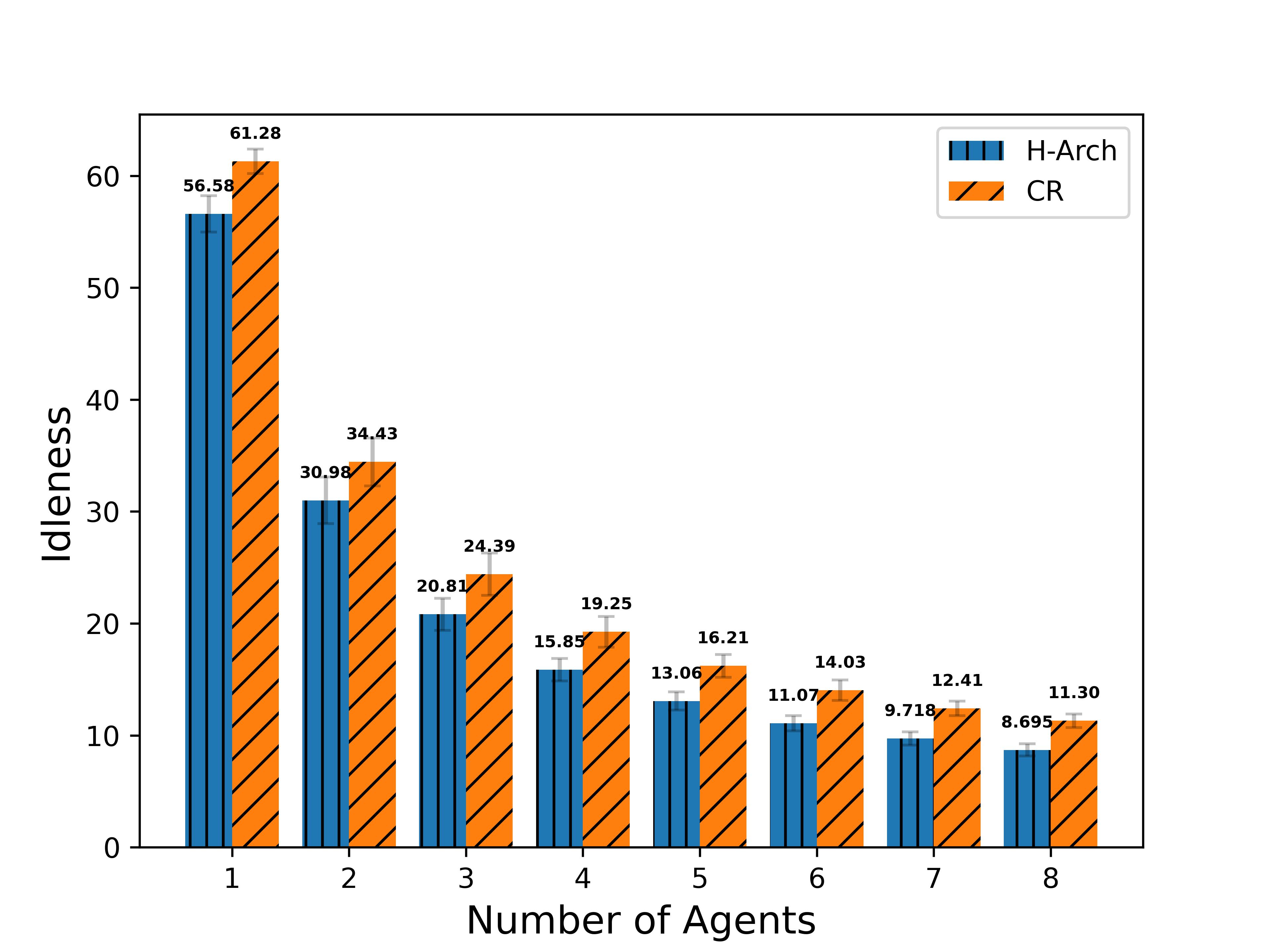}
            \caption{C-0.2 $AVG^h(G)$}
        \end{subfigure} \hfill
        \begin{subfigure}{0.49\textwidth}
            \includegraphics[width=\linewidth]{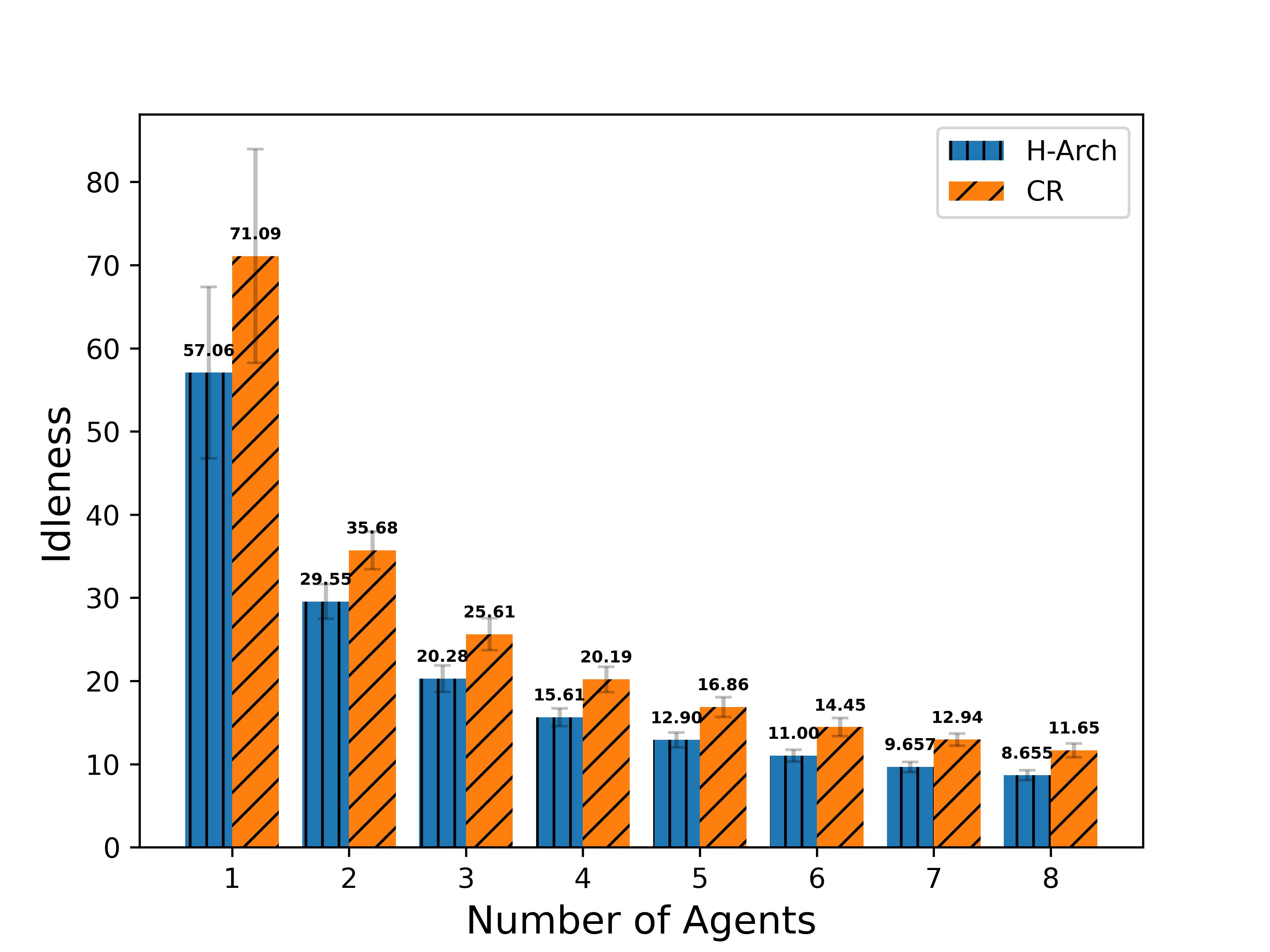}
            \caption{D-0.2 $AVG^h(G)$}
        \end{subfigure} \hfill
    
\caption{Patrolling performance result ($AVG^h(G)$) of proposed deep MARL-based models and CR strategy on four maps with $b_l=0.2$}
\label{fig:pp-eval-max-0.2}
\end{figure}

\clearpage
\section{Fault-tolerance and ability to cooperate with supplementary agents evaluation results on Map B, C, D with $b_l=0.1$} \label{app:long}

\begin{figure}[htb]
        
        \begin{subfigure}[t]{0.32\textwidth}
            \includegraphics[width=\linewidth]{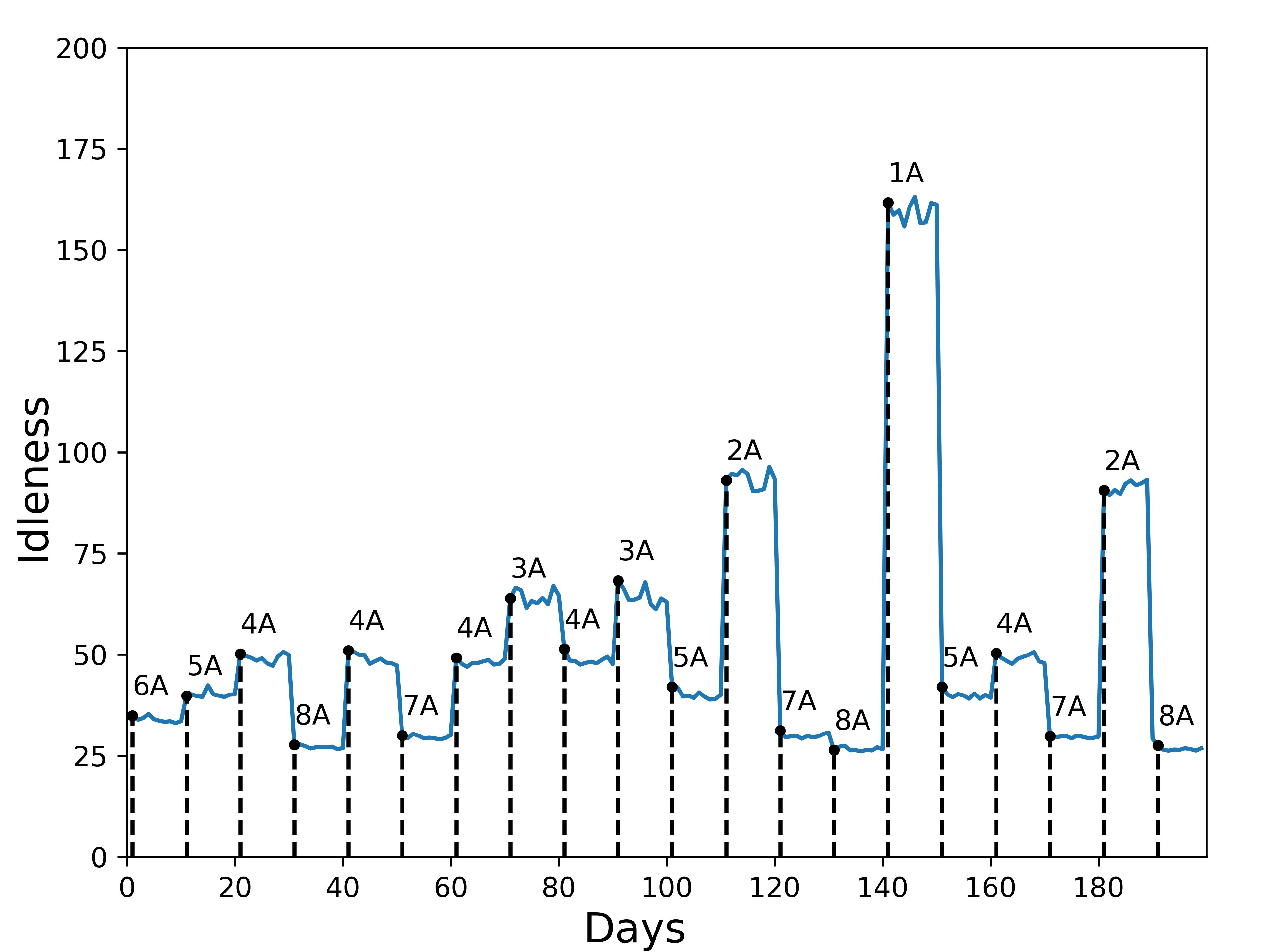}
            \caption{$\overline{MAX^h(G)}$ of Model B-$0.1$}
        \end{subfigure} \hfill
        \begin{subfigure}[t]{0.32\textwidth}
            \includegraphics[width=\linewidth]{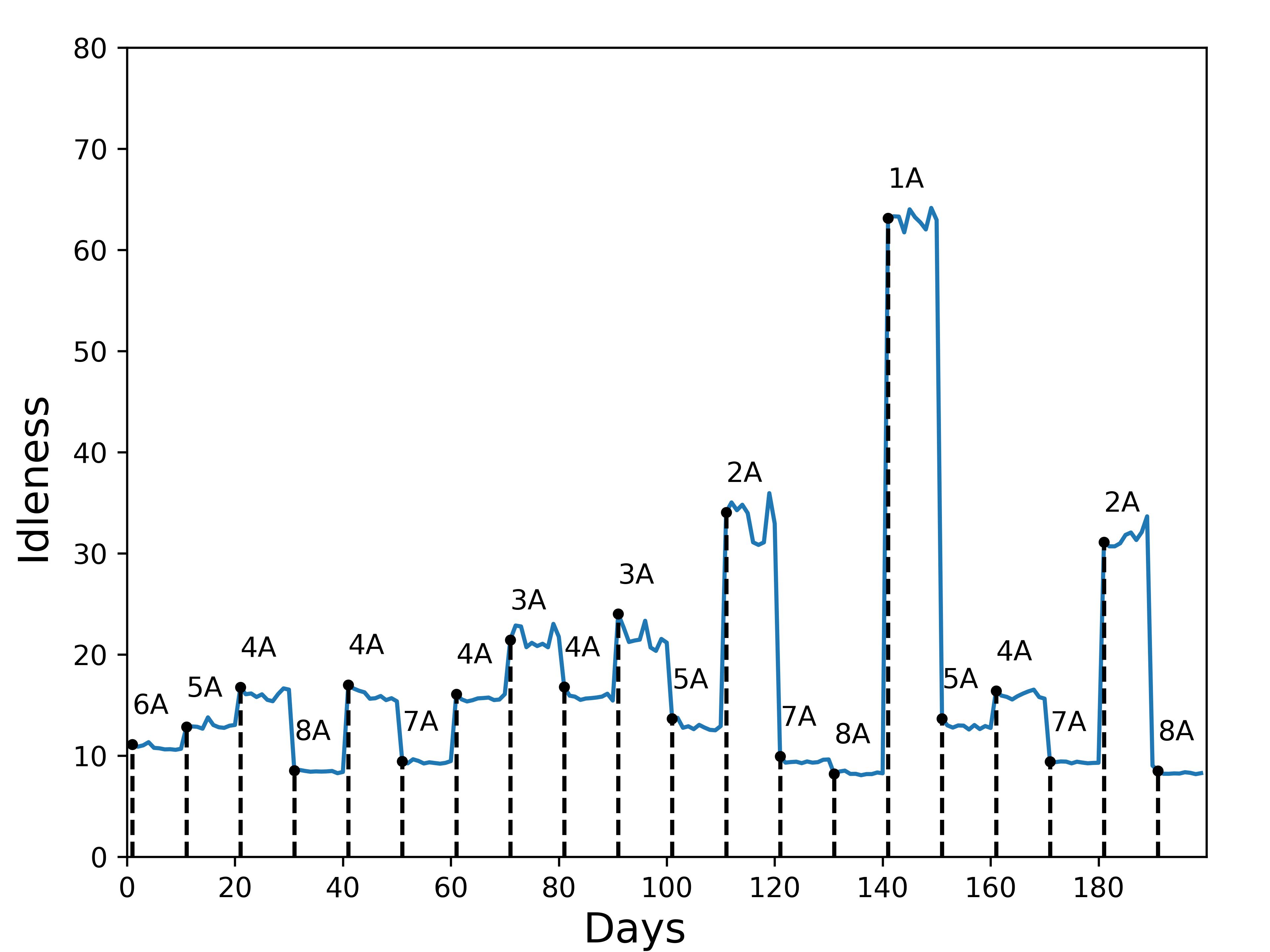}
            \caption{$AVG^h(G)$ of Model B-$0.1$}
        \end{subfigure} \hfill
        \begin{subfigure}[t]{0.32\textwidth}
            \includegraphics[width=\linewidth]{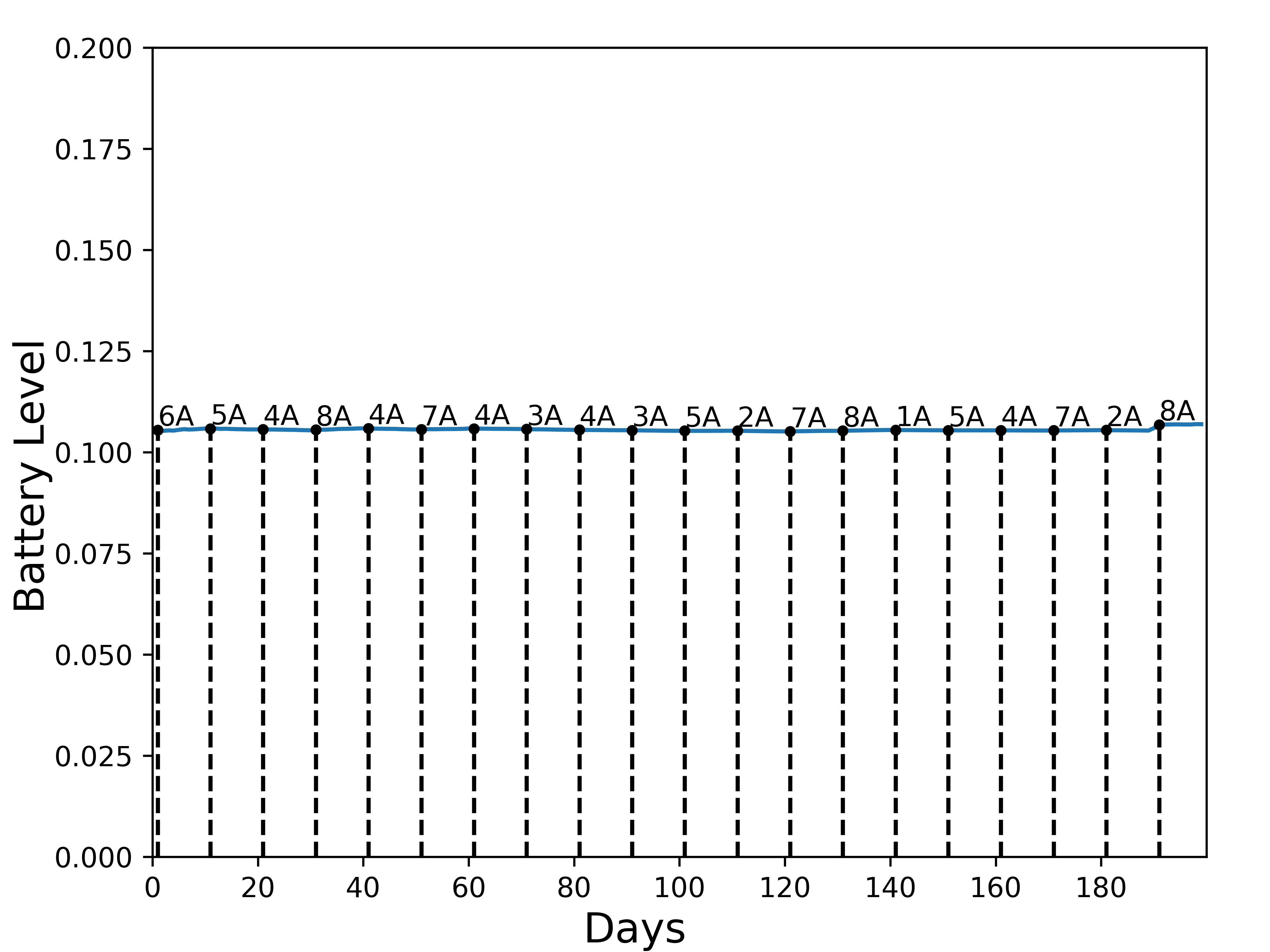}
            \caption{Battery remaining when recharging of Model B-$0.1$}
        \end{subfigure}

        \begin{subfigure}[t]{0.32\textwidth}
            \includegraphics[width=\linewidth]{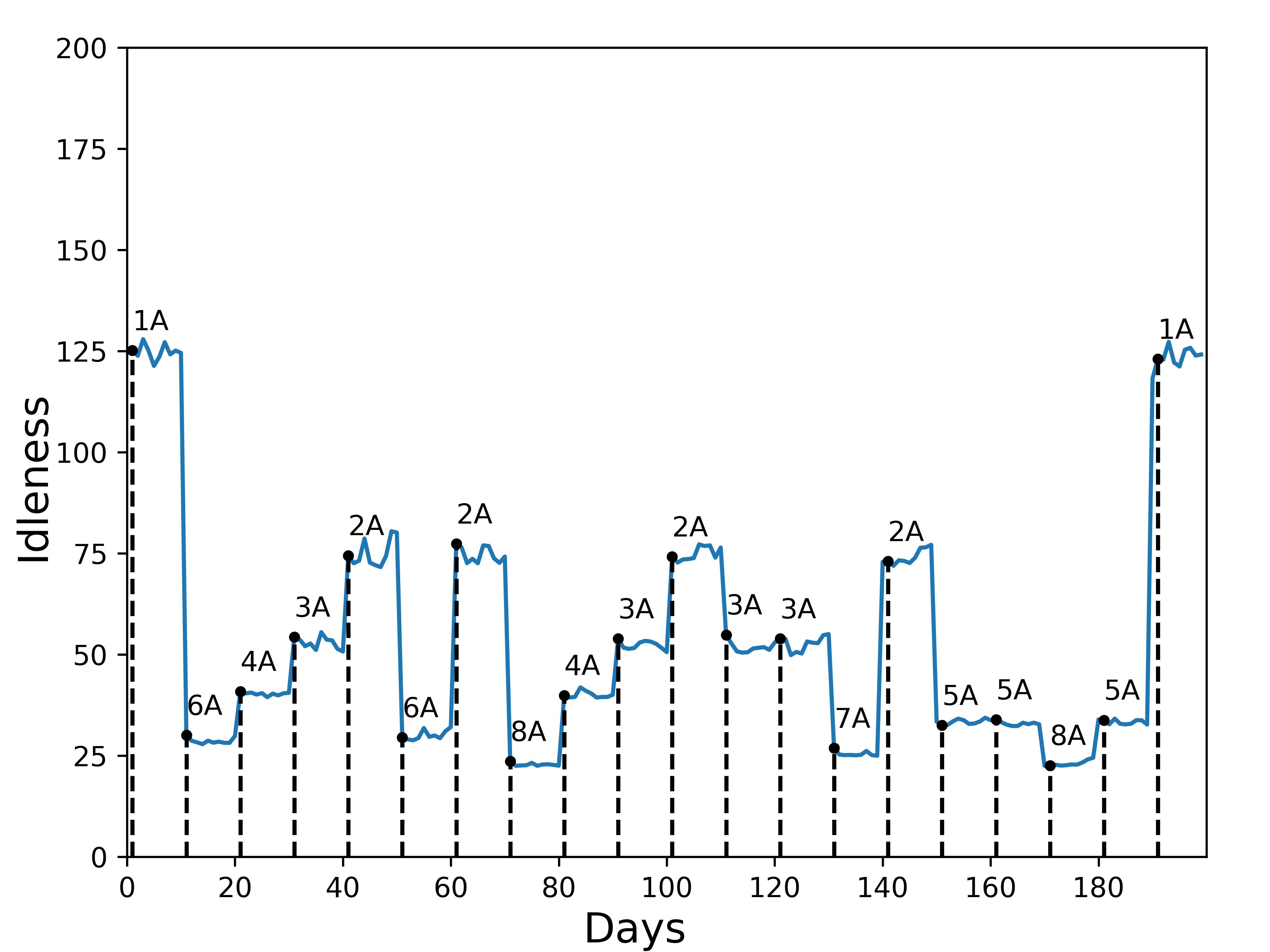}
            \caption{$\overline{MAX^h(G)}$ of Model C-$0.1$}
        \end{subfigure} \hfill
        \begin{subfigure}[t]{0.32\textwidth}
            \includegraphics[width=\linewidth]{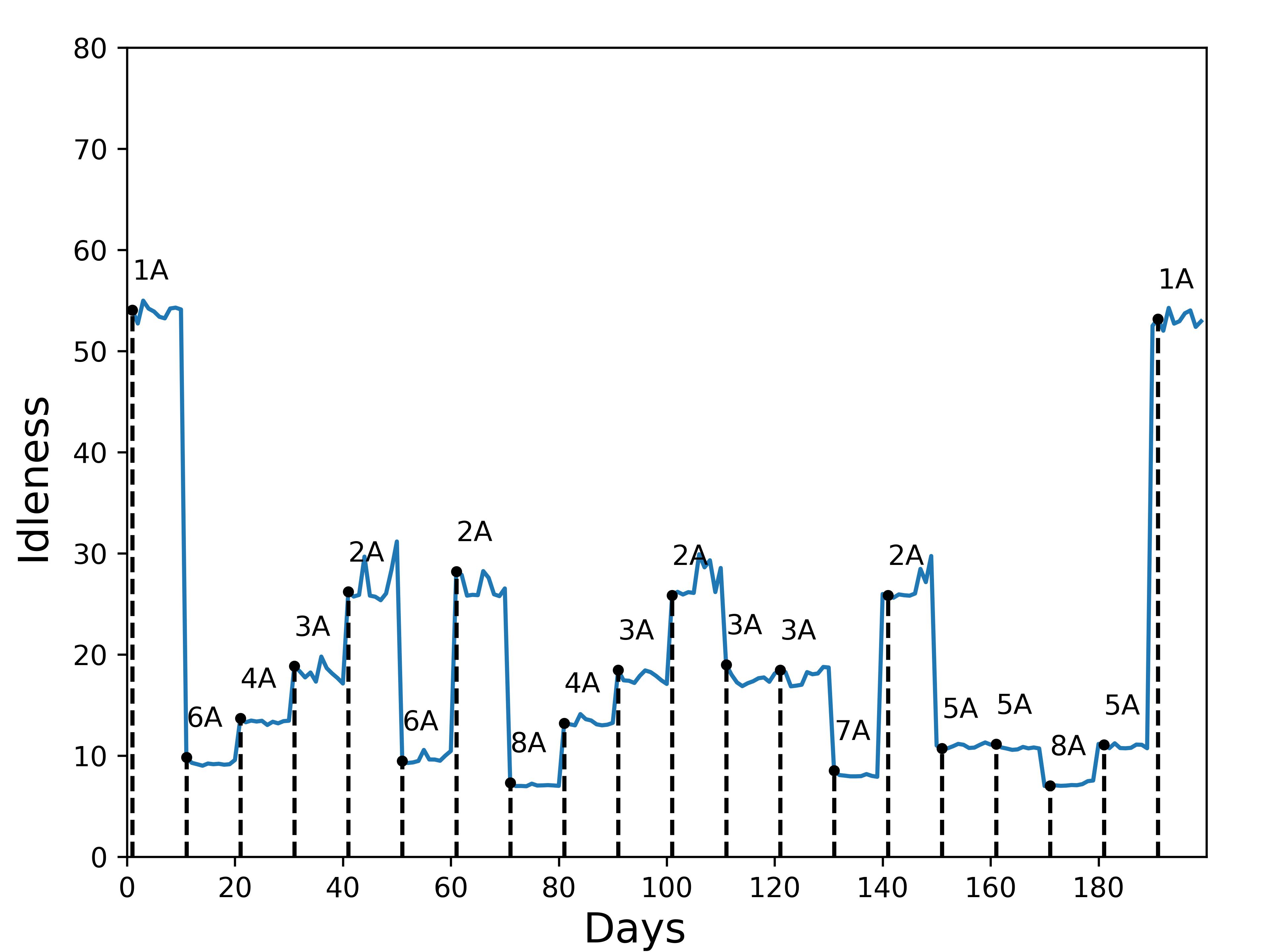}
            \caption{$AVG^h(G)$ of Model C-$0.1$}
        \end{subfigure} \hfill
        \begin{subfigure}[t]{0.32\textwidth}
            \includegraphics[width=\linewidth]{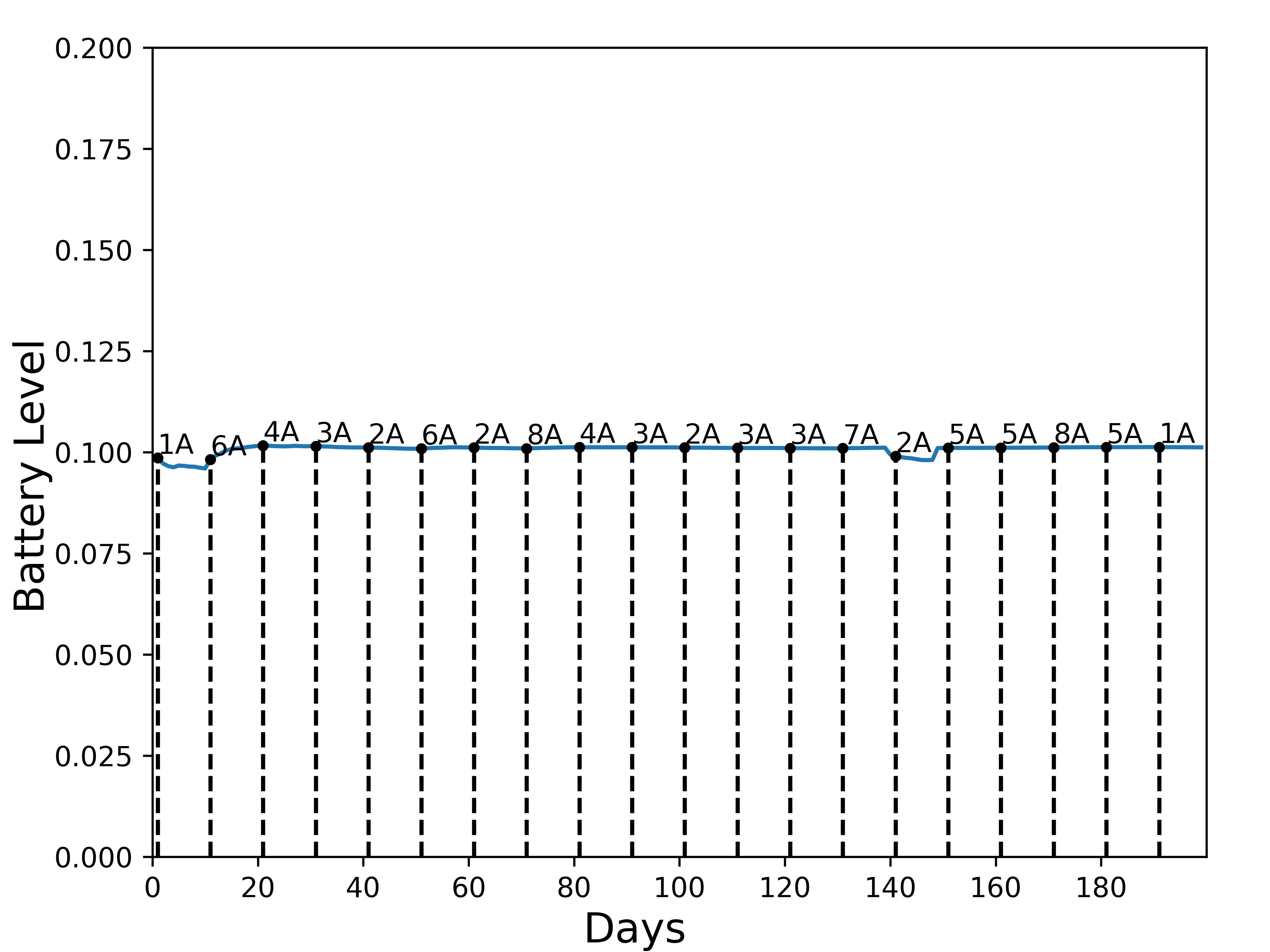}
            \caption{Battery remaining when recharging of Model C-$0.1$}
        \end{subfigure}

        \begin{subfigure}[t]{0.32\textwidth}
            \includegraphics[width=\linewidth]{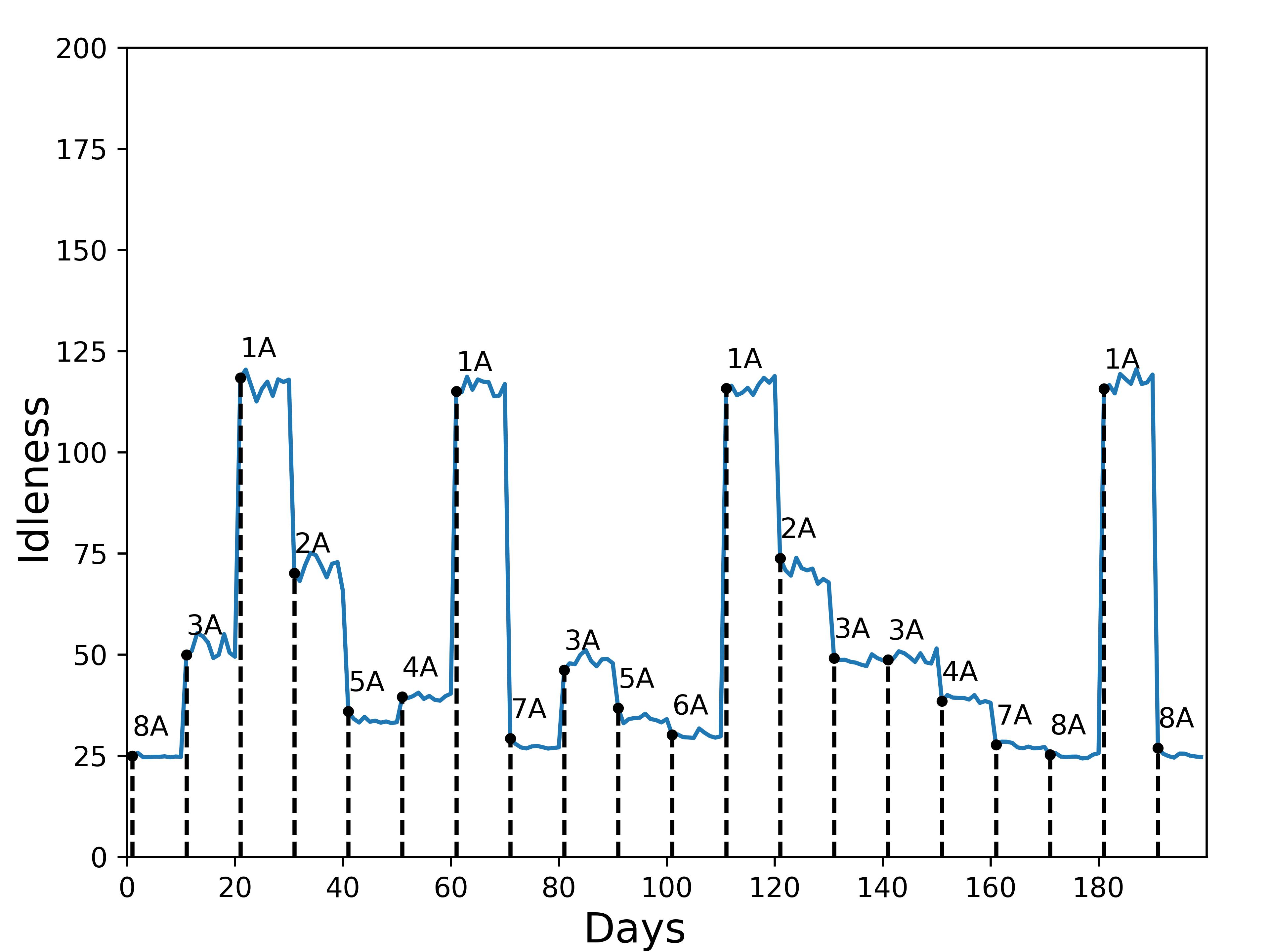}
            \caption{$\overline{MAX^h(G)}$ of Model D-$0.1$}
        \end{subfigure} \hfill
        \begin{subfigure}[t]{0.32\textwidth}
            \includegraphics[width=\linewidth]{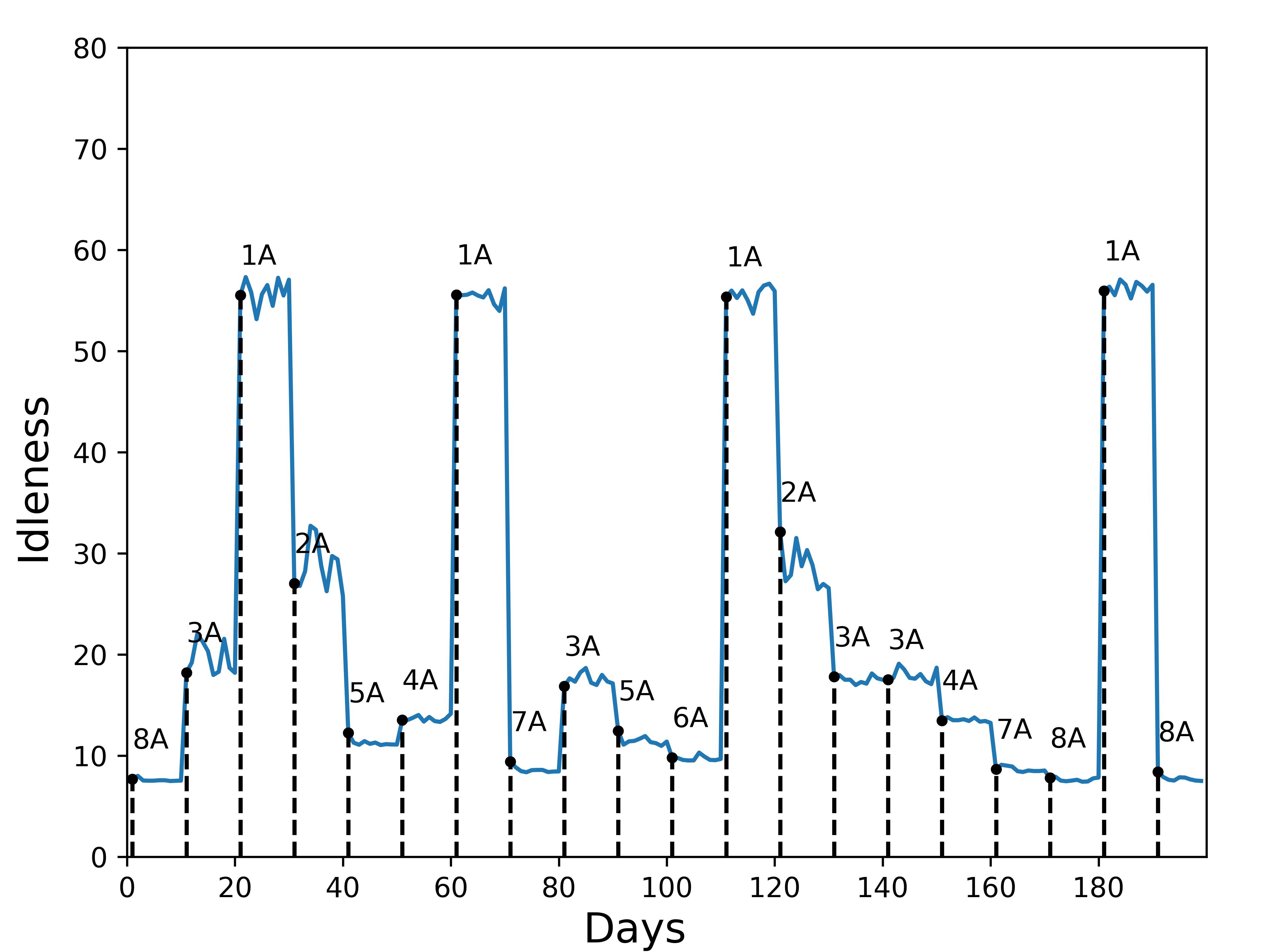}
            \caption{$AVG^h(G)$ of Model D-$0.1$}
        \end{subfigure} \hfill
        \begin{subfigure}[t]{0.32\textwidth}
            \includegraphics[width=\linewidth]{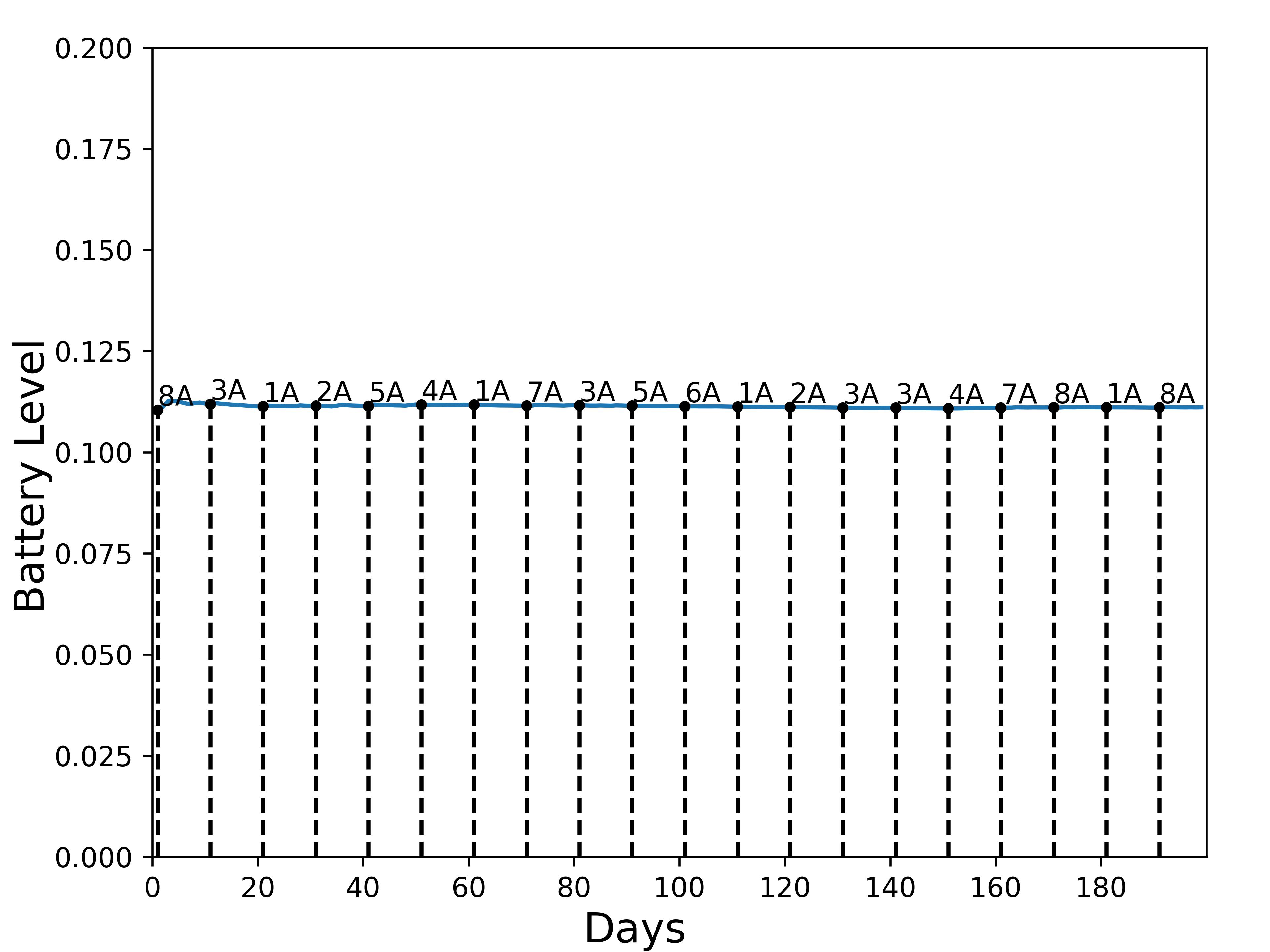}
            \caption{Battery remaining when recharging of Model D-$0.1$}
        \end{subfigure}
    
\caption{$\overline{MAX^h(G)}$), $AVG^h(G)$, and Battery remaining when recharging of model B/C/D-$0.1$ when patrolling with a varying number of agents in test episodes with a horizon of $200$ days. The data are measured daily based. The vertical dash line marks the time step when agent failures occur and when supplementary agents are introduced. The label $nA$ over the solid line indicates the number of patrolling agents during the time between two vertical dash lines. For example, $3A$ represents there are $3$ patrolling agents. }
\label{fig:long-bcd}
\end{figure}